\documentclass{article}

\usepackage[preprint]{neurips_2021}

\usepackage[T1]{fontenc}
\usepackage[T1]{fontenc}    
\usepackage[breaklinks]{hyperref}       
\usepackage{url}            
\usepackage{booktabs}       
\usepackage{amsfonts}       
\usepackage{nicefrac}       
\usepackage{microtype}      
\usepackage{xcolor}         

\newcommand{\ds}{\displaystyle }
\newcommand{\la}{\langle }
\newcommand{\ra}{\rangle }
\newcommand{\ub}[1]{\underline{\mathbf{#1}}}
\newcommand{\ml}{\mathcal{L} }

\usepackage{caption}
\usepackage{subcaption}
\usepackage{enumitem}
\usepackage{amsmath}
\usepackage{amsfonts}
\usepackage{amssymb}
\usepackage{float}

\usepackage{tikz}
\usepackage[ruled,vlined]{algorithm2e}
\SetKwInput{KwInit}{Initialization}
\usepackage[symbol]{footmisc}

\newcommand*\circled[1]{\tikz[baseline=(char.base)]{
            \node[shape=circle,draw,inner sep=2pt] (char) {#1};}}
\usepackage{graphicx}

\title{Eliminating Multicollinearity Issues in Neural Network Ensembles: Incremental, Negatively Correlated, Optimal Convex Blending}

\author{Pola-Lydia Lagari\\
 School of Nuclear Engineering, Purdue University\\
  West Lafayette, IN 47907 \\
  \texttt{lydialagaris@gmail.com}\\ 
   \And
   Lefteri H. Tsoukalas \\
   School of Nuclear Engineering, Purdue University \\
   West Lafayette, IN 47907, USA \\
   \texttt{tsoukala@purdue.edu}\\
   \AND
   Salar Safarkhani \\
   Krannert School of Management, Purdue University \\
   West Lafayette, In 47907, USA \\
   \texttt{salar.safarkhani@gmail.com} \\
   \And
   Isaac E. Lagaris \\
   Department of Computer Science and Engineering, University of Ioannina \\
   Ioannina 45500, Greece \\
   \texttt{lagaris@cs.uoi.gr} \\
}

\begin{document}

\maketitle

\begin{abstract}
Given a \{features\,,\,target\} dataset,  we introduce an incremental algorithm that constructs an  aggregate regressor, using an ensemble of neural networks. 
It is well known that ensemble methods suffer from the multicollinearity issue, which is the manifestation of redundancy arising mainly due to the common training-dataset.  
In the present incremental approach, at each stage we optimally blend the aggregate regressor  with a newly trained neural network under a convexity constraint which, if necessary, induces negative correlations. 
Under this framework, collinearity issues do not arise at all, rendering so the method both accurate and robust.
\end{abstract}
\section{Introduction}
The combination of estimators has been studied by many research teams. \citet{Wolpert:1992}\footnote[2]{This work was available since 1990 as LA-UR-90-3460 technical report} was among the first who studied the effect of estimator combination on the approximation accuracy. 
His work was followed up by \citet{Breiman:1996}, where an explanation for imposing convexity constraints on the linear combination coefficients was offered, based on grounds of generalization quality. \citet{Perrone:1993}  started a systematic study, developing the ``Basic Ensemble Method'' and the ``Generalized Ensemble Method'', that concluded in \citet{Perrone_thesis:1993} PhD Thesis. 
Independently, and around the same time, \citet{HashemDER:1993, Hashem:1995} developed an optimal linear combination method that led to \citet{Hashem:1993} PhD Thesis, and additional \mbox{publications, \citet{Hashem:1996, Hashem:1997}.} 
The potential of ensemble methods to offer improved accuracy was also stressed by \citet{Krogh:1994}, and by \citet{Meir:1995} whose work focused particularly on small and noisy datasets. 
Early work on combining classification neural networks, was published by \citet{Hansen:1990} and later by \citet{Leblanc:1996} and \citet{ZHOU:2002}.
A problem that emerges when combining many estimators is that of collinearity. This is due to the linear dependence of estimators that are trained on the same data. The relevant correlation matrix involved, becomes near singular or even singular and one has to discard a number of estimators in order to circumvent the problem, for example by using techniques such as principal component analysis, \citet{Merz:1999}. 
A different approach was followed by \citet{LIU_YAO:1999} and \citet{Chen_Yao:2009}, termed ``negative correlation'', a technique that encourages the formation of diversity among the estimators during the training process, by adding a proper term in the loss function. 
Diversity management through negative correlation has been examined and reviewed by \citet{Brown:2005}, \citet{Brown_Yao:2005}, \citet{Chan:2005} and by \citet{Reeve_Brown:2018} among others.
For ensembles of classifiers, \citet{Tumer:2002}, have presented a fusion approach based on order statistics.

The structure of this paper is as follows. In section \ref{ANALYSIS}, we first provide a brief background on the collinearity issue, and we proceed with the description and analysis of the proposed method. In section \ref{ALGORITHM}, we lay out the algorithmic procedure aiming to aid the implementation. In section \ref{NUMERICAL}, we present the results of performed  numerical experiments on a number of test functions and network architectures. A summary with conclusions and thoughts for future research are given in section \ref{SUMMARY}. There are also two short appendices describing technical matters.
\section{Analysis and description of the approach} \label{ANALYSIS}
Let us begin with a few definitions. The problem is to approximate an unknown function $y(x)$ by a parametric model $f(x)$,
given  a training set ($T_r$) and a test set ($T_s$):
\begin{eqnarray}
 T_r &=&\{x_i, y_i= y(x_i)\}_{i=1,M}, \\
 T_s &=& \{\hat{x}_i,\hat{y}_i= y(\hat{x}_i)\}_{i=1,L}.
\end{eqnarray}
The expected values of a function $f(x)$ over $T_r$ and over $T_s$, are denoted as $\la f \ra$ and $\la f \ra_s$ correspondingly, and are given by:
\begin{eqnarray}
\la f \ra \equiv \frac{1}{M}\sum_{i=1}^{M} f(x_i), \\
\la f \ra_s \equiv \frac{1}{L}\sum_{i=1}^{L} f(\hat{x}_i).
\end{eqnarray}
The ``misfit''  of a function $f(x)$ is defined as:
\begin{equation}
m_f(x) \equiv f(x)-y(x),
\end{equation}
In this article we will adopt for the model function $f(x)$, neural networks of various architectures and activation functions, all trained to fit  $y(x)$. Let these networks be denoted as:
\begin{equation}
N_j(x),\  \forall j=1,2,\cdots, K_e,
\end{equation}
 with corresponding Mean Squared Error (MSE) values:
\begin{equation}
\la m_j^2 \ra \equiv \la(N_j -y)^2\ra .
\end{equation}
We will assume that each network $N_j(x)$ has  zero bias over $T_r$, namely: 
\begin{equation}
\la m_j\ra = \la N_j -y\ra = 0,
\end{equation}
a property that can be easily arranged to hold (see Appendix \ref{appendix:A}).\\
Let the ensemble estimator be a convex linear combination of ensemble members $N_i(x)$, 
\begin{align}
 N_e(x)=\sum_{i=1}^{K_e}a_i N_i(x).
\end{align}
Convexity imposes the constraints:
\begin{align}\label{CONVEXCON}
\sum_{i=1}^{K_e}a_i = 1 \hbox{\ and \ } a_i\geq 0 ,\ \forall i=1,2,\cdots,K_e.
\end{align}
The ensemble misfit is:
\begin{equation}
M_e(x) = N_e(x)-y(x)=\sum_{i=1}^{K_e}a_i m_i(x),
\end{equation}
and its mean squared error is given by: 
\begin{align}
\la M_e^2\ra = \sum_{i,j}a_ia_j\la m_im_j\ra.
\end{align}
To minimize $\la M^2_e \ra $, subject to the convexity constraints (\ref{CONVEXCON}), it is necessary to construct the relevant  Lagrangian which is written using multipliers $\lambda \in R$ and $\ds \mu \in R^{K_e}$ as:
\begin{eqnarray}
\ml(a,\lambda,\mu)=\frac{1}{2}a^TCa-\lambda(e^Ta-1)-\mu^Ta,\\
\nonumber
\hbox{where\ } e^T=(1,1,\cdots,1), \hbox{\ and\ } C_{ij}=\la m_i m_j\ra.
\end{eqnarray}
The first order optimality conditions and the equality constraint $e^Ta=1$, yield:
\begin{eqnarray}
\lambda &=& \frac{1-e^TC^{-1}\mu}{e^TC^{-1}e},\\
\label{SOLU} 
a&=&\frac{C^{-1}e}{e^TC^{-1}e}+\left[C^{-1} - \frac{C^{-1}ee^TC^{-1}}{e^TC^{-1}e} \right]\mu.
\end{eqnarray}
Eq. (\ref{SOLU}) together with  $a_i \geq 0,\ \mu_i \geq 0$ and the complimentarity conditions $a_i\mu_i =0$, determine the sought solution. Note that if the correlation matrix $C$ is rank deficient, which is often the case due to collinearity, the inverse $C^{-1}$ is not defined and the above procedure is not applicable.
Some methods, in order to avoid the multicollinearity issue, use a simple ensemble average (i.e. $a_i=\frac{1}{K_e},\  \forall i=1,\cdots,K_e$). 
These methods lead to non-optimal results since they neglect the relative importance of the various components, the remedy being the consideration of very large ensembles ($K_e \gg 1$), a tactic that renders the procedure  excessive and inefficient as well.

In the present article we propose a novel method  which determines the coefficients $a_i$ without having to deal with a potentially singular system, avoiding so the above mentioned issues due to multicollinearity. 
Instead, an aggregate network is being built incrementally from ground up, by optimally blending it with a  single new network at every stage, via a convex linear combination. The new network is trained so that its misfit is negatively correlated to the misfit of the current aggregate, a constraint that is sufficient (although not necessary) to guarantee the convexity requirement. 
The method takes in account the relative importance of the different networks, satisfies the convexity requirements, and eliminates multicollinearity complications and associated numerical side effects.
\subsection{The General Framework}
Consider  two zero-bias networks $ N(x)$ and $\hat{N}(x)$, with corresponding misfits:
\begin{equation}
 M(x)=N(x)-y(x), \hbox{\ and\ \ } \hat{M}(x)=\hat{N}(x) -y(x).
\end{equation}
Then  their convex blend may be defined as:
 \begin{equation}\label{BLEND}
\tilde{N}(x) = \beta N(x)+ (1-\beta) \hat{N}(x),\ \beta \in [0,1],
\end{equation} 
with misfit 
$\tilde{M}(x)= \beta M(x)+ (1-\beta) \hat{M}(x)$, and MSE given by:
\begin{equation}
\la\tilde{M}^2\ra = \la(M-\hat{M})^2\ra\beta^2+2\la\hat{M}(M-\hat{M})\ra\beta+\la\hat{M}^2\ra.
\end{equation}
Note that  $\ds \la\tilde{M}^2\ra$ being quadratic in $\beta$,  has a minimum at: 
 \begin{equation}
 \label{beta}
 \beta^* =\max\{\min \{1,\beta_u\},0\},\ \hbox{ with\ }\beta_u =\frac{\la(\hat{M}-M)\hat{M}\ra}{\la(M-\hat{M})^2\ra},\\ 
 \end{equation}
 where $\beta_u$ is the unconstrained minimizer of $\la\tilde{M}^2\ra$. If $\beta_u \notin[0,1]$, then either $\beta^*=1$ or $\beta^*=0$ and there is no blending, as can be readily deduced by inspecting eq. (\ref{BLEND}). \\
 Note from (\ref{beta}) that  $\beta_u \in (0,1)$, implies:
 \begin{align}
\la M \hat{M}\ra < \min\{ \la M^2\ra , \la \hat{M}^2\ra\},
\end{align} 
and in this case $\beta^* = \beta_u$ leading to:
\begin{equation}
\label{MSE_min}
\la\tilde{M}^2\ra^*=\frac{\la M^2\ra\la\hat{M}^2\ra-\la M \hat{M}\ra^2}{\la M^2\ra+\la\hat{M}^2\ra-2\la M \hat{M}\ra}.
\end{equation}
From eq. (\ref{MSE_min}), one may deduce, after some manipulation, that
 $\ds \la\tilde{M}^2\ra^* \ < \ \min\{\la M^2\ra,\la \hat{M}^2\ra\}$. \\
The important conclusion is that a convex combination of two zero-bias networks may be arranged so as to yield a zero-bias network with a reduced MSE.

\subsection{Recursive Convex Blending}
The idea is to construct recursively an aggregate network, by combining each time the current aggregate with a newly trained network.
Namely, if by $N^{(a)}_k(x)$ we denote the $k^{th}$ aggregate network, then the $(k+1)^{th}$ aggregate will be given by the convex blend:
\begin{equation}
N^{(a)}_{k+1}(x)=\beta_{k+1}N^{(a)}_k(x)+(1-\beta_{k+1})N_{k+1}(x).
\end{equation}
Let $M_k(x) \equiv N^{(a)}_k(x) -y(x)$ be the aggregate misfit. Then $N_{k+1}(x)$ is trained so that its misfit $m_{k+1}(x)$ satisfies: 
\begin{equation} \la M_km_{k+1}\ra <  \min\{\la M_k^2\ra, \la m_{k+1}^2\ra\},
\end{equation}
in order to guarantee convex blending (i.e. $ \beta_{k+1} \in [0,1]$).
This is a constrained optimization problem stated formally as:
\begin{eqnarray}
&\min\la m_{k+1}^2\ra,\hbox{\ subject to:} \\
& \la M_km_{k+1}\ra \ \  \leq\ \  \min\{\la M_k^2\ra,\la m^2_{k+1}\ra\}.   \label{INEQ}
\end{eqnarray}
Note that since the positive quantity: $\min\{\la M_k^2\ra,\la m^2_{k+1}\ra\} $, is  expected to be  small, condition (\ref{INEQ}) may be satisfied by imposing that the aggregate and the new network misfits are negatively correlated or uncorrelated, i.e. $\la M_km_{k+1}\ra \leq 0$, which is a stronger requirement than   (\ref{INEQ}), and which in addition prevents $\beta_{k+1}$ from assuming the limiting no-blend values \mbox{of $0$ and $1$}\footnote[3]{For different blending restrictions see Appendix \ref{appendix:B}}.\\
\noindent
The next aggregate misfit is then given by:
\begin{eqnarray}\label{MRECUR}
M_{k+1}(x) &=& \beta_{k+1}M_k(x)+(1-\beta_{k+1})m_{k+1}(x),\\
\hbox{with\ }\beta_{k+1} &=& \frac{\la m^2_{k+1}\ra-\la M_km_{k+1}\ra}{\la(M_k-m_{k+1})^2\ra}. \label{BETAK}
\end{eqnarray}
Setting $M_1(x) = m_1(x)$ (and hence $\beta_1 = 0$), and noting that $M_k$ is a linear combination of $m_{1}(x)$, $m_{2}(x),\cdots,m_{k}(x)$, one may express $M_k$ as:
\begin{equation}
M_k(x) = \sum_{i=1}^k a^{(k)}_{i} m_{i}(x),
\end{equation}
with the coefficients $a^{(k)}_i$ given by:
\begin{equation}\label{BLCOEF}
a^{(k)}_i = 
\begin{cases}
1-\beta_k, \hspace{2cm} i=k\\
\displaystyle (1-\beta_i)\prod_{l=i+1}^k\beta_l, \hspace{7pt} \forall i=1,2,\cdots,k-1
\end{cases}.
\end{equation}
One may verify that  
$ \ds
\sum_{i=1}^k a^{(k)}_{i} = 1 \hbox{\ and \ }a^{(k)}_{i} \geq 0, 
$
and the aggregate network is then given by:
\begin{equation}\label{ZBFAN}
N^{(a)}_k(x) = \sum_{i=1}^k a^{(k)}_{i} N_{i}(x).
\end{equation}
Note  that the aggregate network $N^{(a)}_k(x)$ is not the simple average of the member networks as is the case in \citet{AHMAD:2009} and in the negative correlation approaches \citet{LIU_YAO:1999,Chan:2005,Chen_Yao:2009,Brown:2005}. 
The combination coefficients $a^{(k)}_{i}$ given by eq. (\ref{BLCOEF} ),
do take in account the relative importance of the different network contributions, and are being built iteratively by optimal pairwise convex blending, avoiding so problems due to multicollinearity. This is in contrast to the approach described by eq. (\ref{SOLU}) which had been adopted by several authors in the past.  
\subsection{Algorithmic Procedure and Implementation\label{ALGORITHM}}
The above results and ideas are employed to design an algorithm for practical use.

\begin{description}
\item[Initialization] \hfill 
\begin{itemize}
\item Create the empty list, \texttt{MList}, to store the aggregate misfit in each step.
\item Create the empty list, \texttt{BList}, to store $\beta$'s.
\item Create the empty list, \texttt{ModelList} to store the trained neural network model.
\item Set $b_L,\ b_U $ both $\in (0,1)$, the lower and upper bound for the $\beta$'s.
\item Set $K_e$, the number of networks to be contained in the ensemble.
\end{itemize}
\item[Main Part] \hfill
\begin{enumerate}
\item Train the first zero bias network, $N_1(x)$ and push it to the \texttt{ModelList}.
\item Set $M_1(x_i) = N_1(x_i)-y_i, \forall \{x_i,y_i\} \in T_r$ and push it to \texttt{MList}.
\item Set $\beta_1 = 0$ and push it to the \texttt{BList}.
\item $k \leftarrow 1$
\item Read $M_k(x),\  \forall x\in T_r$ from \texttt{MList}.
\item  Train the zero bias network $N_{k+1}(x)$ s.t. $ \la M_km_{k+1}\ra \  \leq\ \min\{\la M_k^2\ra,\la m^2_{k+1}\ra\}$.\\
This can be facilitated via a penalty method, i.e. by optimizing:
$$
\la m^2_{k+1}\ra + \lambda \max\{\la m_{k+1}M_k\ra,0\}
$$
for a sequence of increasing $\lambda$ values, until $\beta_{k+1} \in[b_L, b_U ]$\footnote{$\beta_{k+1}$ is calculated according to eq. (\ref{BETAK})}.
\item Calculate $M_{k+1}$ from eq. (\ref{MRECUR}) and $\la M^2_{k+1}\ra$.
\item
Push $M_{k+1}$ to \texttt{MList}, $\beta_{k+1}$ to \texttt{BList}, and $N_{k+1}$ to \texttt{ModelList}.
\item $k \leftarrow k+1$
\item If $k < K_e$ repeat from \circled{5}.
\item 
Training Task Completed.
\end{enumerate}
\item [Final Network Assembly] \hfill 

From \texttt{BList} and \texttt{ModelList}, recover the  $\beta$'s and the trained networks respectively.
\begin{enumerate}[label=(\alph*)]
\item  From $\beta_1,\cdots,\beta_{K_e}$, calculate the member coefficients $a_i^{(K_e)}$ using eq. (\ref{BLCOEF}).
\item Construct the final zero-bias aggregate network $N^{(a)}_{K_e}(x)$, according to eq. (\ref{ZBFAN}).
\end{enumerate} 

\end{description}
\section{Numerical Experiments}\label{NUMERICAL}

We consider two model functions in our numerical study,
\begin{eqnarray}
\label{EQF1}
f_1(x) &=& x\sin(x^2),\ \hbox{and} \\ 
\label{EQF2}
f_2(\mathbf{x}) &=& 
10d + \sum_{i=1}^d[x_i^2-10cos(2\pi x_i)], \hbox{\  with \ }
\mathbf{x}\in[-1.5,1.5]^d.
\end{eqnarray}
\begin{figure}[ht]
\centering
\begin{subfigure}{0.48\textwidth}
    \centering
    \includegraphics[width=\textwidth]{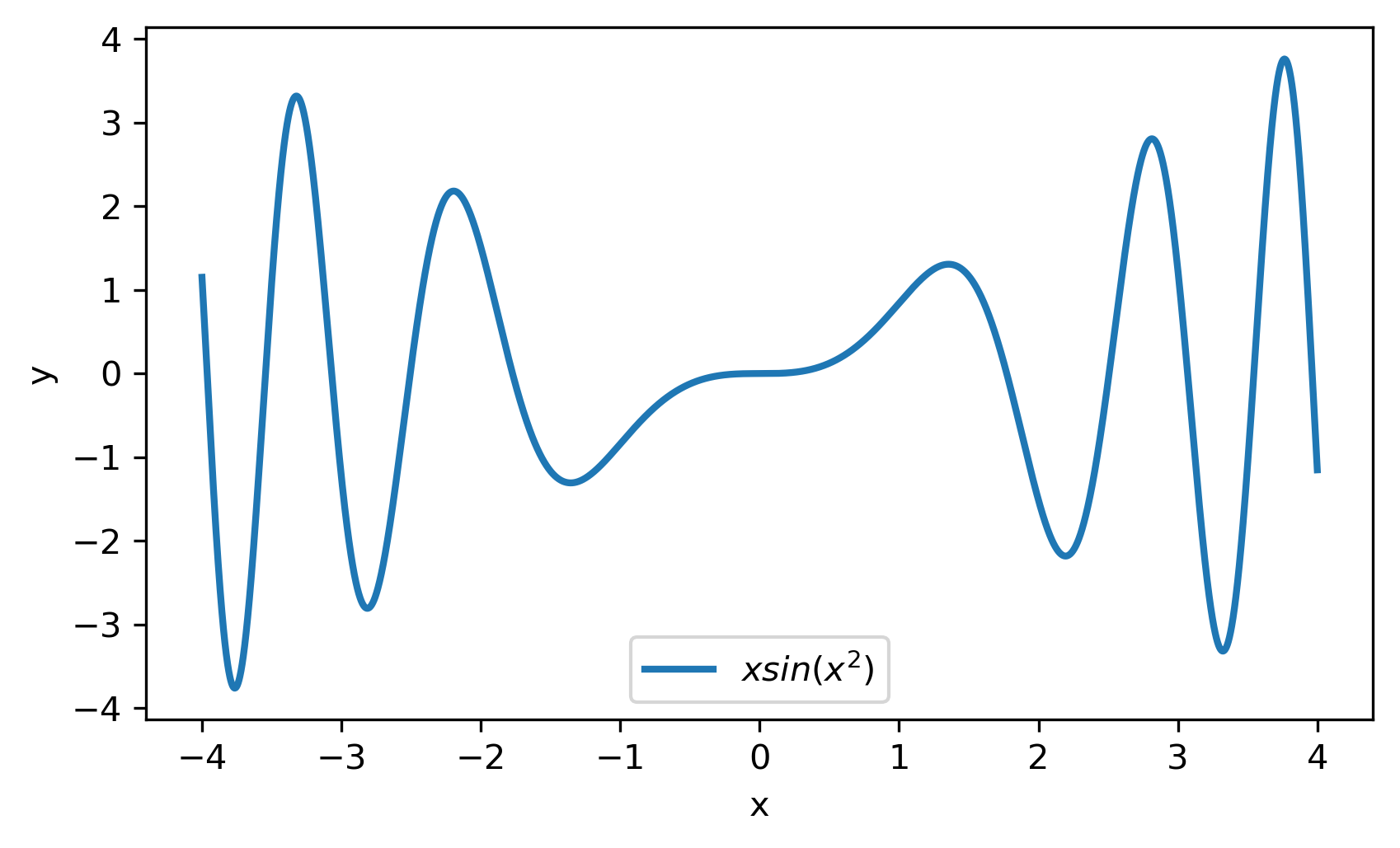}
\end{subfigure}\hfill
\begin{subfigure}{0.48\textwidth}
    \centering
    \includegraphics[width=\textwidth]{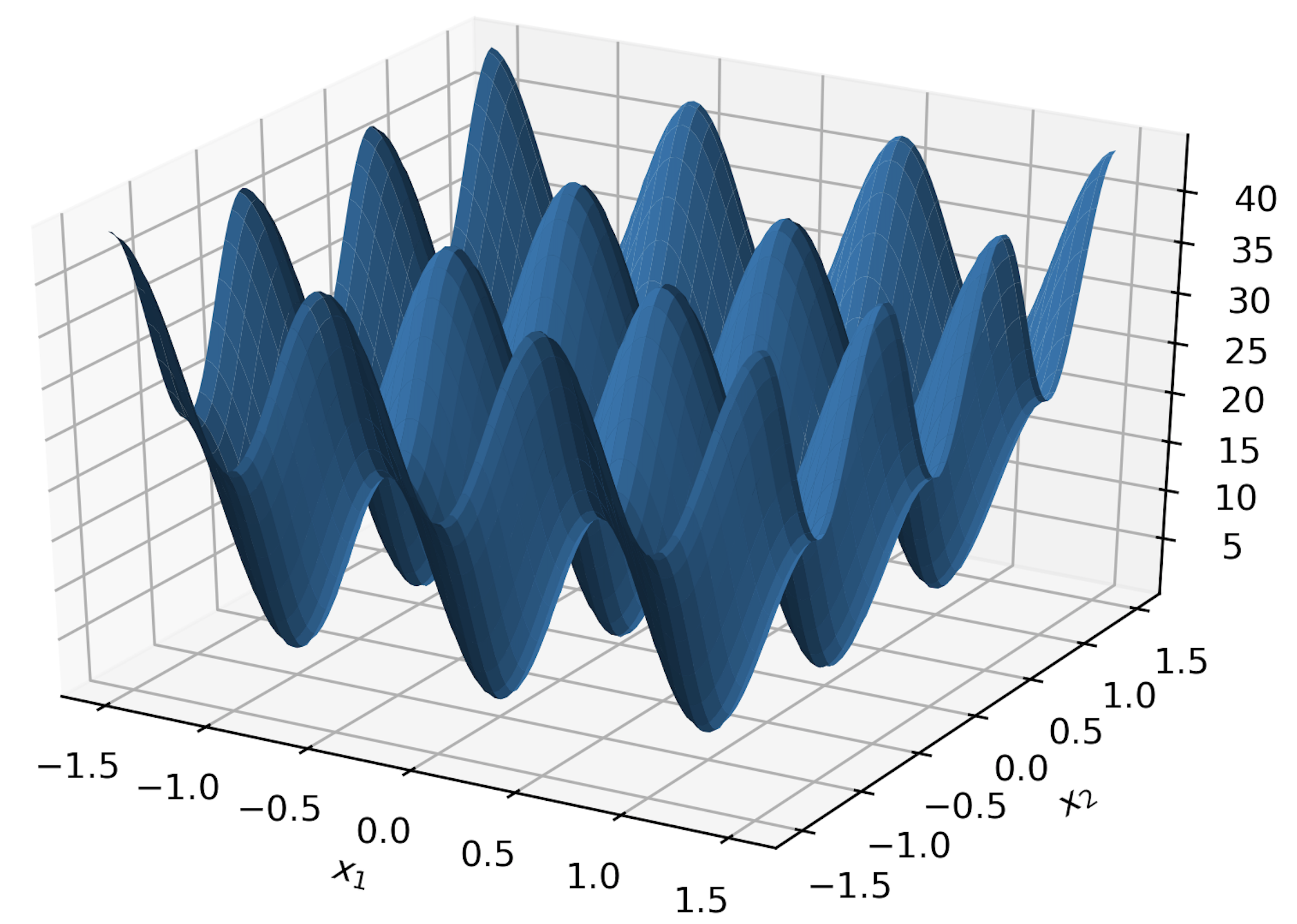}
\end{subfigure}
\caption{Plots for $f_1(x)$ of eq.(\ref{EQF1}) with $x \in [-4, 4]$, and $f_2(x)$  of eq.  (\ref{EQF2}) with $d=2$ }
\label{ONEDTEST}
\end{figure}
We  have performed four different experiments to assess the effectiveness of the method.
Common features and practices shared across all experiments are listed below.
\begin{description} 
\item[Datasets] \hfill

For every experiment we define a dataset $D$, and we split it in two disjoint sets, the training set $T_r$, and the test set $T_s$, with $D=T_r \cup T_s$ and 
$T_r \cap T_s = \emptyset$.
\item[Training] \hfill

Training was performed using the BFGS Quasi-Newton optimization algorithm, contained in ``TensorFlow Probability'' \citet{tensorflow}, with the maximum number of iterations set to 20,000.
\item[Regularization] \hfill 

We have used  $L_2$-mean weight decay, with regularization factor $\nu_{reg}$, chosen so as to inhibit excessive growing of the network weights. 
\item[Blending range] \hfill 

We set the lower and upper bounds for $\beta$'s to $b_L=0$ and $b_U=0.99$. 
\item[Penalty strategy] \hfill

In step No.~6 of the section \ref{ALGORITHM} algorithm, we set the initial penalty to $\lambda=4$.
At each subsequent iteration, until condition  $b_L<\beta_{k+1}<b_U$ is satisfied,  $\lambda$ is doubled. 
If the above condition is not satisfied within 10 iterations, we discard the model.
\item[ANN architecture] \hfill

For all the test cases we use a single hidden layer neural network of the form:\\ 
$\ds N(x)= \sum_{i=1}^{Nodes}\gamma_i h(\delta_i^Tx + \phi_i)$, where $h(x)$ is the network's activation function. \\
We use three types of activation functions:
\begin{enumerate}
\item Sigmoid: \hskip 4.9em  $h(z)= (1+exp(-z))^{-1} $.
\item Softplus: \hskip 4.9em $ h(z)= \ln(1+exp(z))$.
\item Hyperbolic tangent: $h(z)=\tanh(z)$.
\end{enumerate}
\item[Notation] \hfill

In all tables, ``Nodes'' denotes the number of neurons in the hidden layer, ``Type'' stands for the activation type, ``MSE'' for the mean squared error over the training set, ``$\beta$'' for the blending coefficient given by eq. (\ref{BETAK}), ``AG. MSE'' for the mean squared error of the aggregate over the training set, ``AG. MSE/TE'' for the mean squared error of the aggregate  over the test set, and  ``$a$''  for the member participation coefficient, given by eq. (\ref {BLCOEF}).
\item[Randomness] \hfill

To ease the reproduction of the numerical results, we set all random seeds to 12345 using Python's Numpy.
\item[Code] \hfill

We made the code publicly available via the Github repository with URL: \url{https://github.com/salarsk1/Ensemble-NN}, under an MIT license.
\end{description}

\subsection{Test Case 1}
\begin{figure}[h!]
\centering
\begin{subfigure}{0.33\textwidth}
    \centering
    \includegraphics[width=\textwidth]{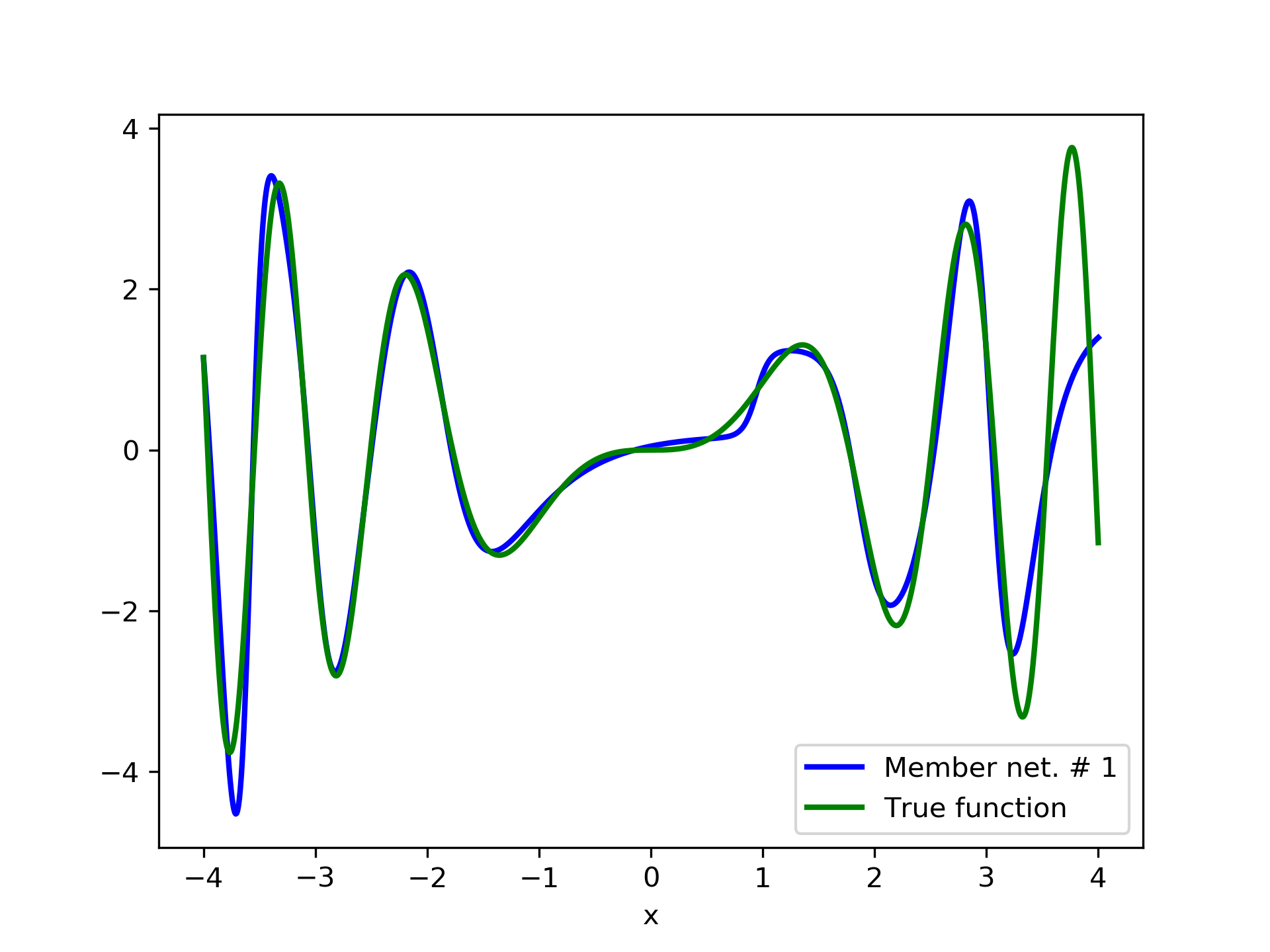}
\end{subfigure}\hfill
\begin{subfigure}{0.33\textwidth}
    \centering
    \includegraphics[width=\textwidth]{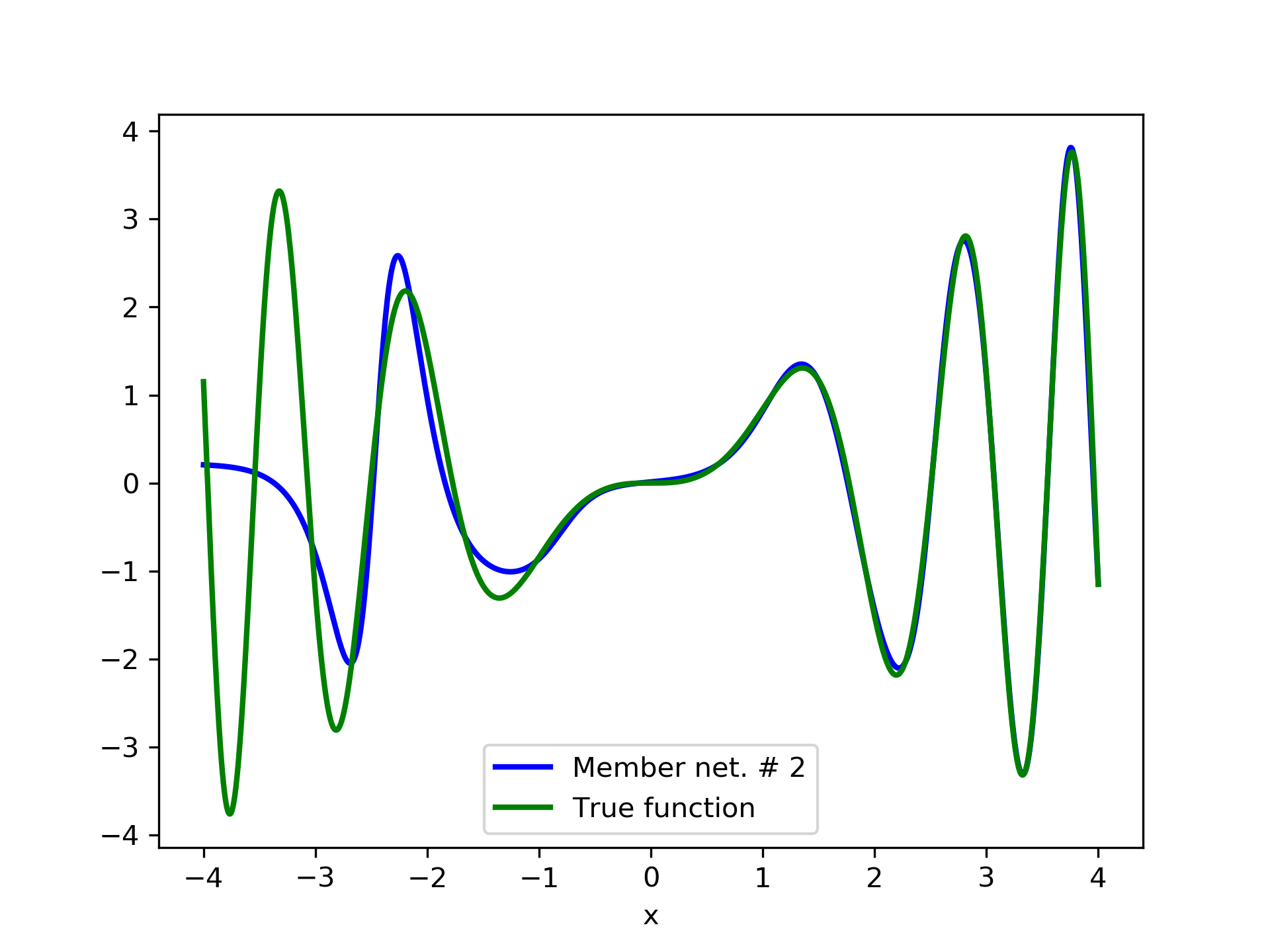}
\end{subfigure}\hfill
\begin{subfigure}{0.33\textwidth}
    \centering
    \includegraphics[width=\textwidth]{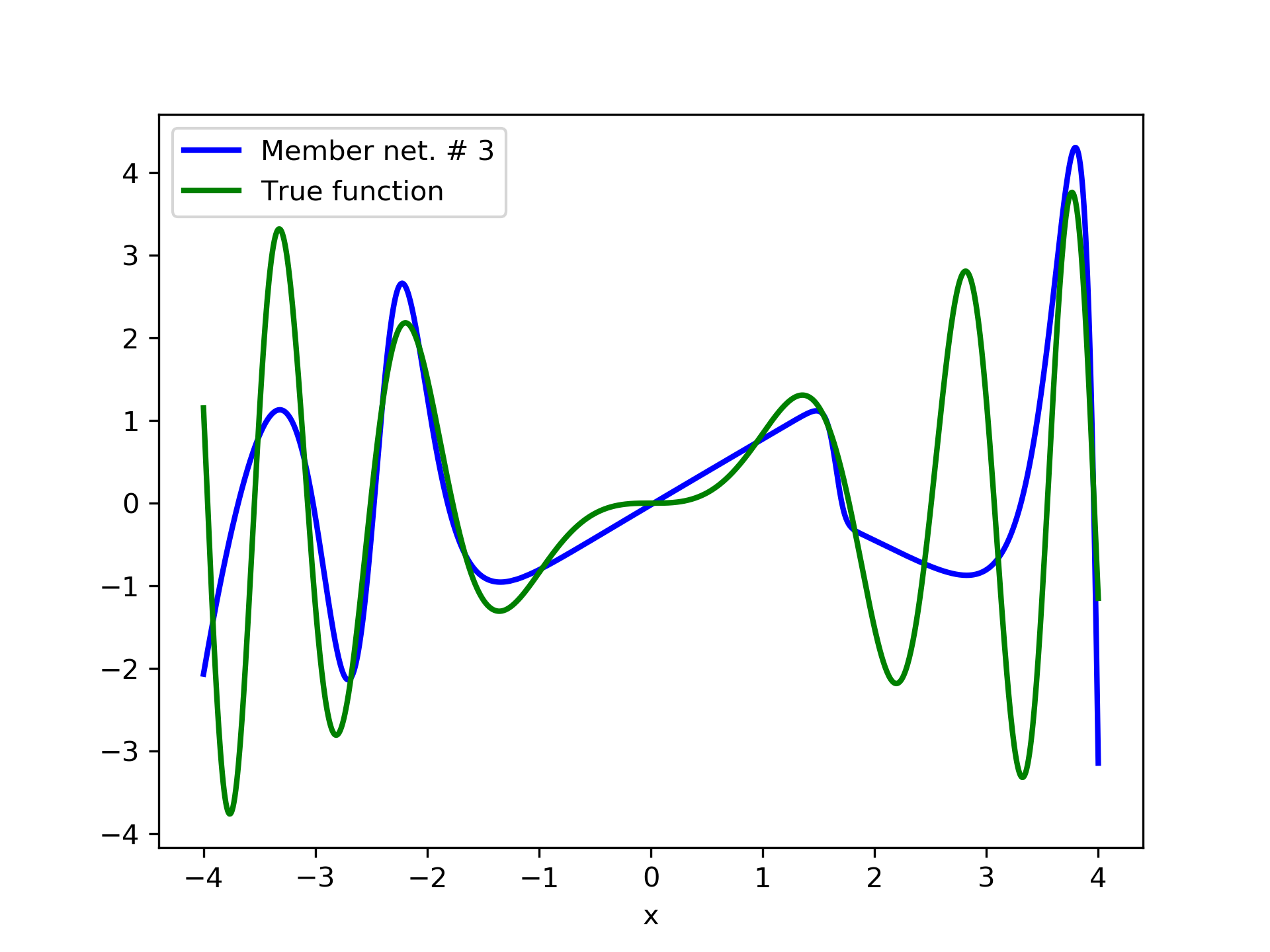}
\end{subfigure}
\begin{subfigure}{0.33\textwidth}
    \centering
    \includegraphics[width=\textwidth]{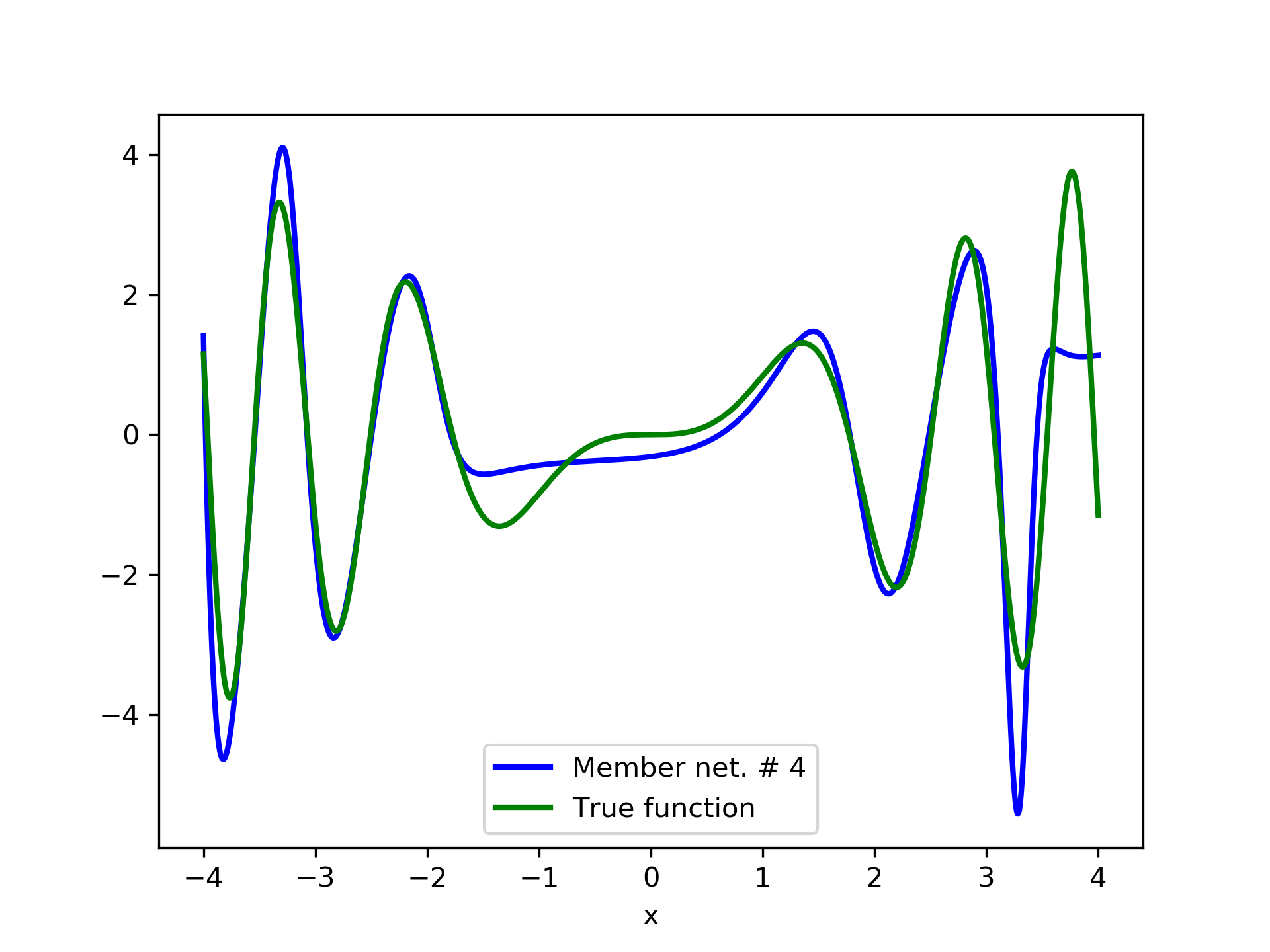}
\end{subfigure}\hfill
\begin{subfigure}{0.33\textwidth}
    \centering
    \includegraphics[width=\textwidth]{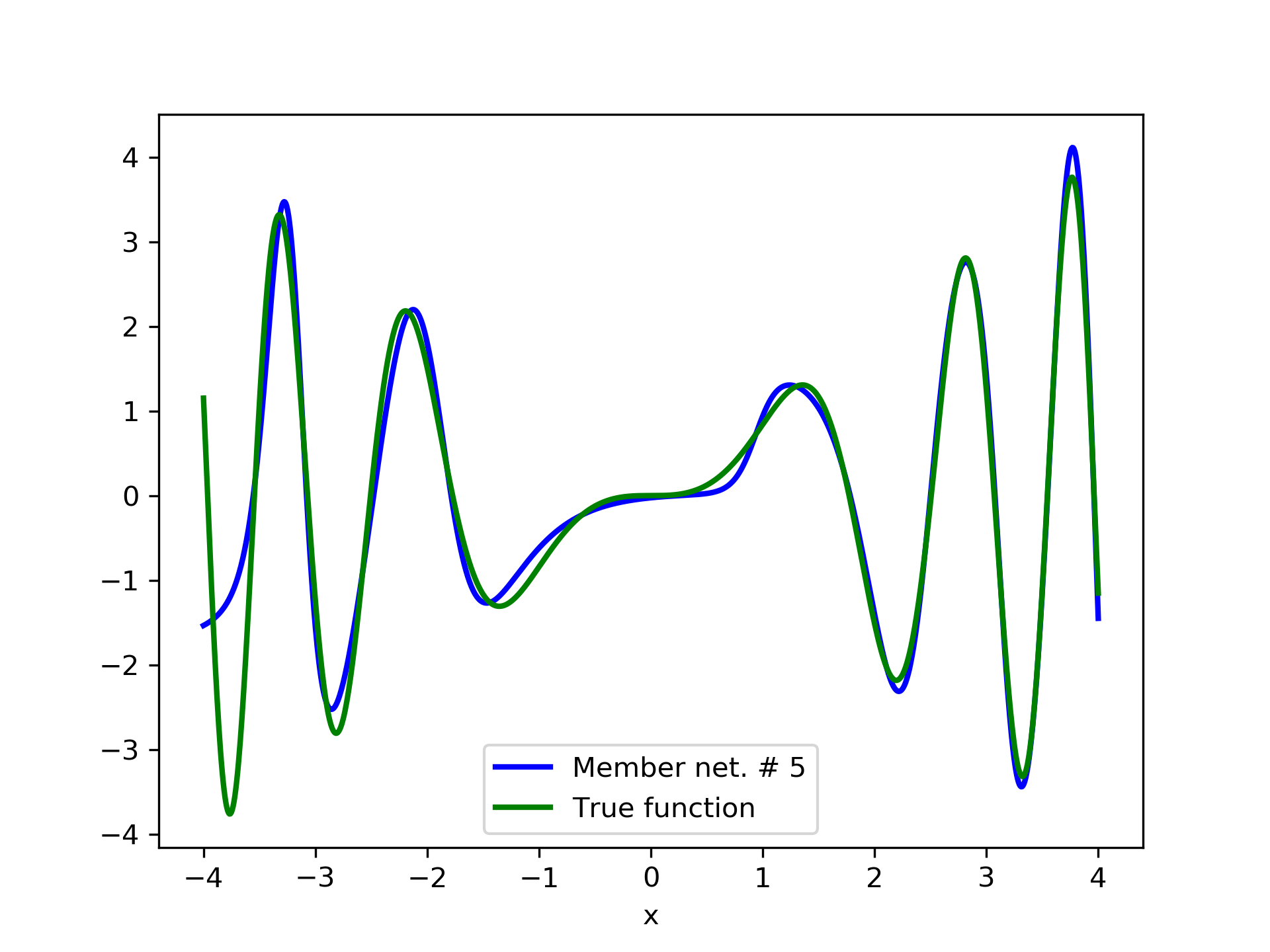}
\end{subfigure}\hfill
\begin{subfigure}{0.33\textwidth}
    \centering
    \includegraphics[width=\textwidth]{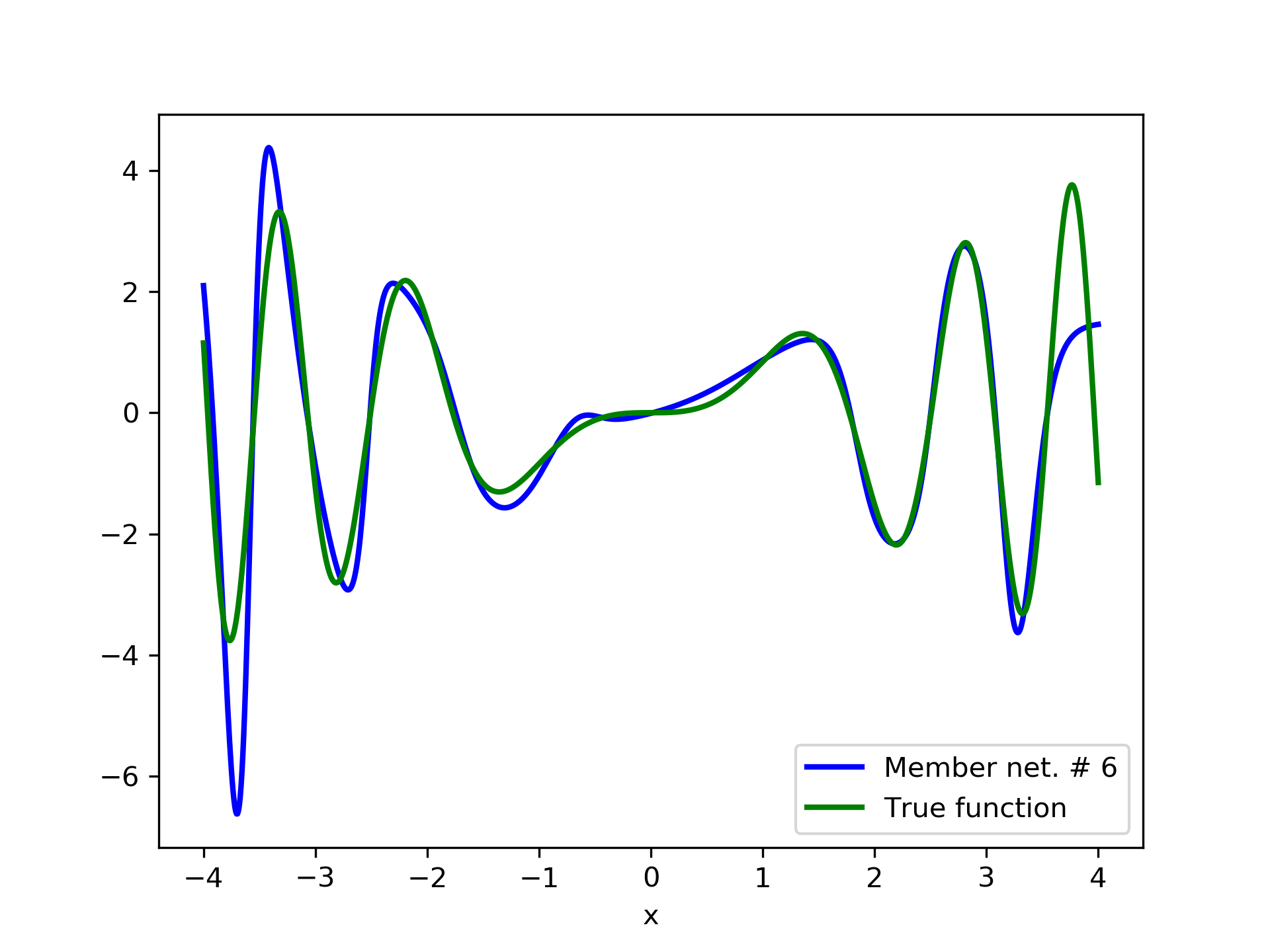}
\end{subfigure}
\caption{Plot of the ANN members related to table \ref{EXPF1},  along with the plot of $f_1(x)$.}
\label{MEMBERS}
\end{figure}

\begin{figure}[h!]
\centering
\begin{subfigure}{0.33\textwidth}
    \centering
    \includegraphics[width=\textwidth]{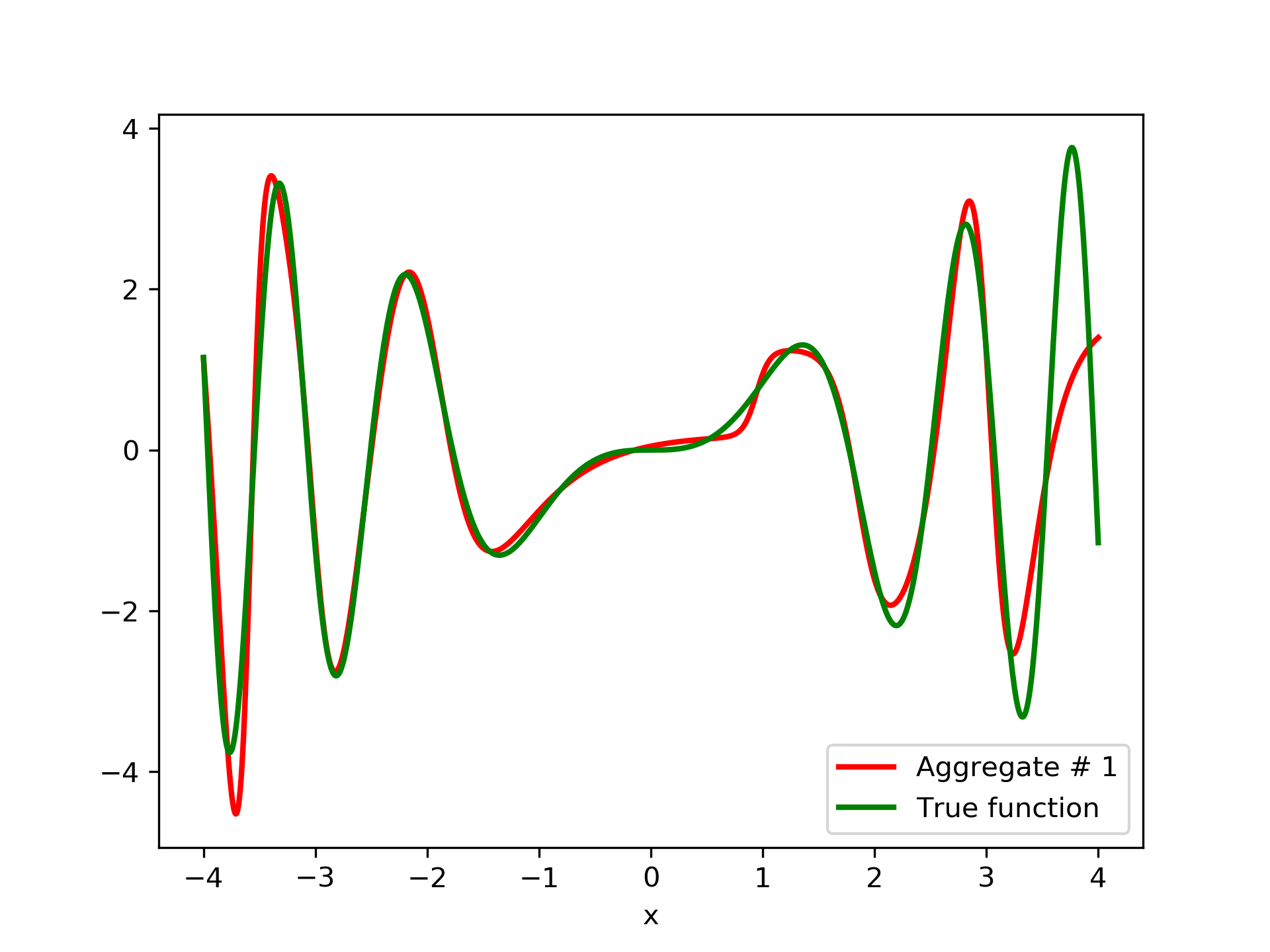}
\end{subfigure}\hfill
\begin{subfigure}{0.33\textwidth}
    \centering
    \includegraphics[width=\textwidth]{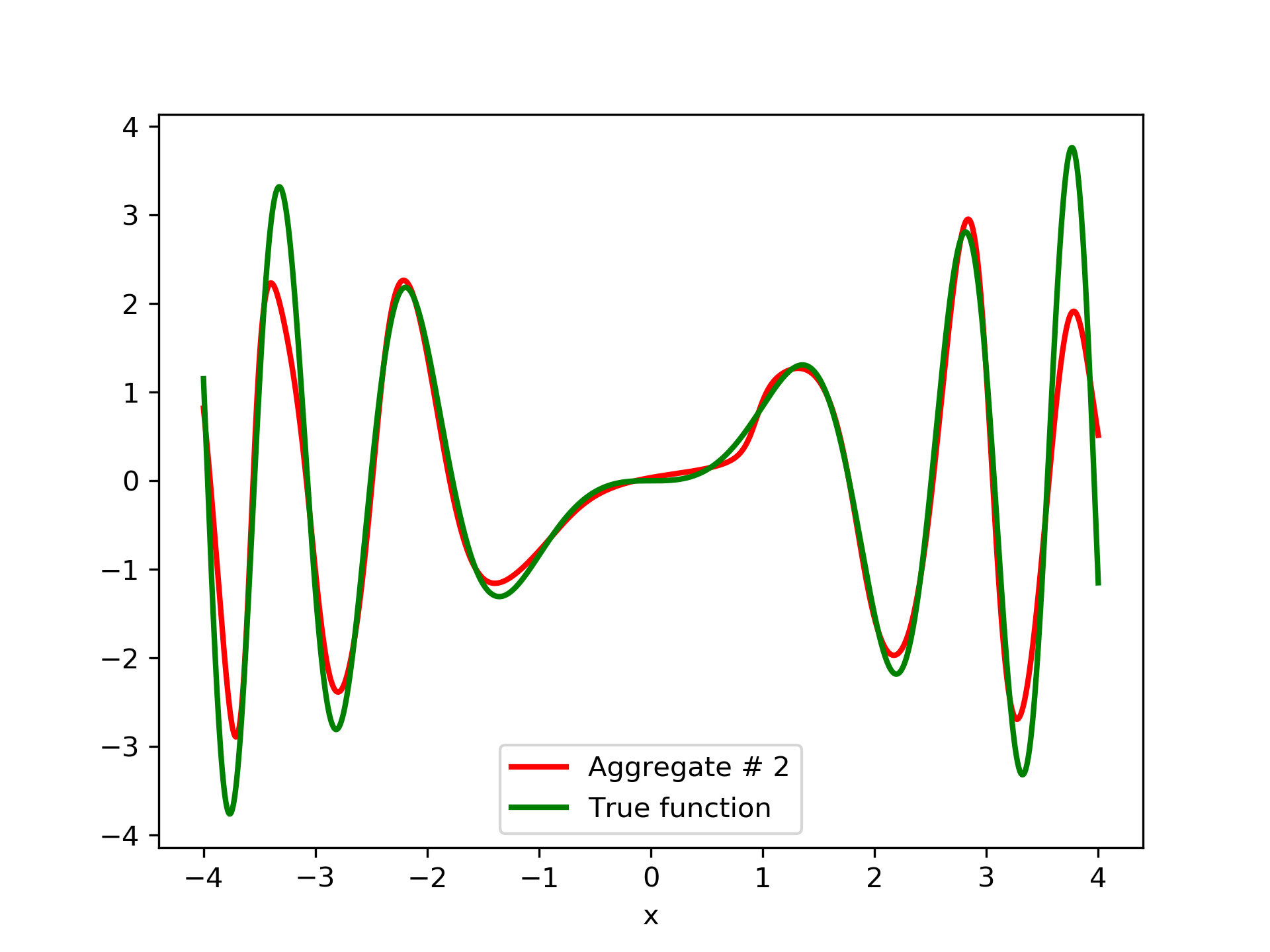}
\end{subfigure}\hfill
\begin{subfigure}{0.33\textwidth}
    \centering
    \includegraphics[width=\textwidth]{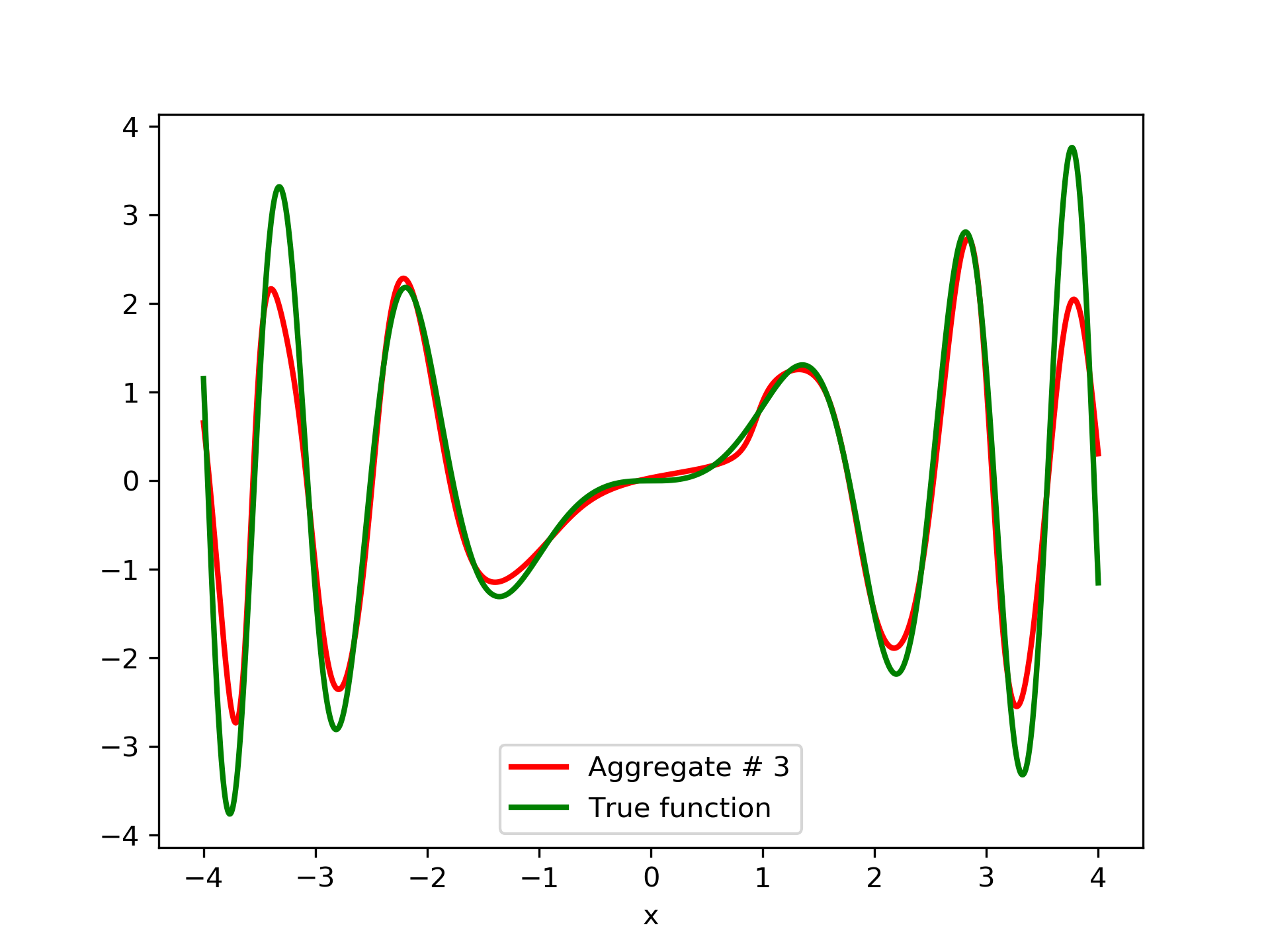}
\end{subfigure}
\begin{subfigure}{0.33\textwidth}
    \centering
    \includegraphics[width=\textwidth]{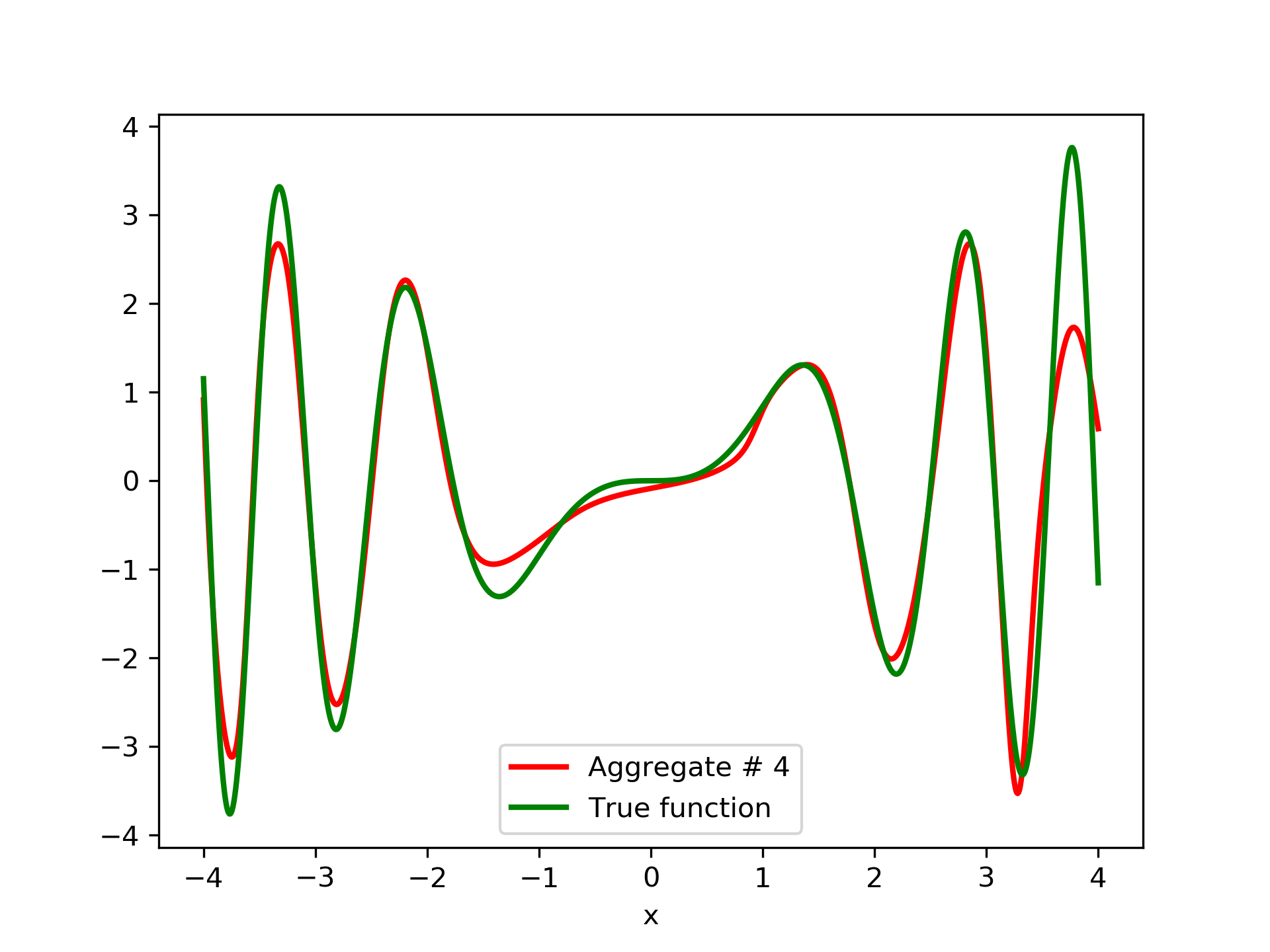}
\end{subfigure}\hfill
\begin{subfigure}{0.33\textwidth}
    \centering
    \includegraphics[width=\textwidth]{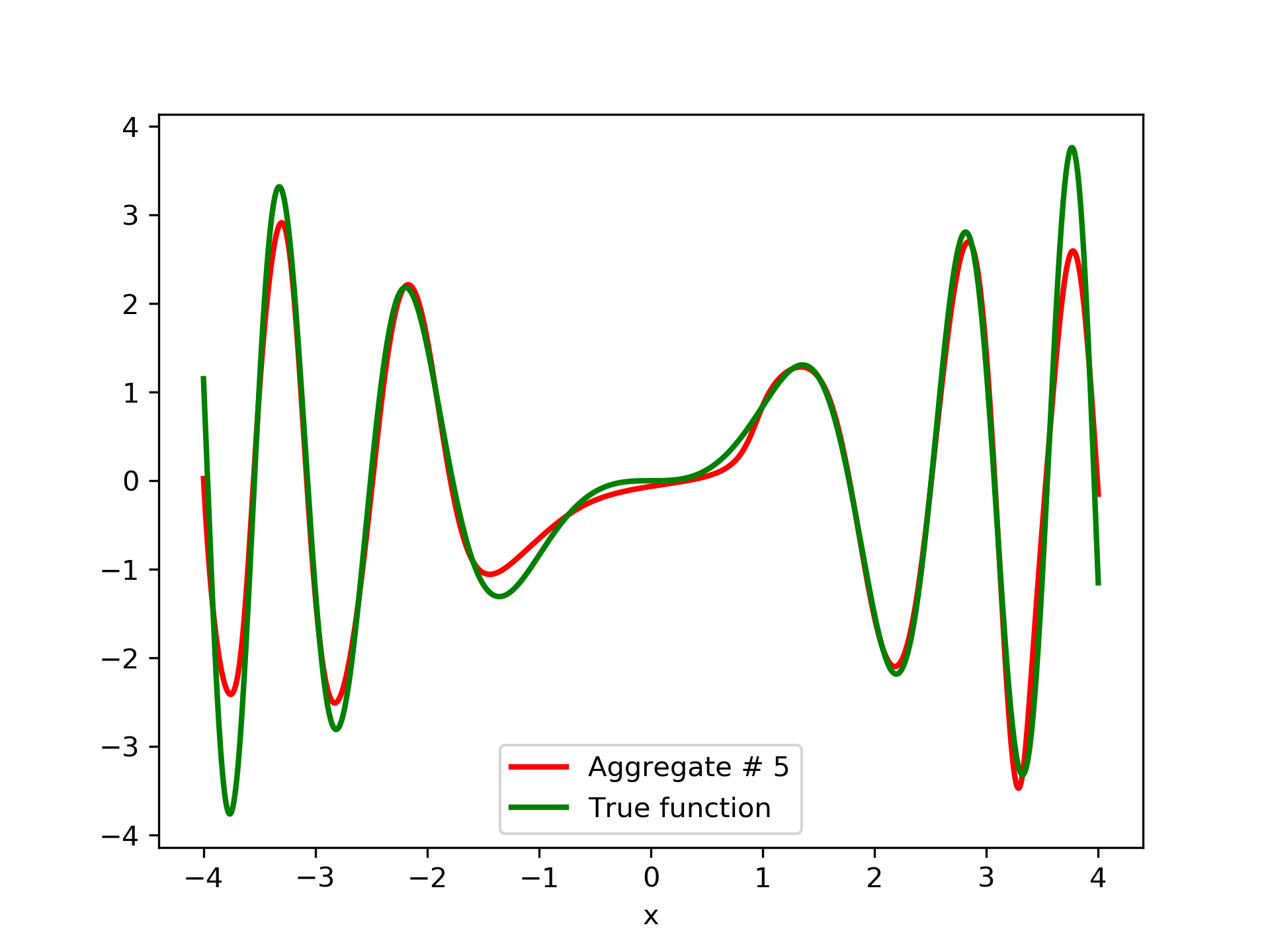}
\end{subfigure}\hfill
\begin{subfigure}{0.33\textwidth}
    \centering
    \includegraphics[width=\textwidth]{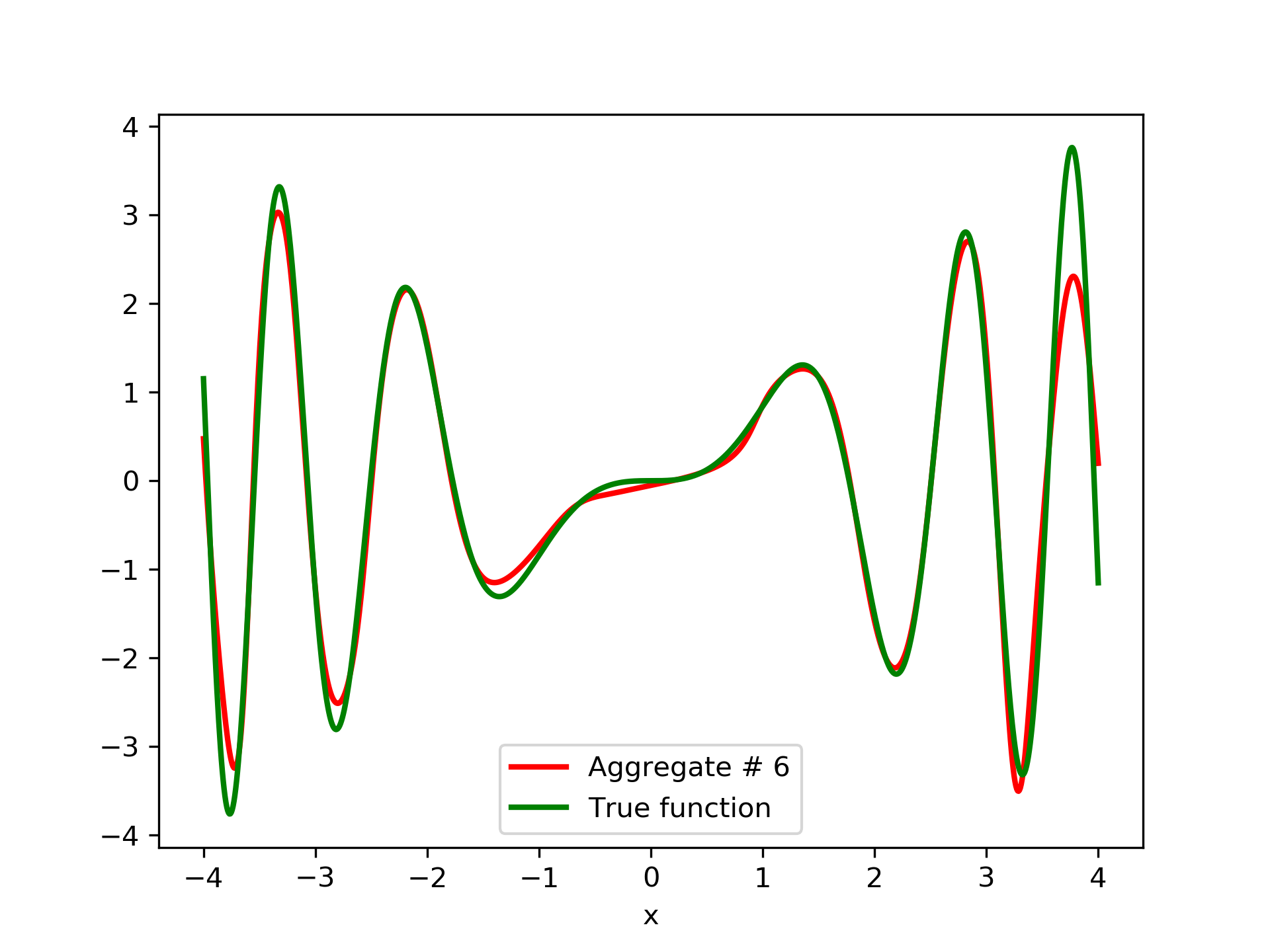}
\end{subfigure}
\caption{Plot of the aggregate ANNs related to table \ref{EXPF1},  along with the plot of $f_1(x)$.}
\label{AGGREGATES}
\end{figure}

Model function: $f_1(x), x \in [-4, 4]$. 
We construct a set $D$ containing $M_D = 1000$ equidistant points, $x_i = -4+(i-1)\frac{8}{999}$. The training set $T_r$ contains $M_r = 38$ equidistant points,
$ z_j = -4+(j-1)\frac{216}{999}=x_{27j-26}$. Note that $T_r \subset D$ and the test set is $T_s = D - T_r$.
The ensemble contained six networks.  
Details of this experiment are listed in table \ref{EXPF1}. 
Among the 6 models, model No. 5 has the minimum training MSE, which is 0.38498. The training MSE of the final aggregate is 0.14035, which is $63\%$ lower than that of model No. 5.
The plots of the six member-networks $N_i(x)$ together with the target function $f_1(x)$ are depicted in fig. \ref{MEMBERS}. Likewise, plots of the corresponding aggregate networks are depicted in \mbox{fig. \ref{AGGREGATES}.}
\vskip -0.2cm
 \begin{table}[h!]
\caption{Experiment with $f_1(x)$ with $x\in[-4,4]$ and a training set of 38 equidistant  points. Regularization factor: $\nu_{reg}=0.002$.}
\begin{center}
 \begin{tabular}{|c|c|c|c|c|c|c|c|}
\hline
     ANN & Nodes & Type & MSE & $\beta$ &  AG. MSE &  AG. MSE/TE &   $a$ \\
\hline
 1 &      9 &      tanh &  0.45326 &  0.00000 &  0.45326 &     0.32955 &  0.20175 \\
\hline
   2 &     11 &   sigmoid &  0.85077 &  0.65221 &  0.29531 &     0.21960 &  0.10759 \\
\hline
   3 &     11 &  softplus &  1.92921 &  0.94300 &  0.28932 &     0.22312 &  0.01870 \\
\hline
   4 &      9 &      tanh &  0.43278 &  0.65750 &  0.23590 &     0.16807 &  0.17088 \\
\hline
   5 &     11 &   sigmoid &  0.38498 &  0.63707 &  0.16427 &     0.10779 &  0.28423 \\
\hline
   6 &     12 &   sigmoid &  0.45224 &  0.78315 &  0.14035 &     0.08970 &  0.21685 \\
\hline
\end{tabular}
\end{center}
\label{EXPF1}
\end{table}

\subsection{Test Case 2}
\begin{figure}
\begin{center}
\includegraphics[scale=.8]{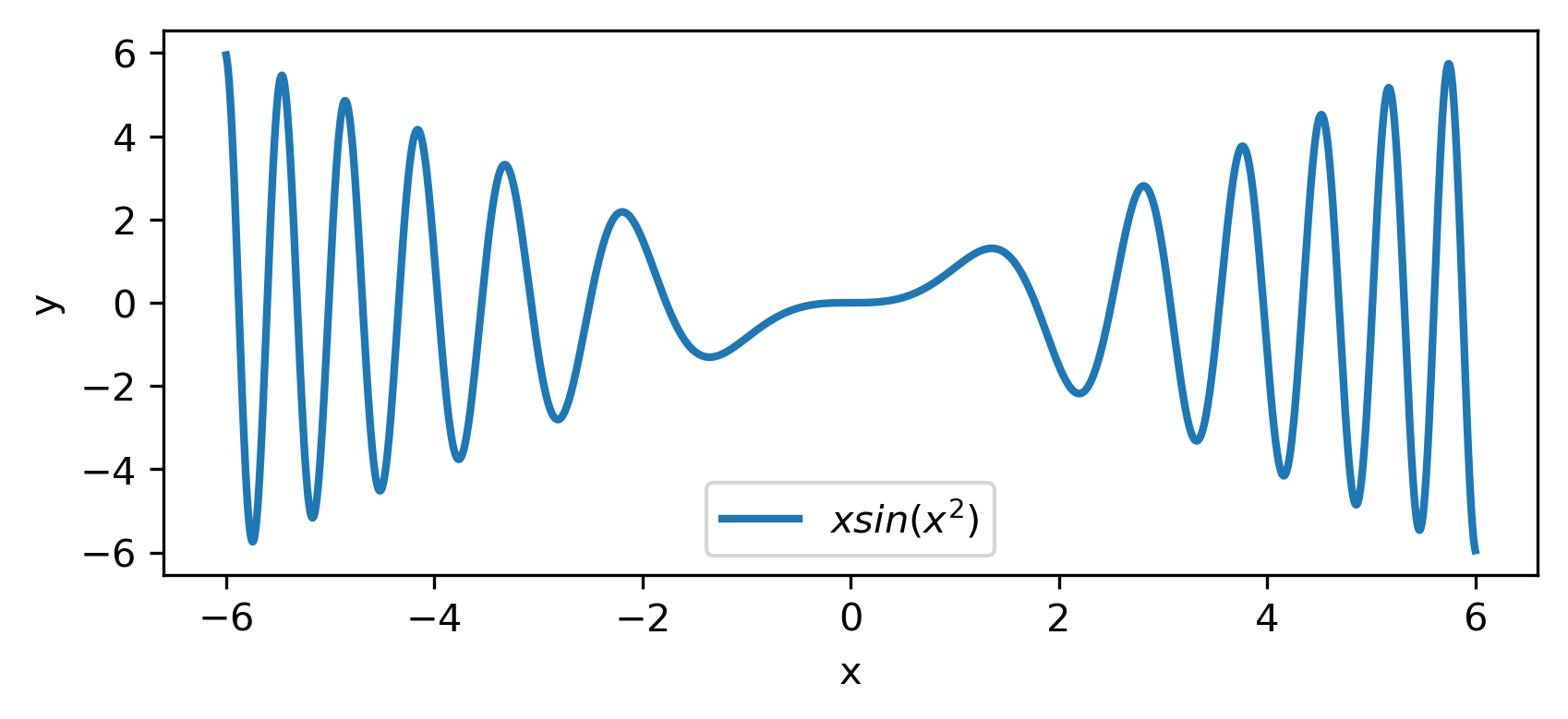}
\end{center}
\caption{Plot of $f_1(x)$ with $x\in[-6,6]$, illustrating the rapidly oscillating pattern as $|x|$ grows}
\label{F1W}
\end{figure}

To escalate the difficulty of the task, in this case we have experimented with $f_1(x)$ in a wider range $x\in[-6,6]$, which introduces intense oscillatory behavior illustrated in \mbox{fig. \ref{F1W}.}
We construct a set $D$ containing $M_D = 1201$ equidistant points, $x_i = -6+\frac{i-1}{100}$. The training set $T_r$ contains $M_r = 121$ equidistant points,
$ z_j = -6+\frac{j-1}{10}=x_{10j-9}$. Note that $T_r \subset D$ and the test set is $T_s = D - T_r$.
The generalization performance is quite satisfactory but not of the quality of the first example, as can be seen by inspecting  \mbox{table \ref{EXPF1W}.}
This was expected since the rapid oscillations as $|x|$ grows, render the task harder.
\begin{table}[ht]
\caption{Experiment with $f_1(x)$ with $x\in[-6,6]$ and a training set of 121 equidistant  points. Regularization factor: $\nu_{reg}=0.003$.}
\begin{center}
\begin{tabular}{|c|c|c|c|c|c|c|c|}
\hline
     ANN & Nodes & Type & MSE &     $\beta$ &  AG. MSE &  AG. MSE/TE & $a$ \\
\hline
 1 &     23 &      tanh &   2.87200 &  0.00000 &  2.87200 &     2.73192 &  0.04995 \\
 \hline
   2 &     25 &   sigmoid &  12.93168 &  0.86162 &  2.60566 &     2.62689 &  0.00802 \\
 \hline
   3 &     27 &   sigmoid &   3.19976 &  0.55508 &  1.53797 &     1.60549 &  0.04646 \\
   \hline
   4 &     23 &  softplus &  18.90312 &  0.95861 &  1.50554 &     1.56396 &  0.00451 \\
   \hline
   5 &     24 &   sigmoid &   5.05116 &  0.79406 &  1.24986 &     1.30413 &  0.02825 \\
   \hline
   6 &     29 &      tanh &   4.18786 &  0.81902 &  1.09904 &     1.13224 &  0.03032 \\
   \hline
   7 &     26 &      tanh &   4.07782 &  0.83087 &  0.97027 &     1.00334 &  0.03410 \\
   \hline
   8 &     23 &   sigmoid &   4.48691 &  0.88752 &  0.91286 &     0.93800 &  0.02555 \\
   \hline
   9 &     24 &      tanh &   3.70554 &  0.86650 &  0.84497 &     0.84990 &  0.03500 \\
   \hline
  10 &     25 &      tanh &   2.00931 &  0.73126 &  0.66317 &     0.68001 &  0.09634 \\
  \hline
  11 &     28 &   sigmoid &   0.98583 &  0.61088 &  0.44286 &     0.45987 &  0.22836 \\
  \hline
  12 &     27 &      tanh &   1.27605 &  0.78000 &  0.37085 &     0.38586 &  0.16552 \\
  \hline
  13 &     26 &  softplus &  10.20572 &  0.97406 &  0.36387 &     0.37846 &  0.02004 \\
  \hline
  14 &     26 &      tanh &   2.98190 &  0.95094 &  0.35688 &     0.37098 &  0.03985 \\
  \hline
  15 &     26 &      tanh &   1.12186 &  0.81227 &  0.31371 &     0.32395 &  0.18773 \\
 \hline
\end{tabular}
\end{center}
\label{EXPF1W}
\end{table}
We have noticed that if one of the member networks $N_i(x)$  overfits the data, then a number of additional networks are necessary in the ensemble in order to eliminate the overfit effect. 
So it seems that a preferred tactic would be to use small networks, that are less prone to over-fitting, rather than large ones. This also serves the need of the so called ``{\it few-shot learning}'', that refers to problems with limited training examples \citet{Wang:2020}.
\begin{figure}[h]
\centering
\begin{subfigure}{0.33\textwidth}
    \centering
    \includegraphics[width=\textwidth]{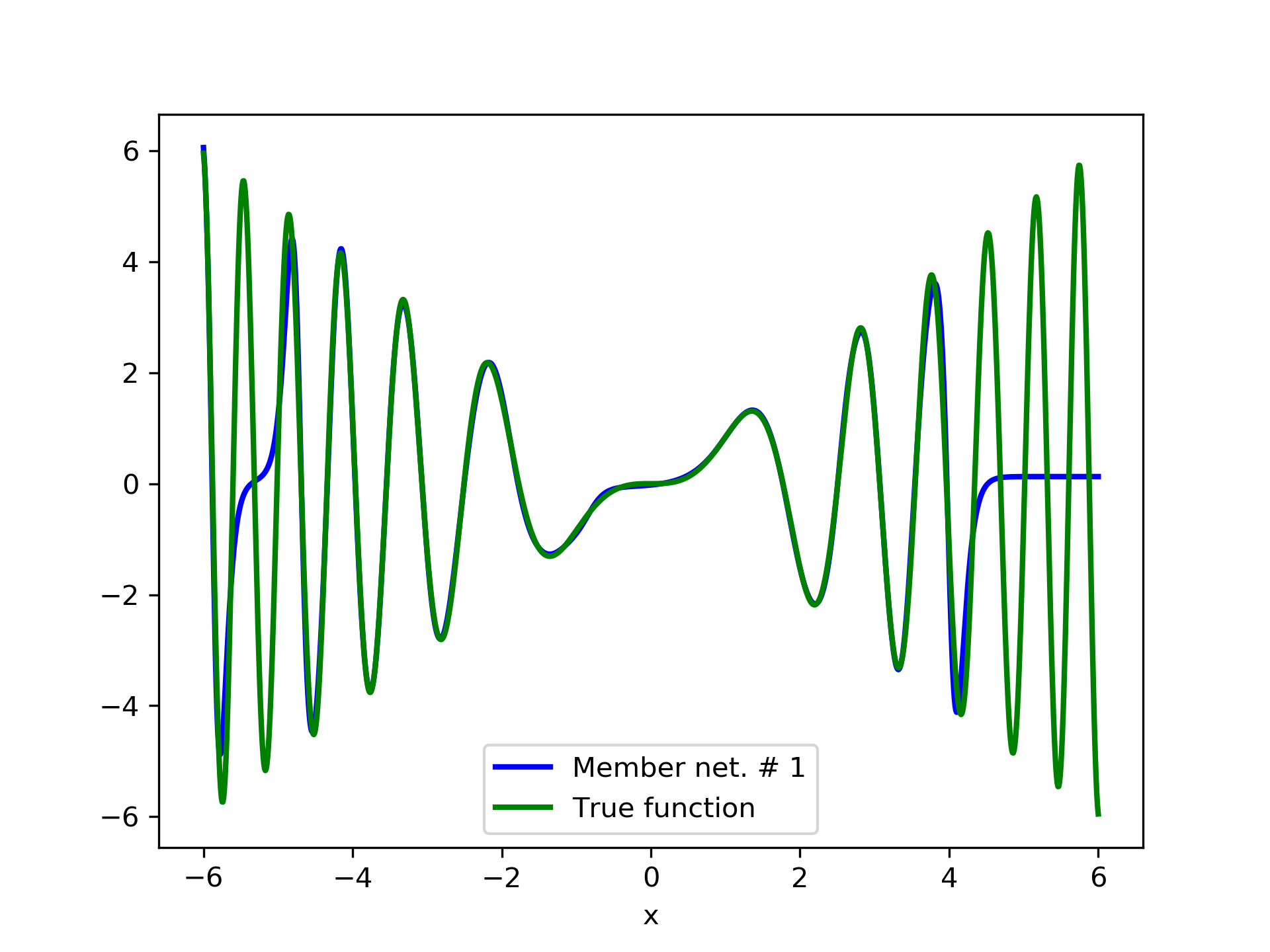}
\end{subfigure}\hfill
\begin{subfigure}{0.33\textwidth}
    \centering
    \includegraphics[width=\textwidth]{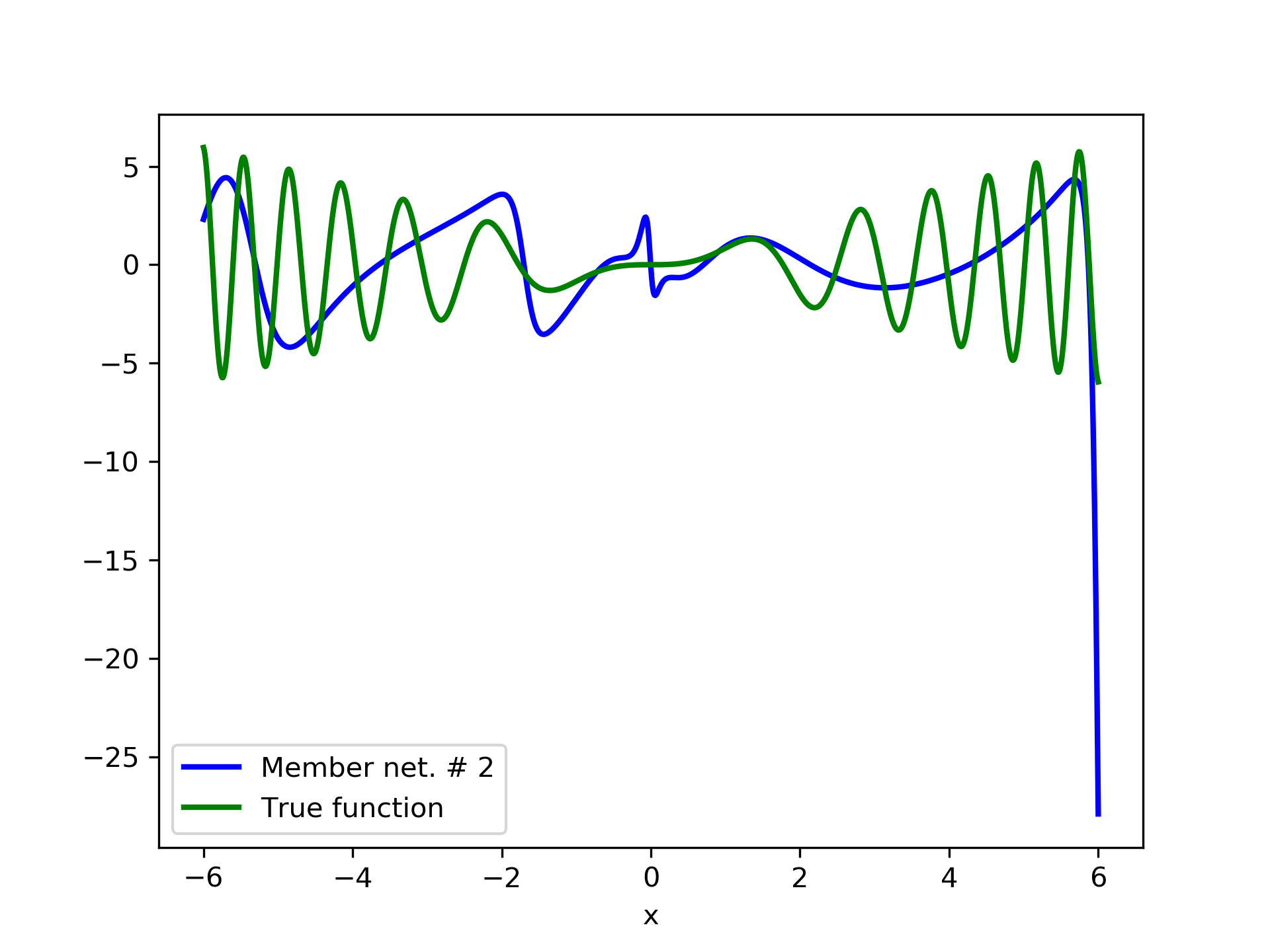}
\end{subfigure}\hfill
\begin{subfigure}{0.33\textwidth}
    \centering
    \includegraphics[width=\textwidth]{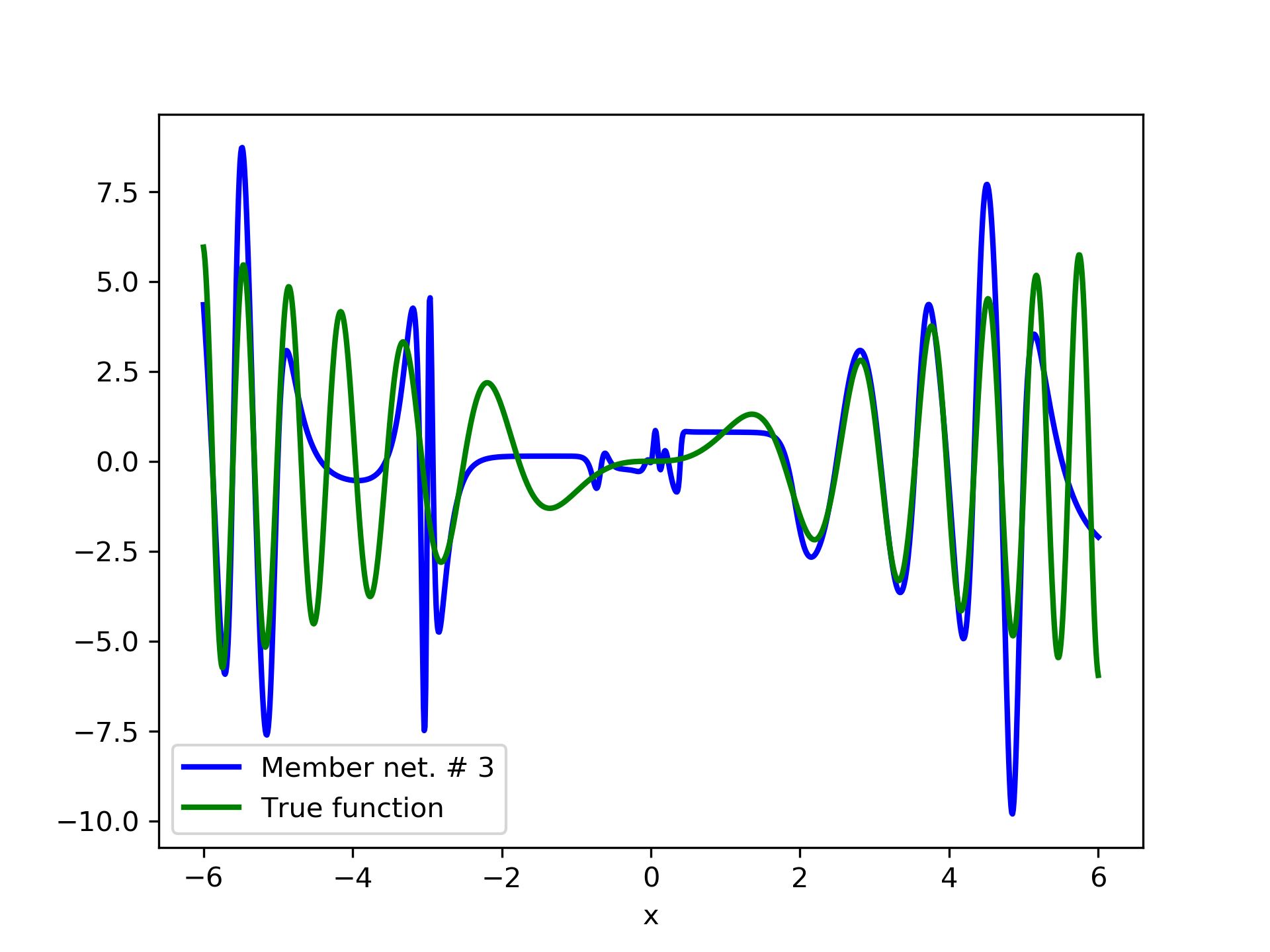}
\end{subfigure}
\begin{subfigure}{0.33\textwidth}
    \centering
    \includegraphics[width=\textwidth]{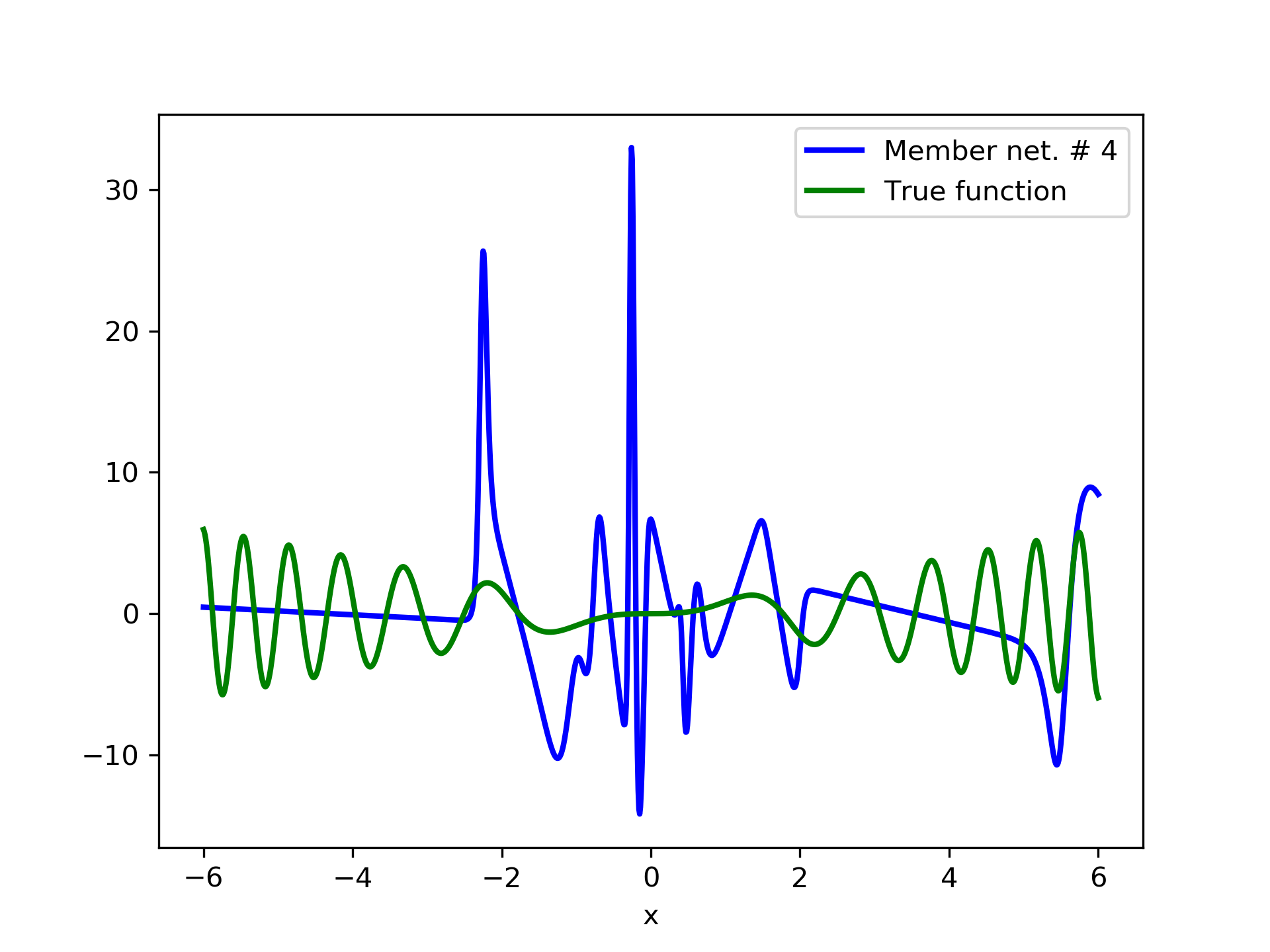}
\end{subfigure}\hfill
\begin{subfigure}{0.33\textwidth}
    \centering
    \includegraphics[width=\textwidth]{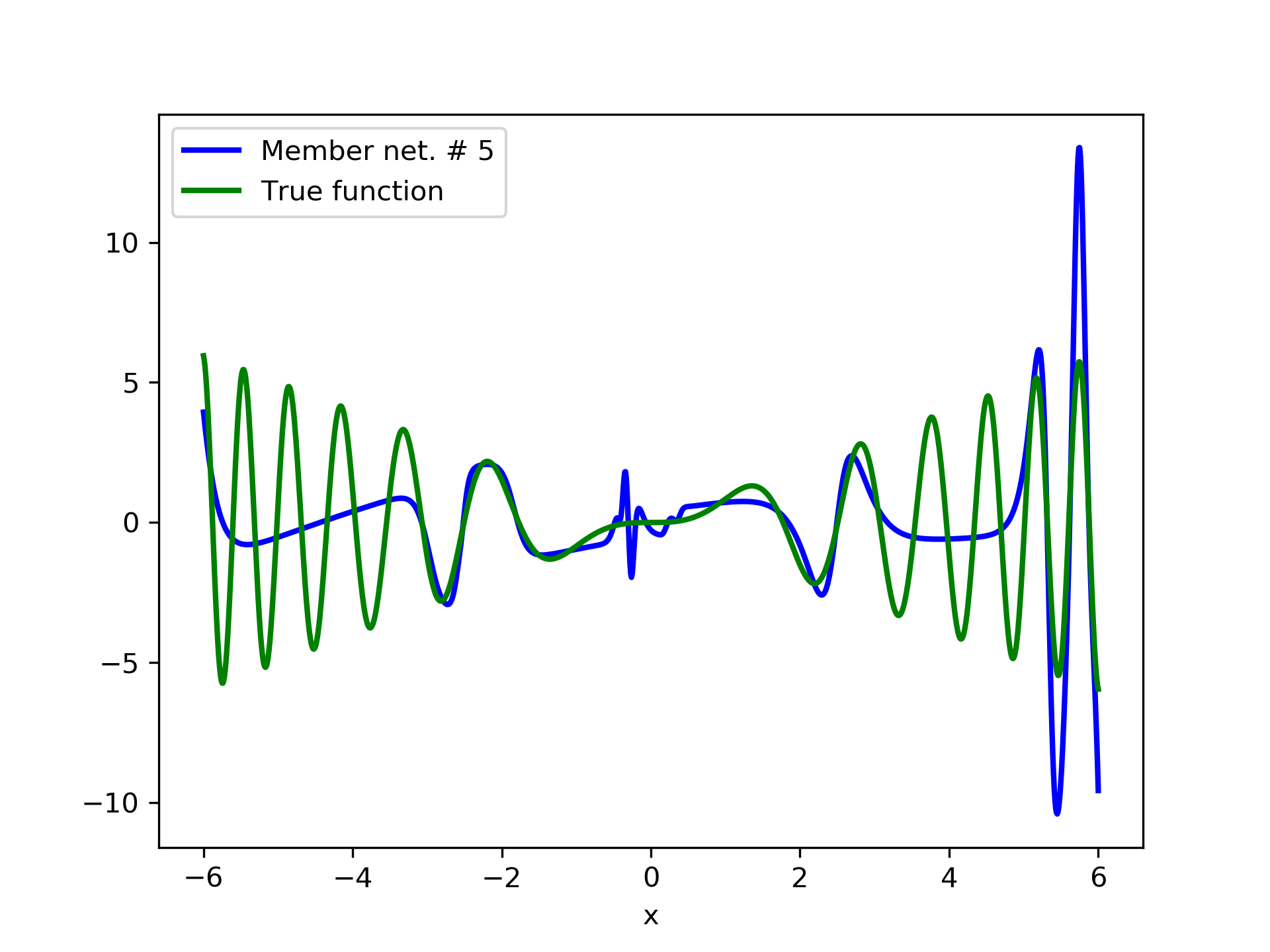}
\end{subfigure}\hfill
\begin{subfigure}{0.33\textwidth}
    \centering
    \includegraphics[width=\textwidth]{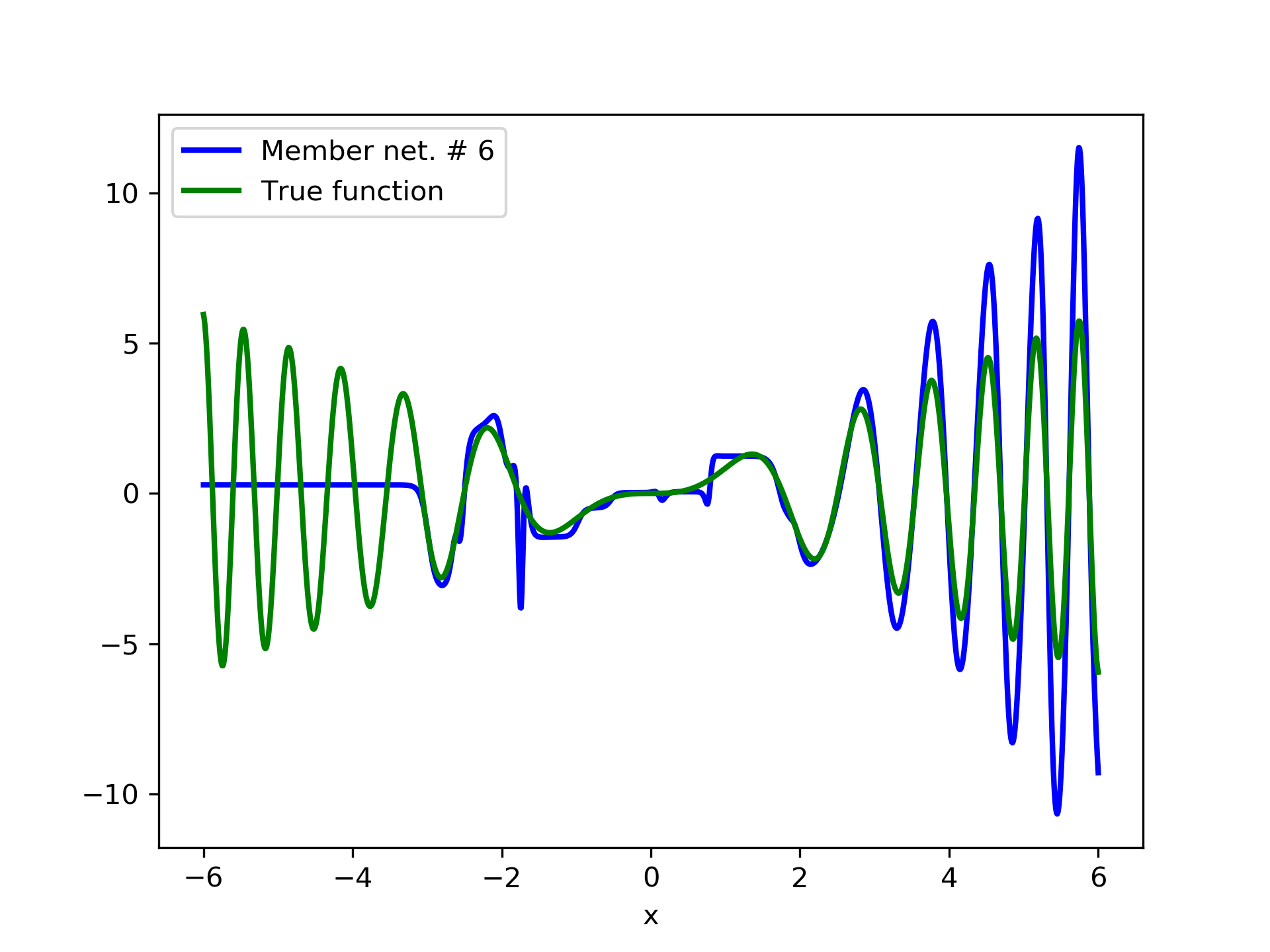}
\end{subfigure}
\begin{subfigure}{0.33\textwidth}
    \centering
    \includegraphics[width=\textwidth]{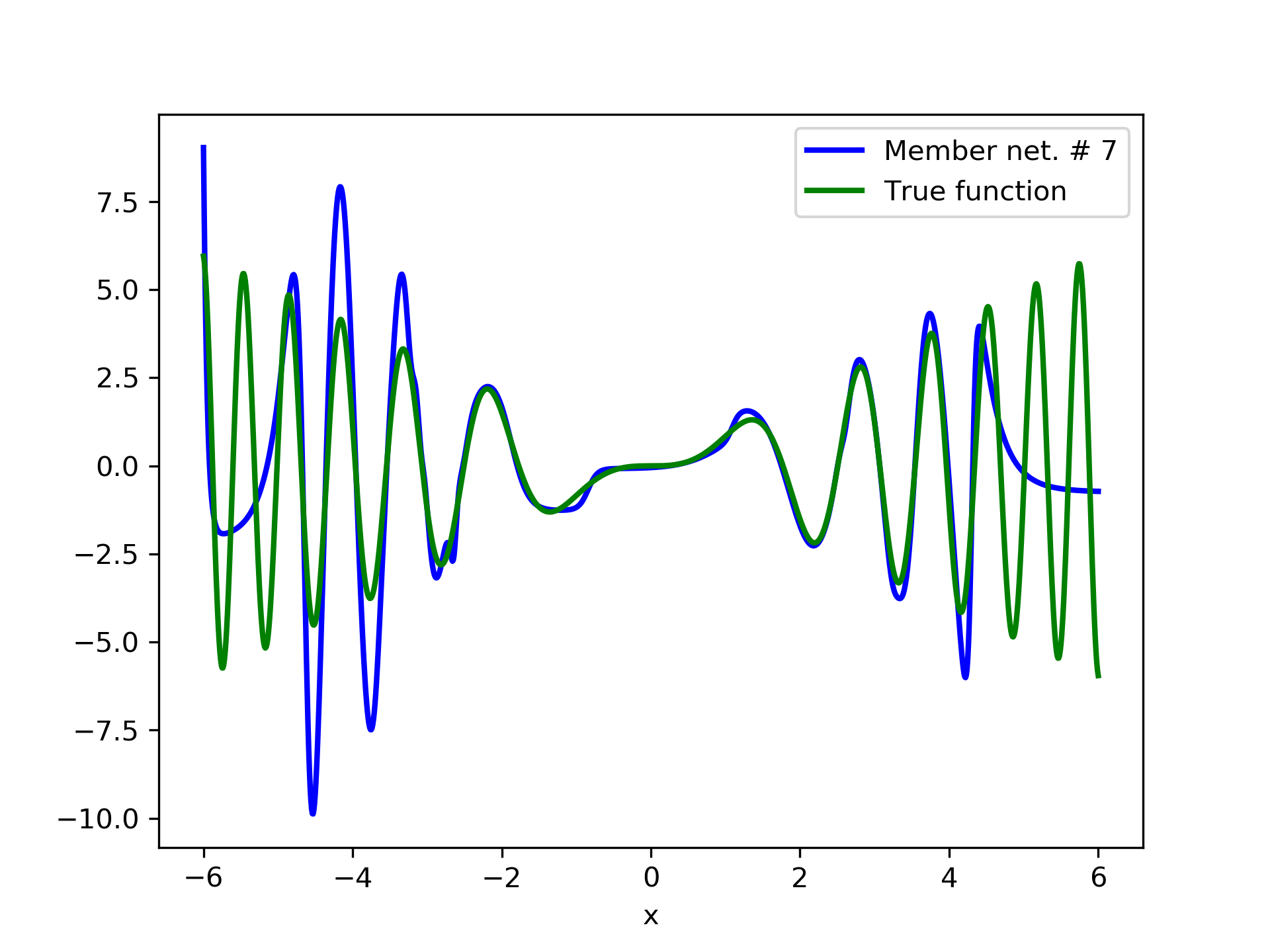}
\end{subfigure}\hfill
\begin{subfigure}{0.33\textwidth}
    \centering
    \includegraphics[width=\textwidth]{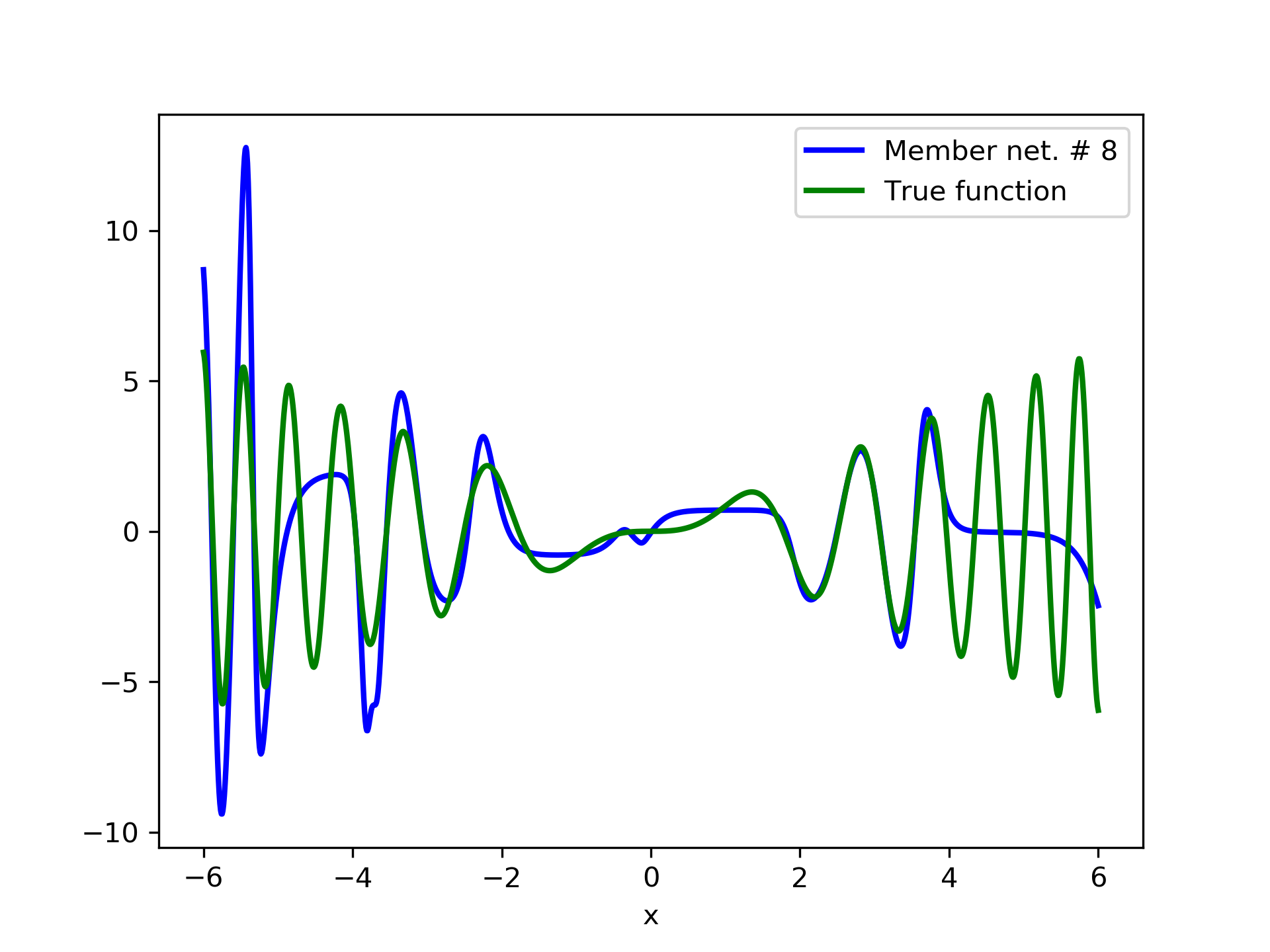}
\end{subfigure}\hfill
\begin{subfigure}{0.33\textwidth}
    \centering
    \includegraphics[width=\textwidth]{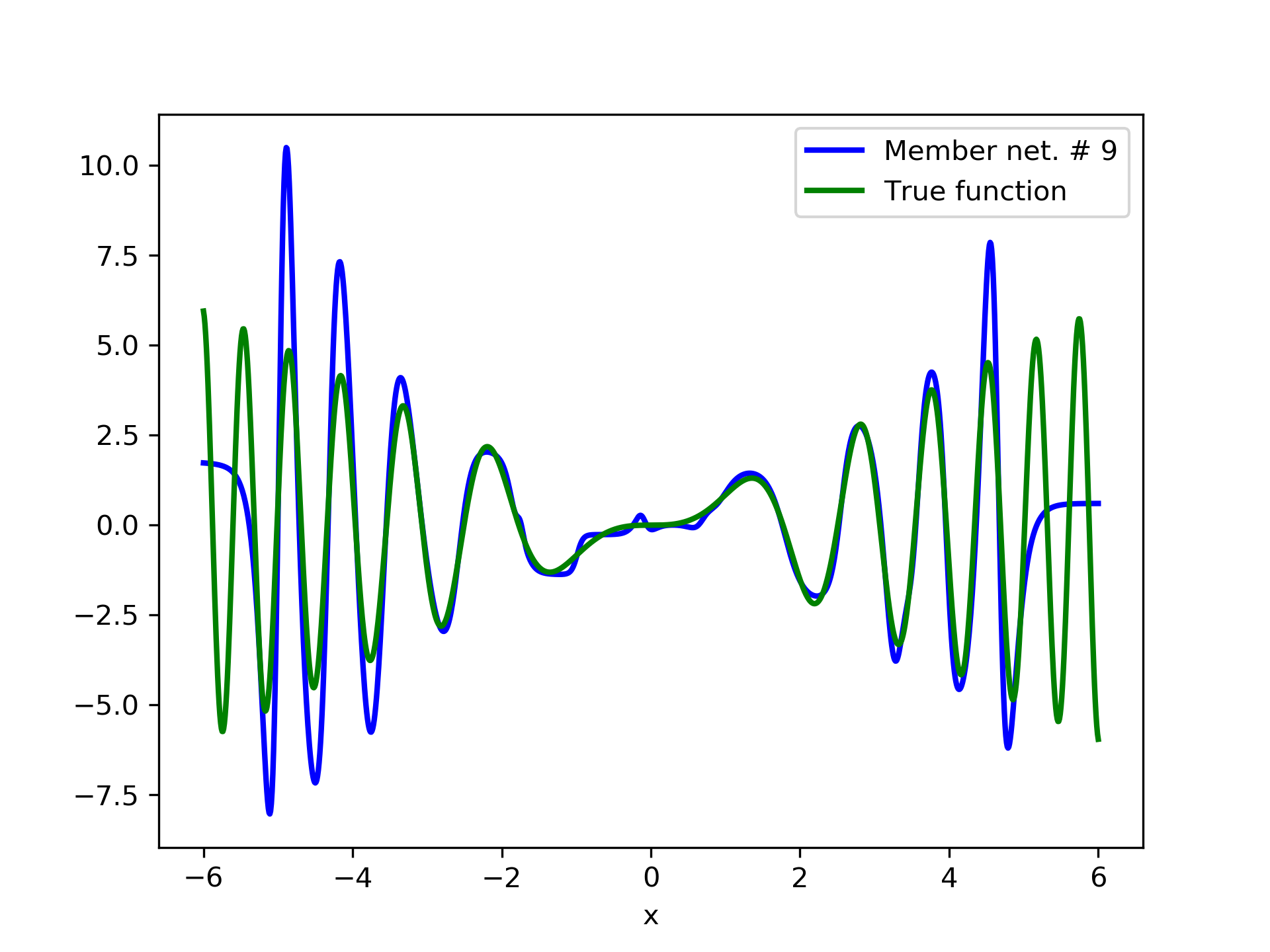}
\end{subfigure}
\begin{subfigure}{0.33\textwidth}
    \centering
    \includegraphics[width=\textwidth]{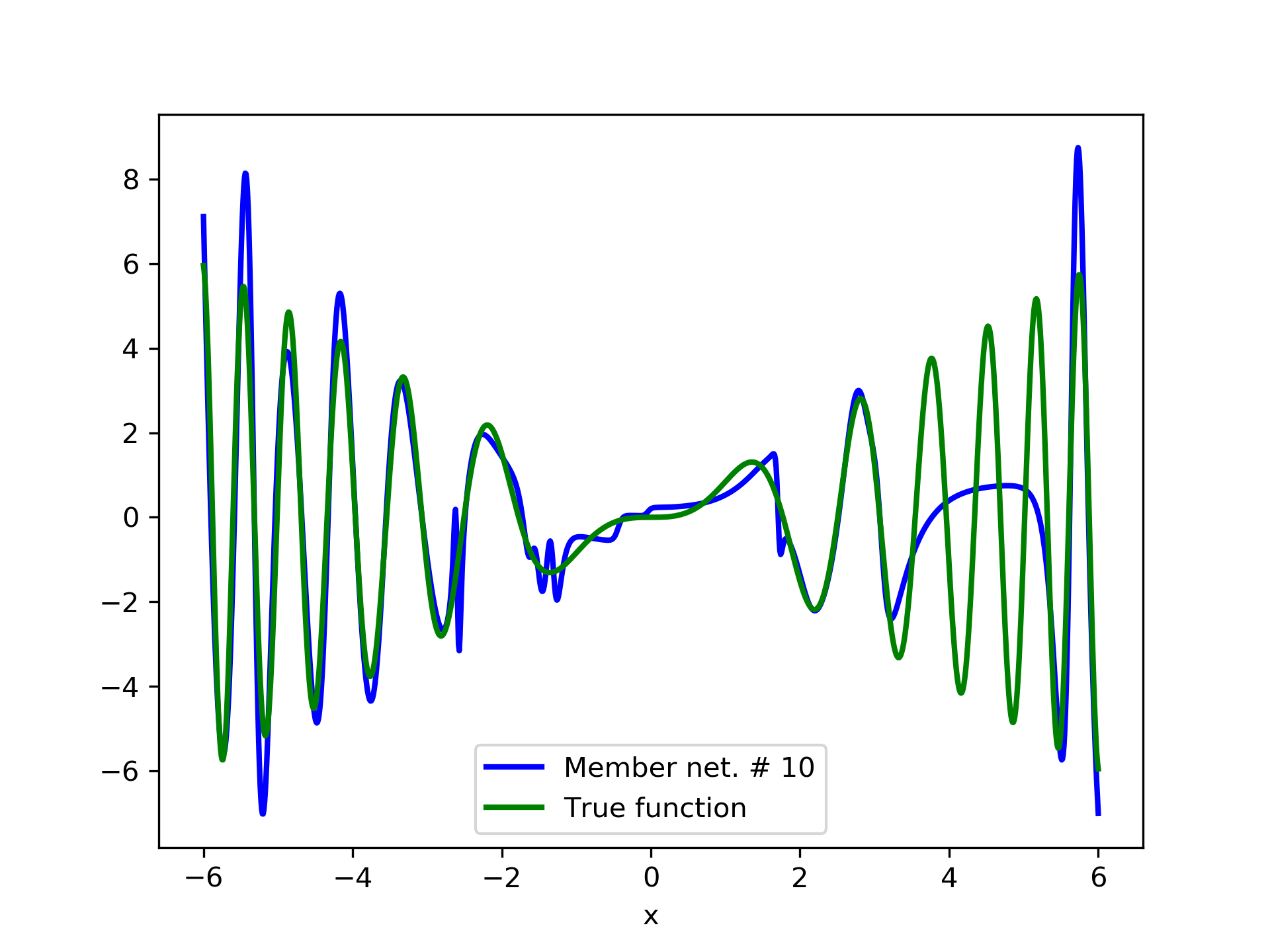}
\end{subfigure}\hfill
\begin{subfigure}{0.33\textwidth}
    \centering
    \includegraphics[width=\textwidth]{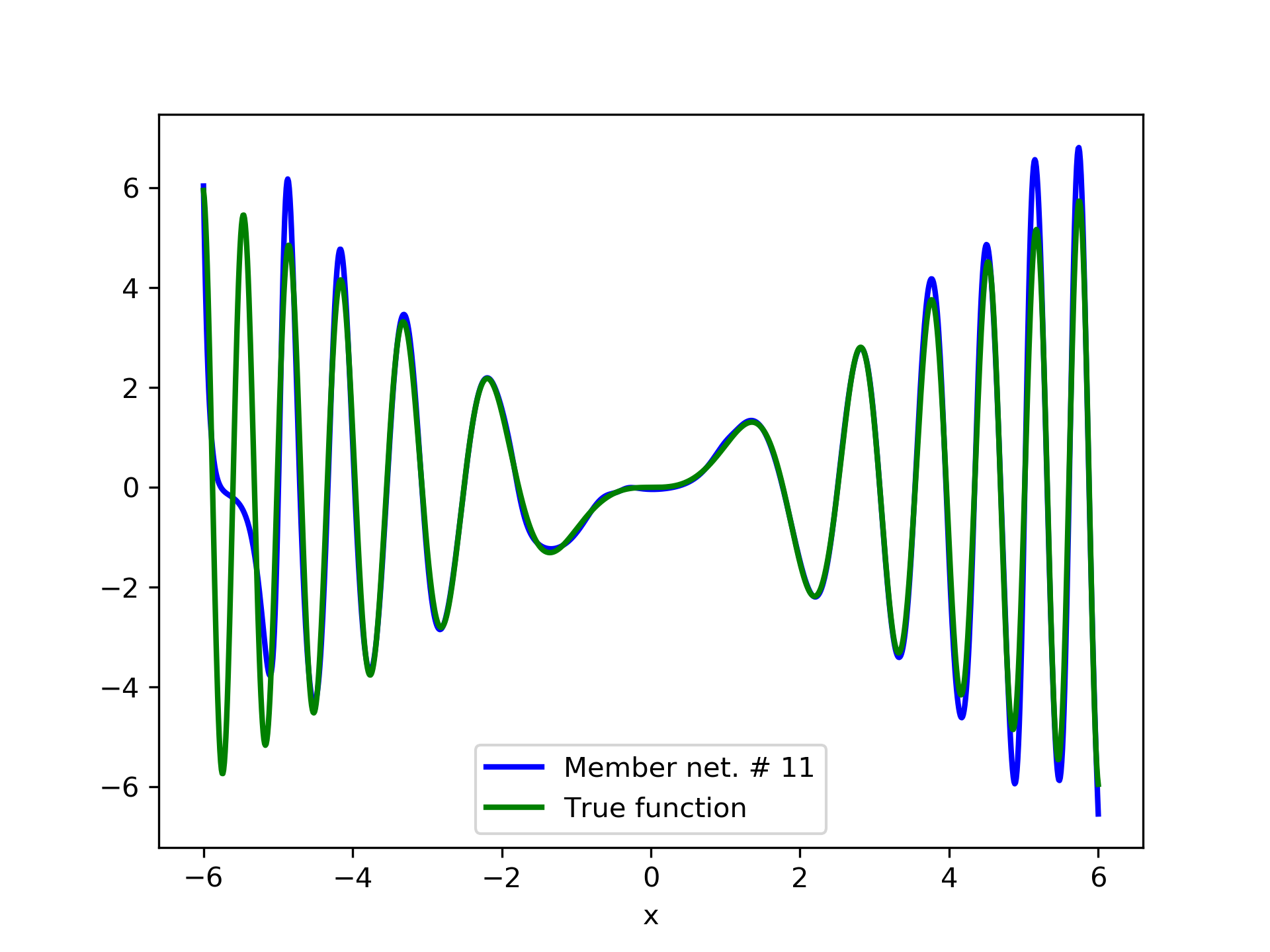}
\end{subfigure}\hfill
\begin{subfigure}{0.33\textwidth}
    \centering
    \includegraphics[width=\textwidth]{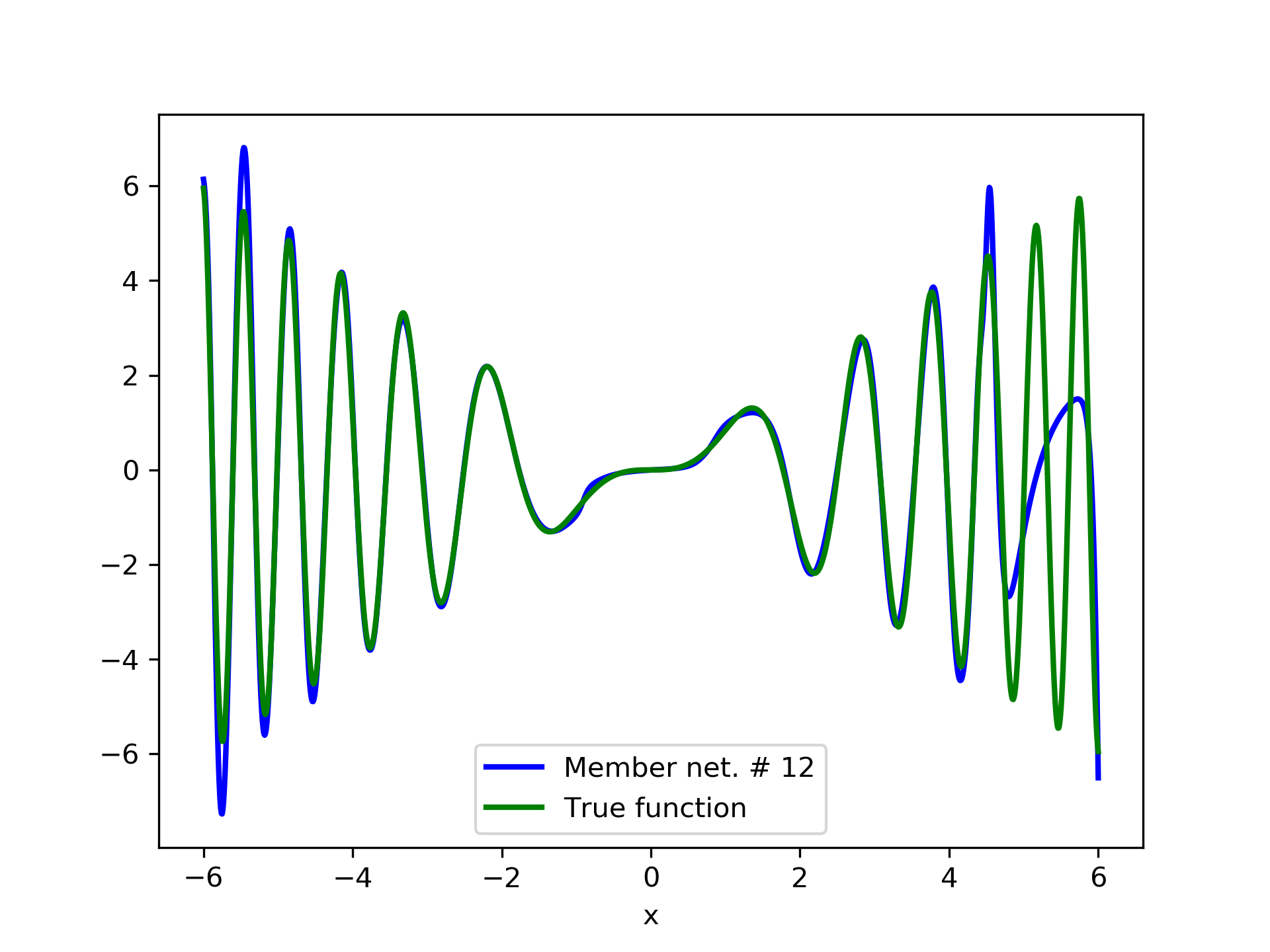}
\end{subfigure}
\begin{subfigure}{0.33\textwidth}
    \centering
    \includegraphics[width=\textwidth]{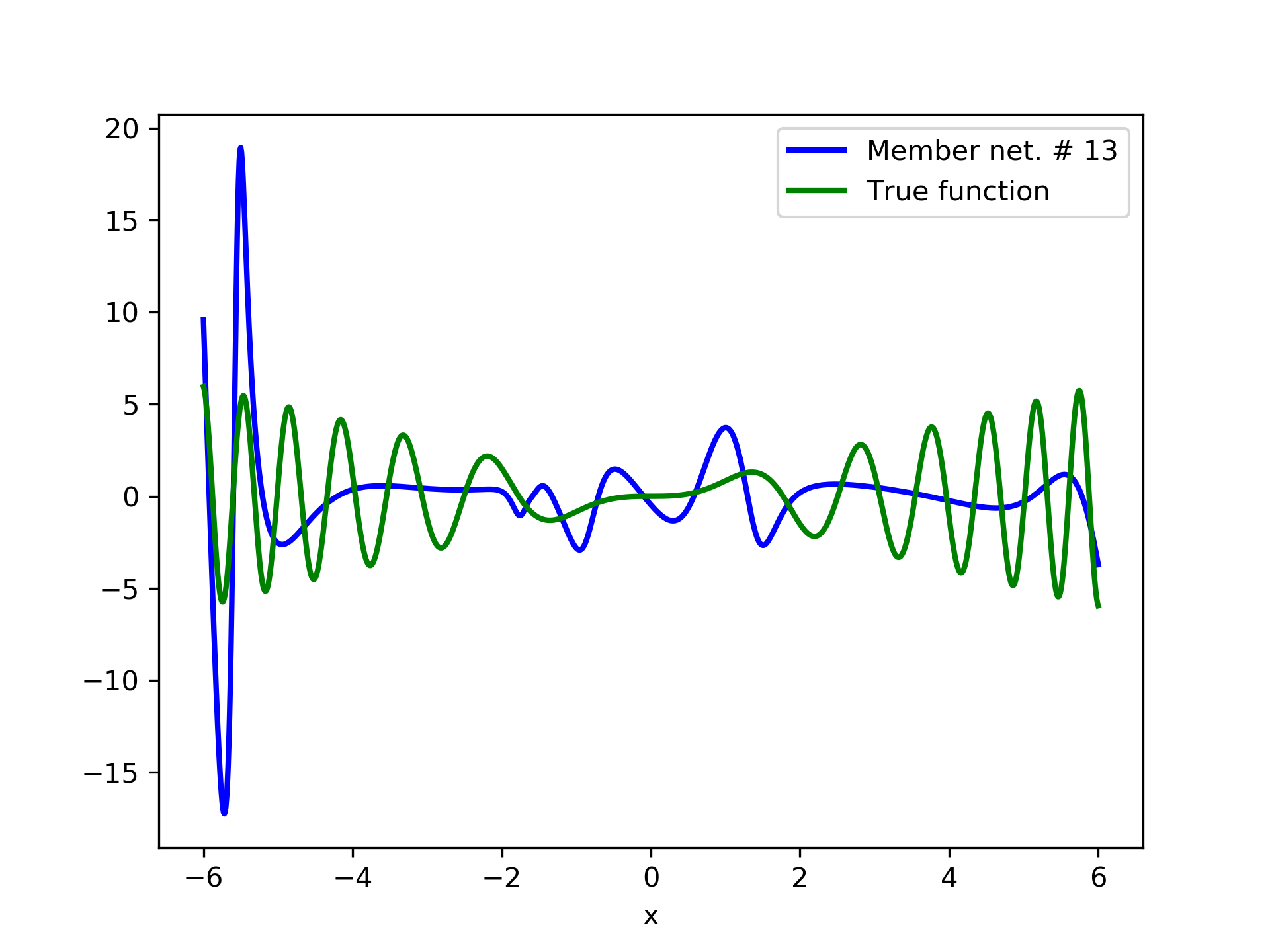}
\end{subfigure}\hfill
\begin{subfigure}{0.33\textwidth}
    \centering
    \includegraphics[width=\textwidth]{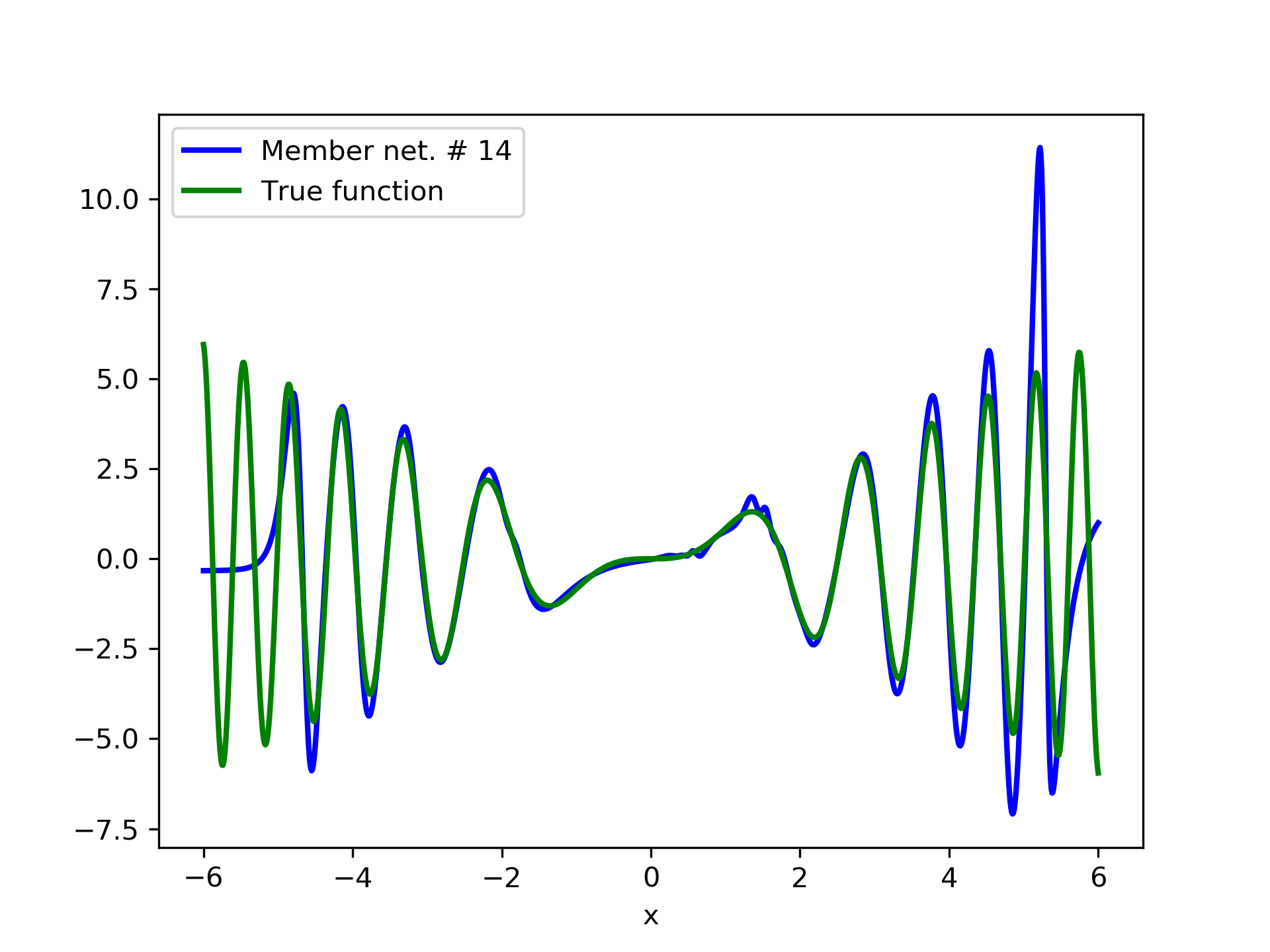}
\end{subfigure}\hfill
\begin{subfigure}{0.33\textwidth}
    \centering
    \includegraphics[width=\textwidth]{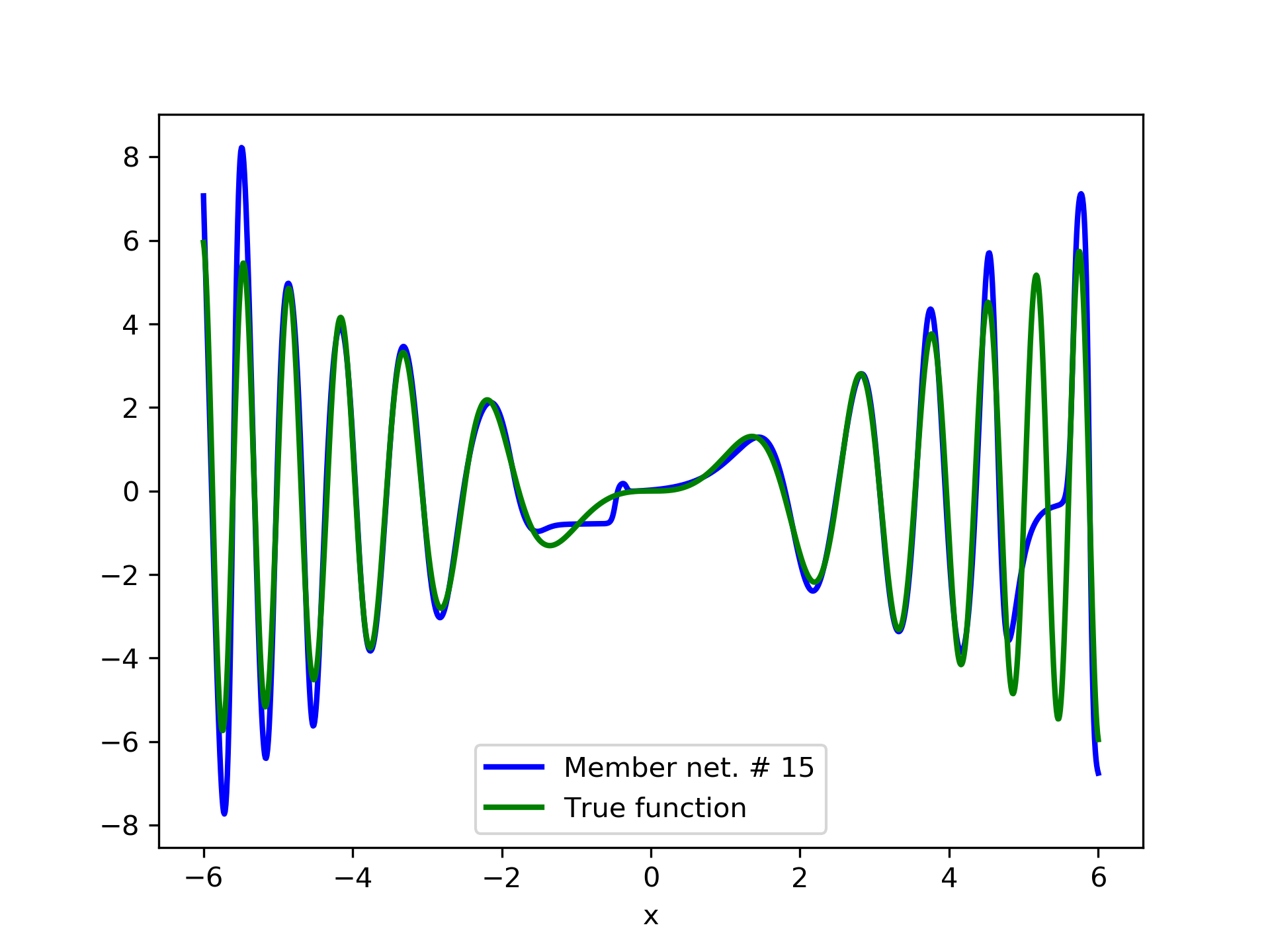}
\end{subfigure}
\caption{Plot of the ANN members related to table \ref{EXPF1W}, along with the plot of $f_1(x)$.}
\label{MXMEMBERS}
\end{figure}

\begin{figure}[h]
\centering
\begin{subfigure}{0.33\textwidth}
    \centering
    \includegraphics[width=\textwidth]{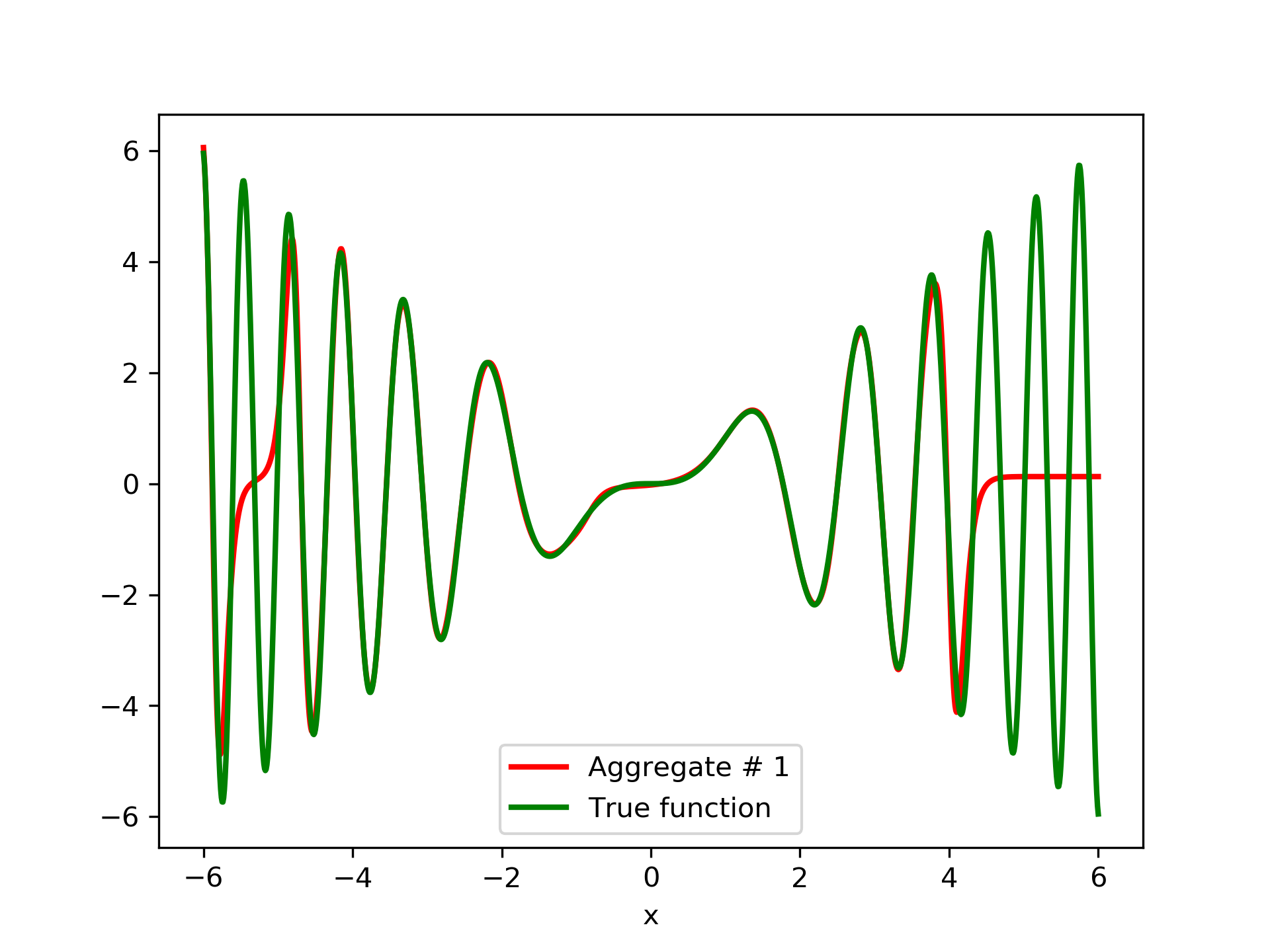}
\end{subfigure}\hfill
\begin{subfigure}{0.33\textwidth}
    \centering
    \includegraphics[width=\textwidth]{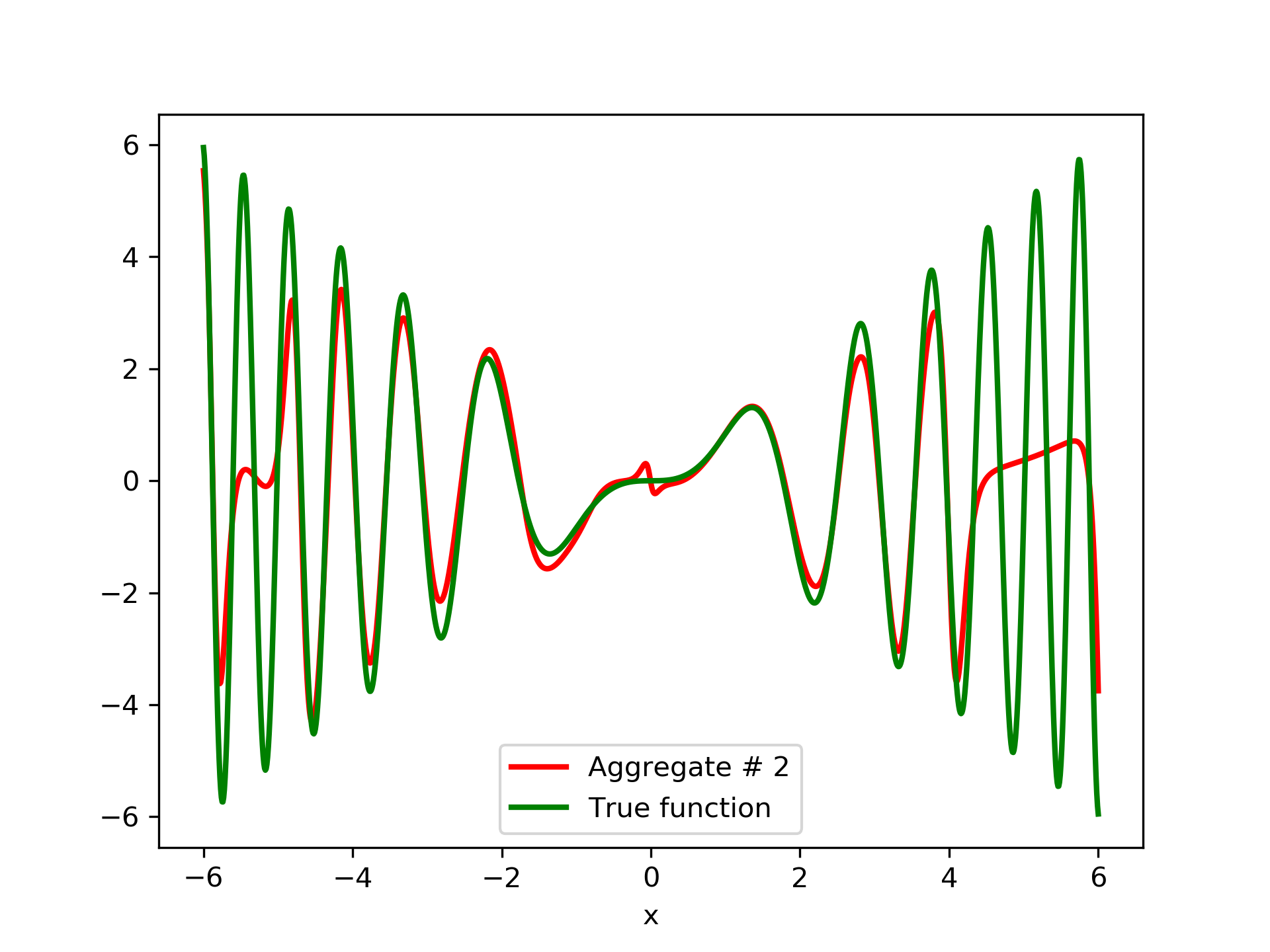}
\end{subfigure}\hfill
\begin{subfigure}{0.33\textwidth}
    \centering
    \includegraphics[width=\textwidth]{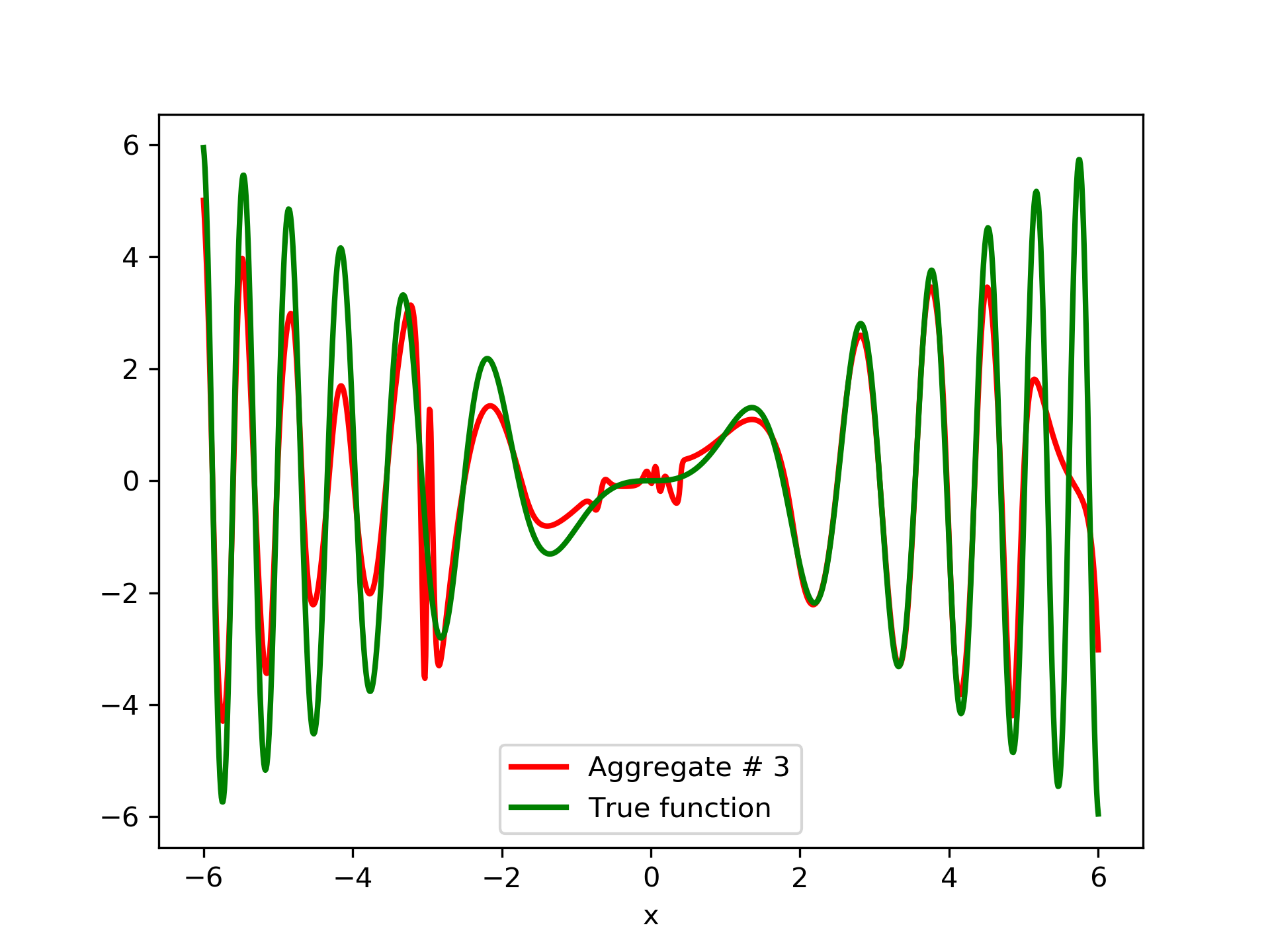}
\end{subfigure}
\begin{subfigure}{0.33\textwidth}
    \centering
    \includegraphics[width=\textwidth]{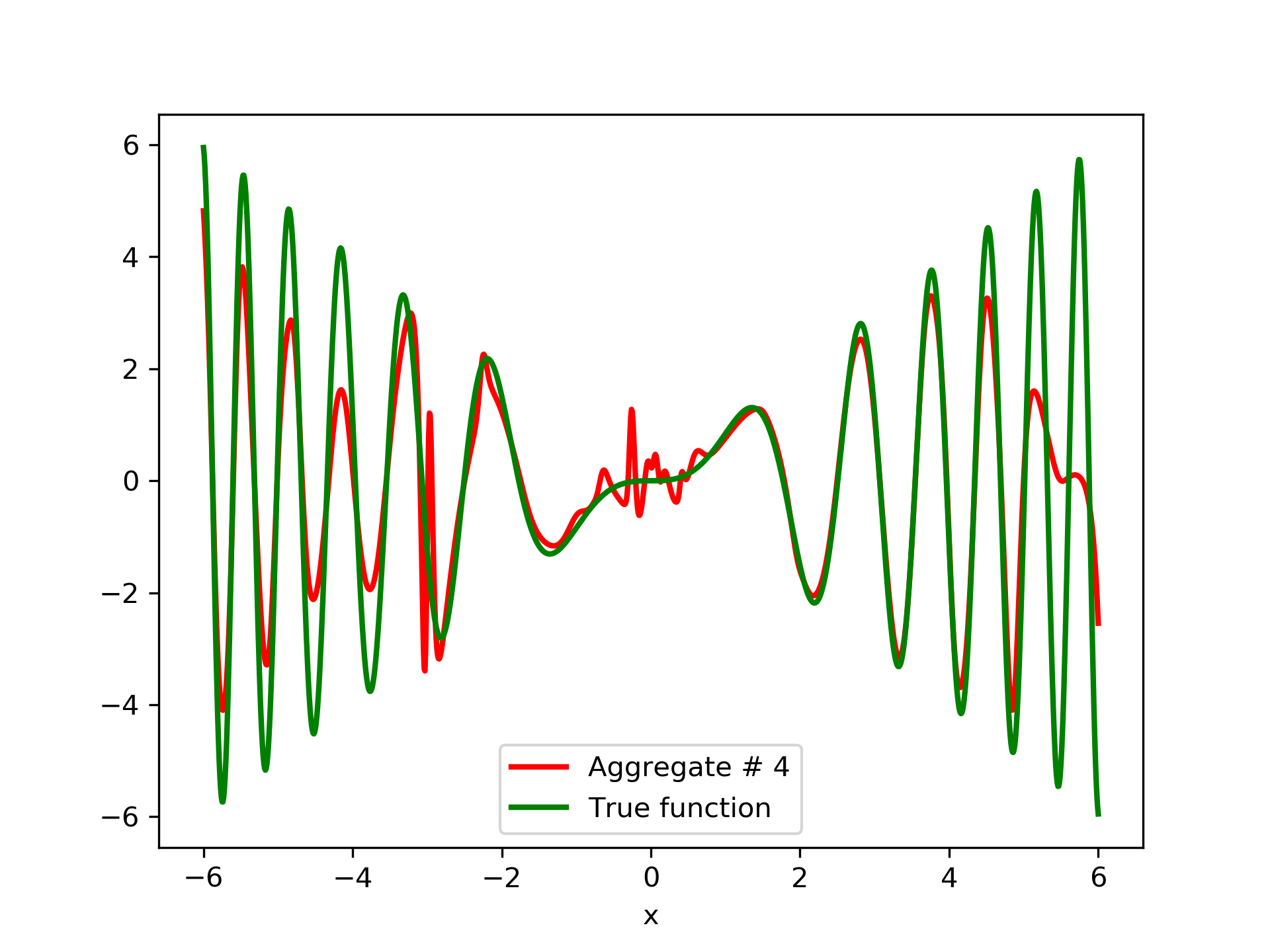}
\end{subfigure}\hfill
\begin{subfigure}{0.33\textwidth}
    \centering
    \includegraphics[width=\textwidth]{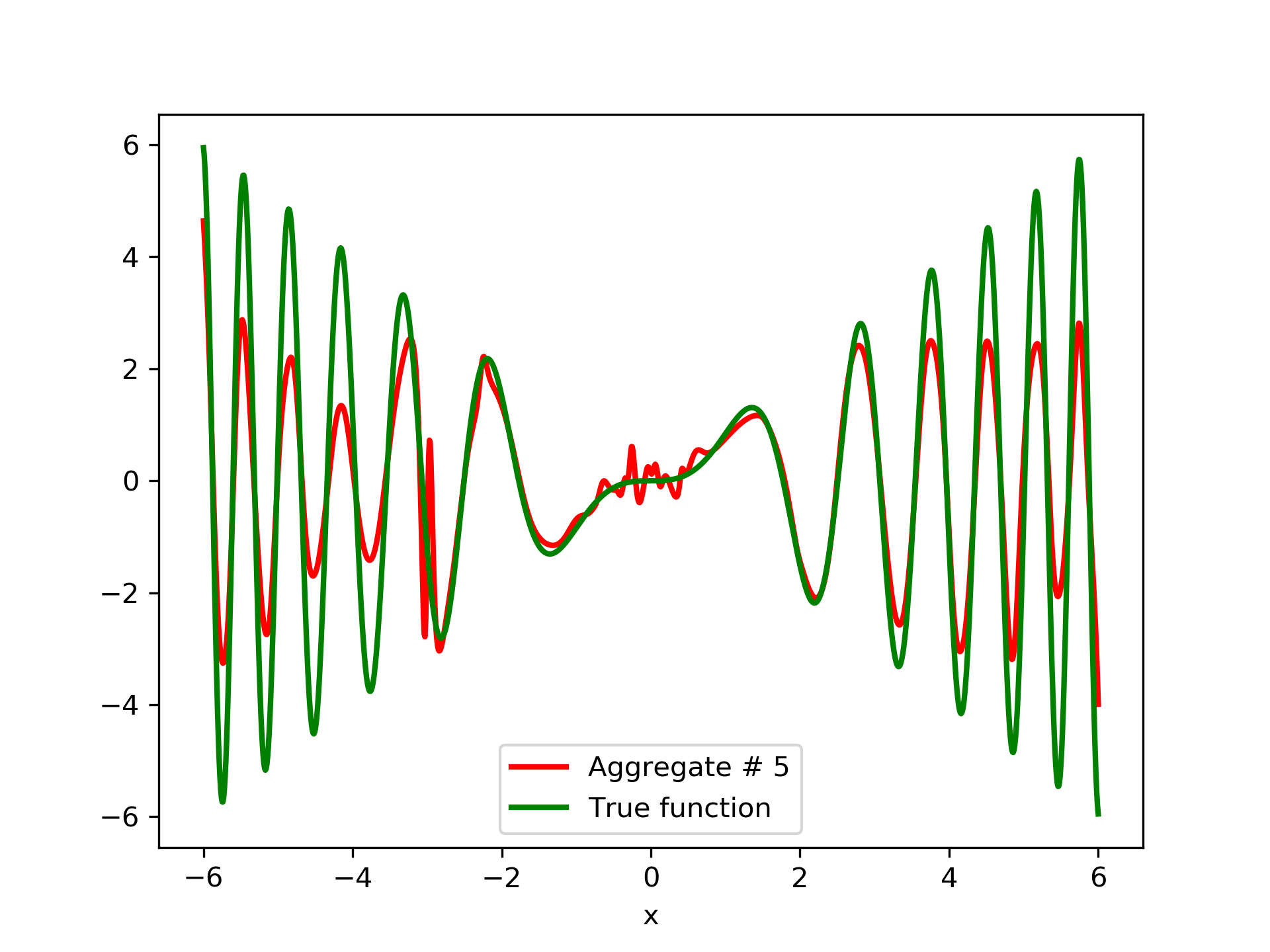}
\end{subfigure}\hfill
\begin{subfigure}{0.33\textwidth}
    \centering
    \includegraphics[width=\textwidth]{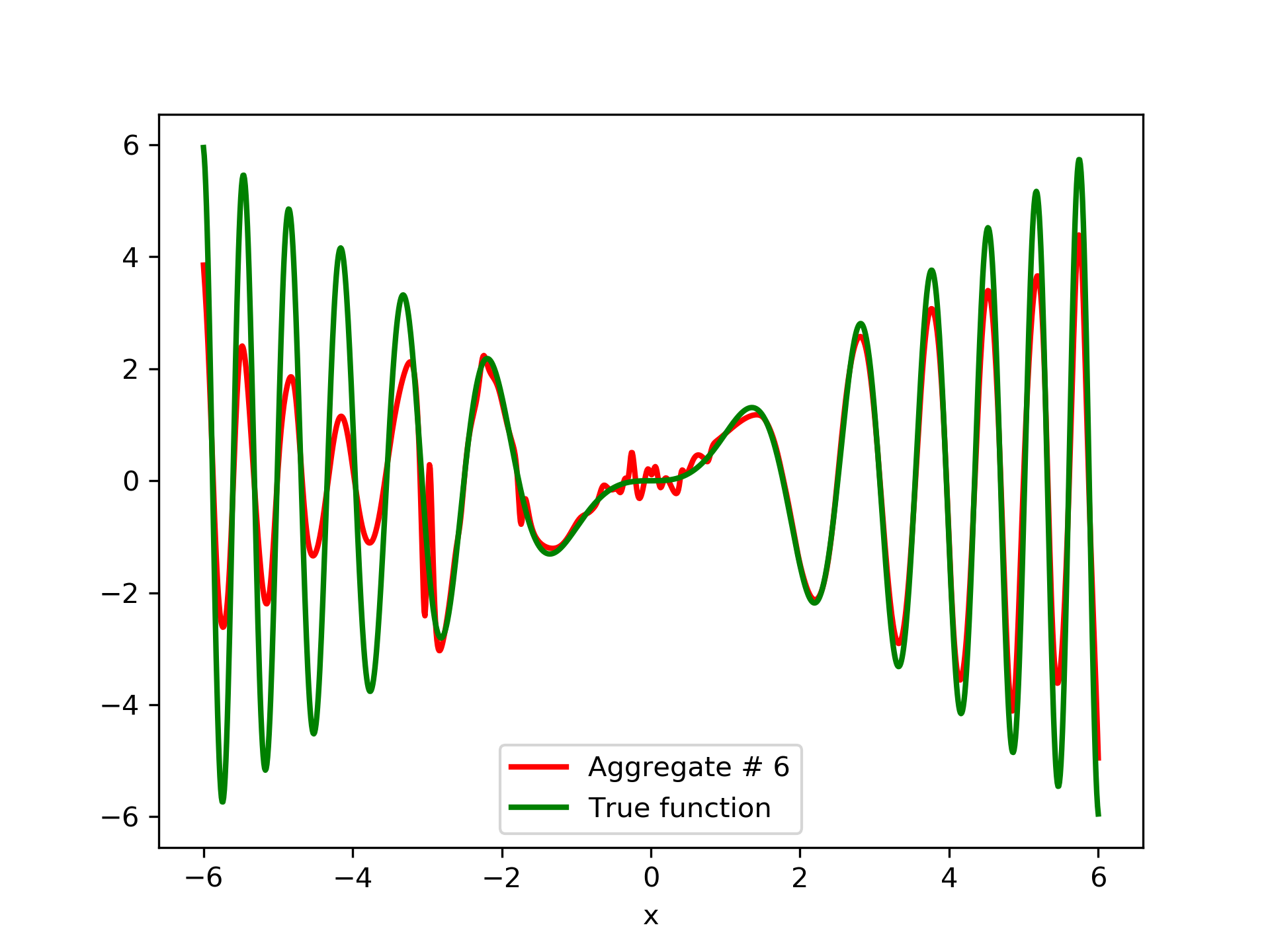}
\end{subfigure}
\begin{subfigure}{0.33\textwidth}
    \centering
    \includegraphics[width=\textwidth]{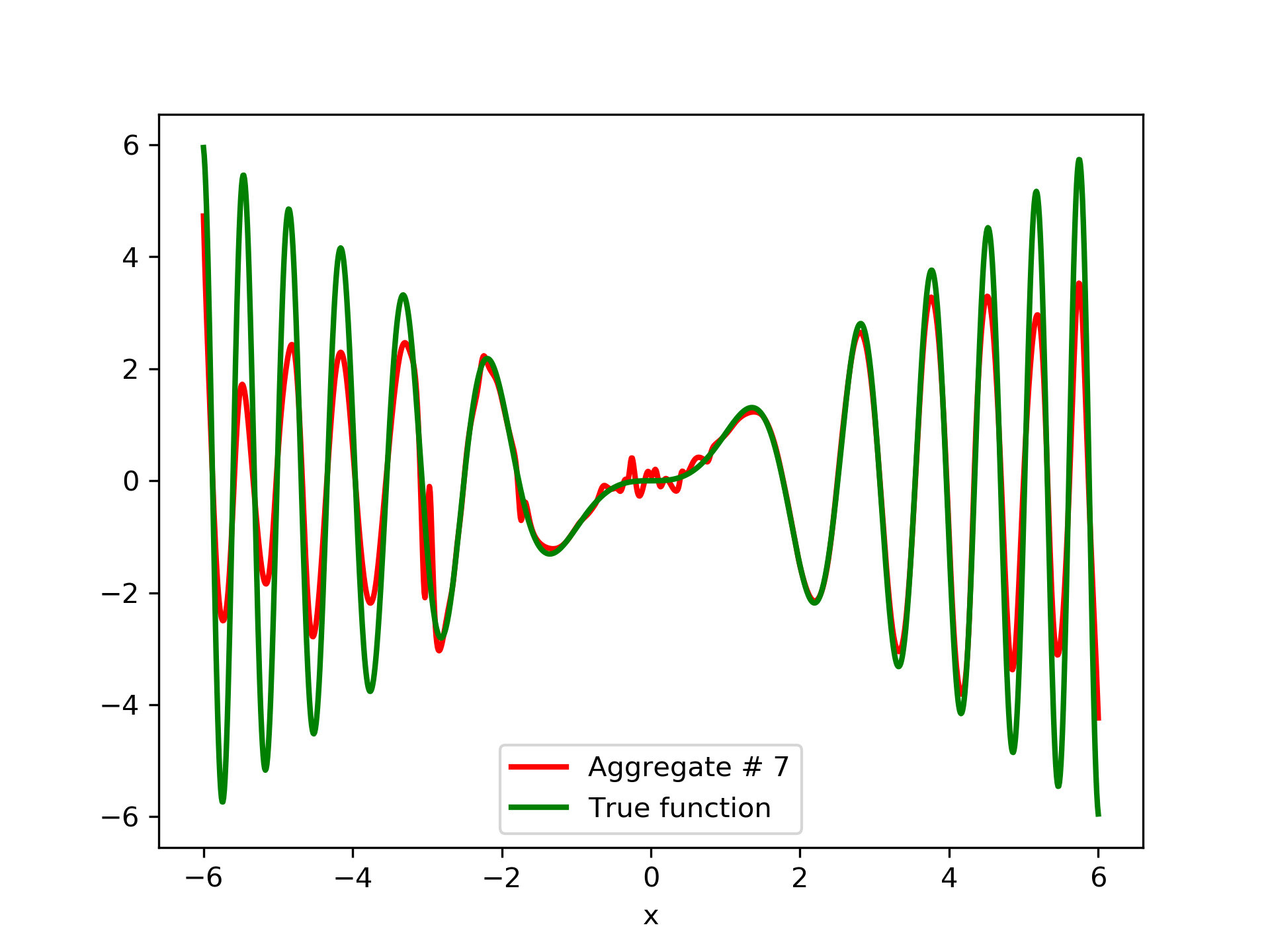}
\end{subfigure}\hfill
\begin{subfigure}{0.33\textwidth}
    \centering
    \includegraphics[width=\textwidth]{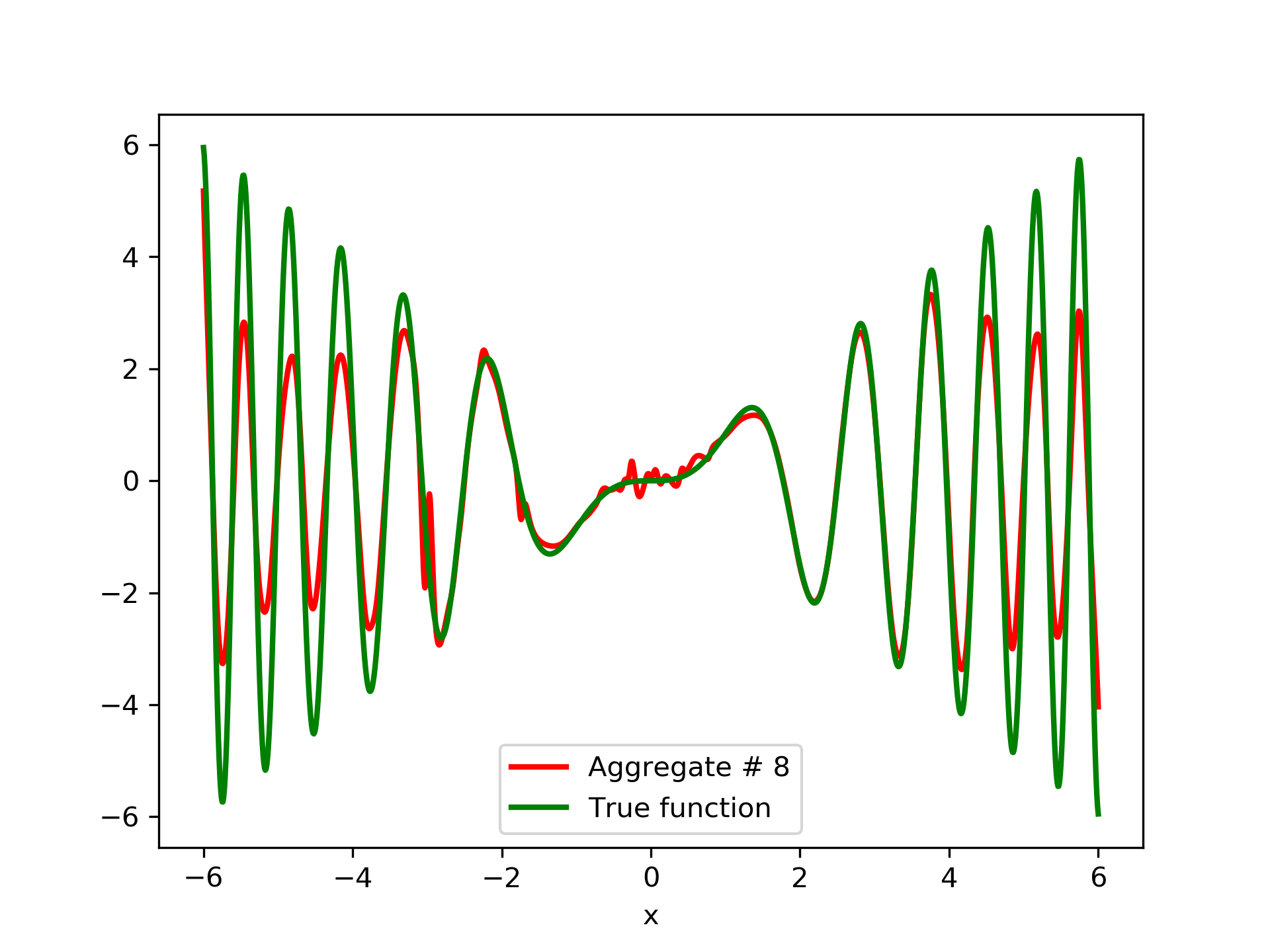}
\end{subfigure}\hfill
\begin{subfigure}{0.33\textwidth}
    \centering
    \includegraphics[width=\textwidth]{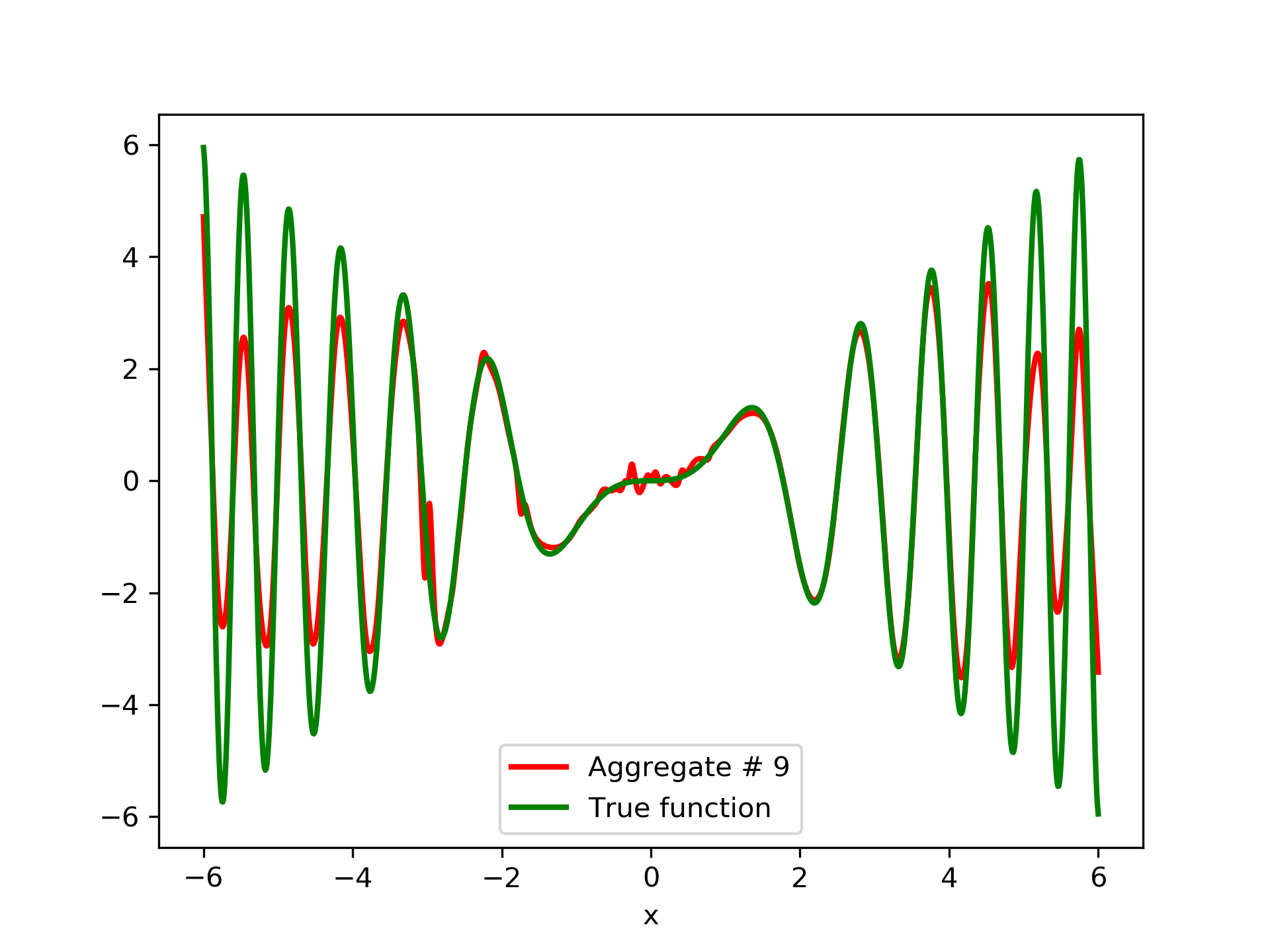}
\end{subfigure}
\begin{subfigure}{0.33\textwidth}
    \centering
    \includegraphics[width=\textwidth]{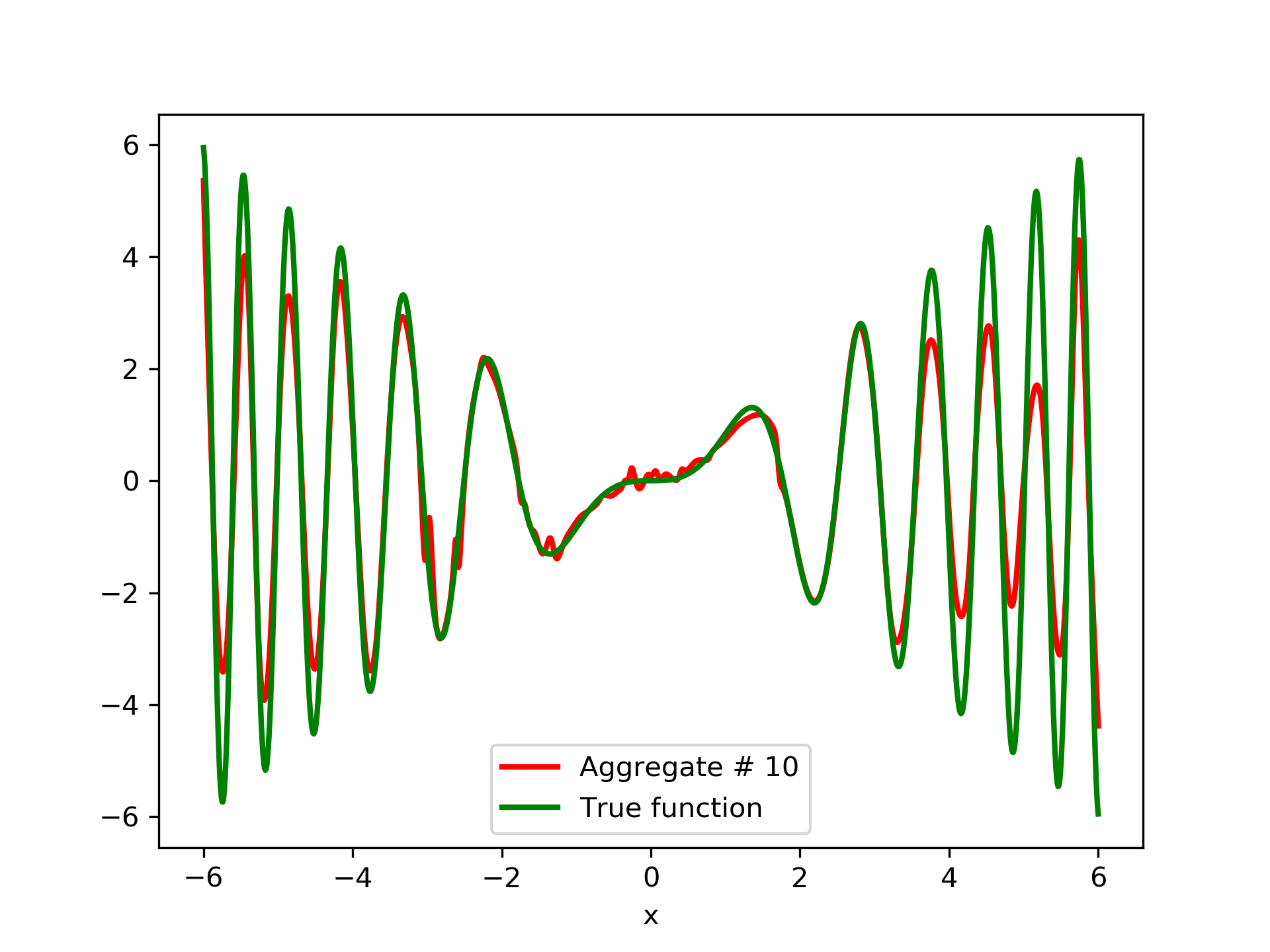}
\end{subfigure}\hfill
\begin{subfigure}{0.33\textwidth}
    \centering
    \includegraphics[width=\textwidth]{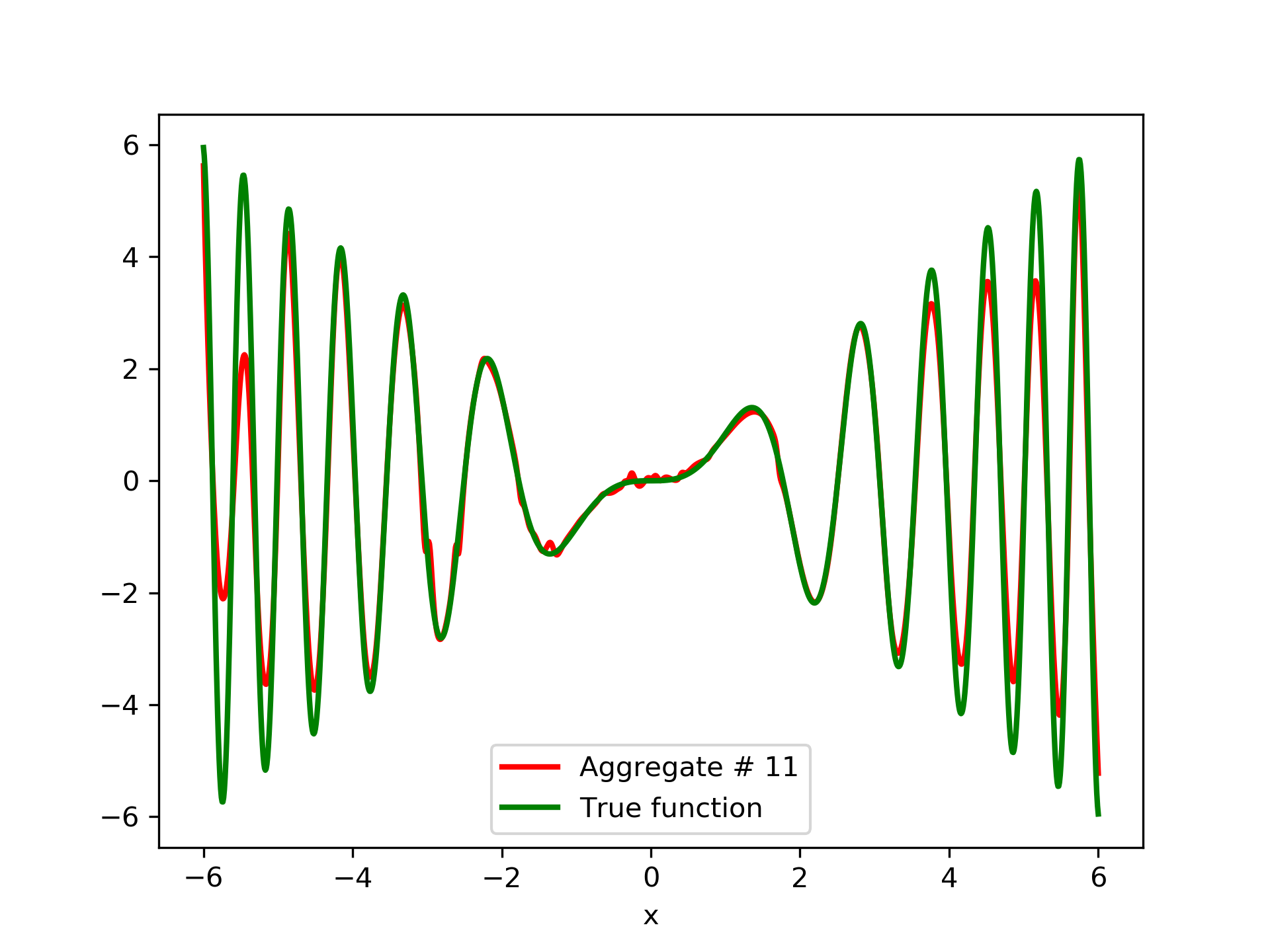}
\end{subfigure}\hfill
\begin{subfigure}{0.33\textwidth}
    \centering
    \includegraphics[width=\textwidth]{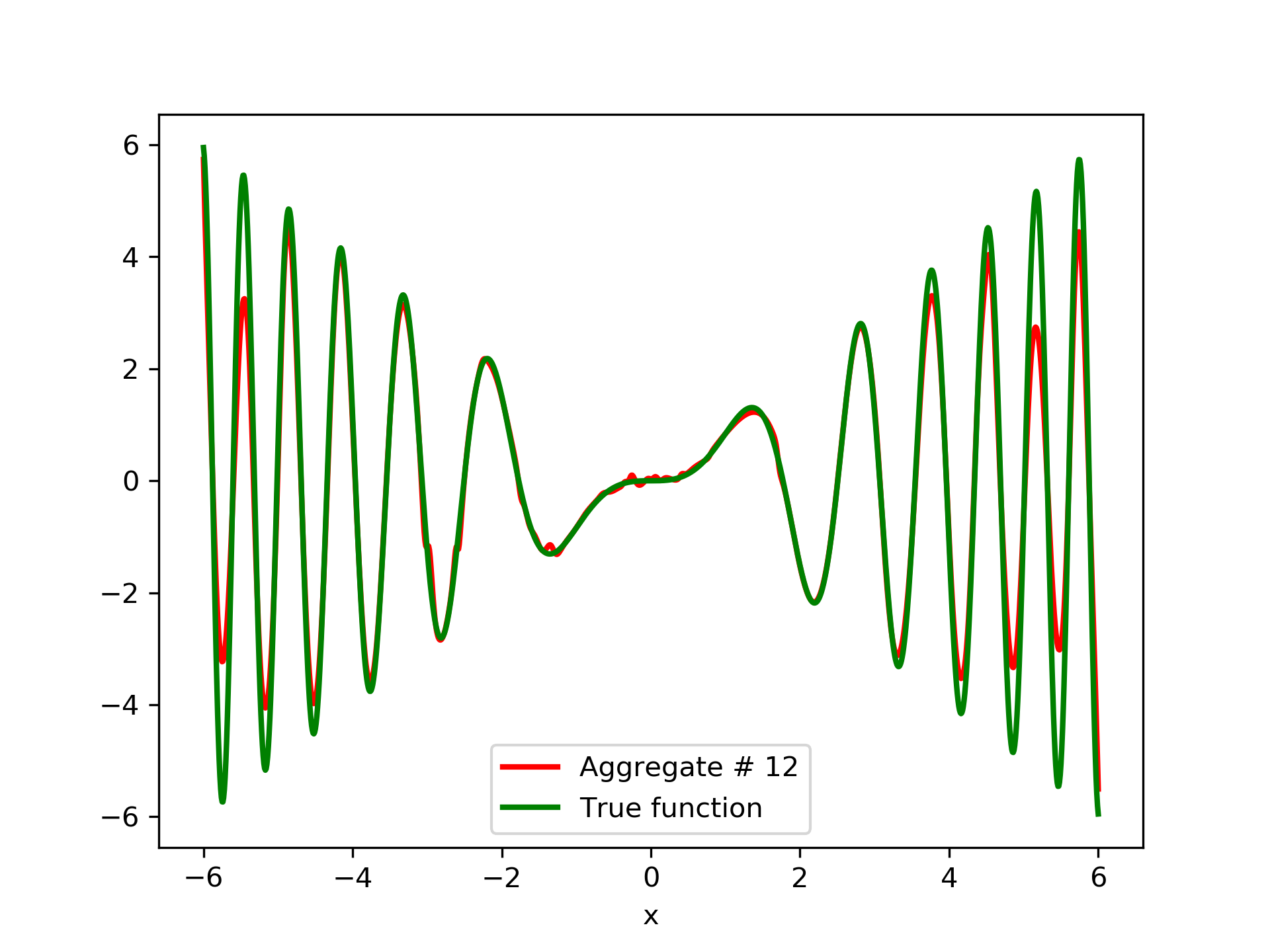}
\end{subfigure}
\begin{subfigure}{0.33\textwidth}
    \centering
    \includegraphics[width=\textwidth]{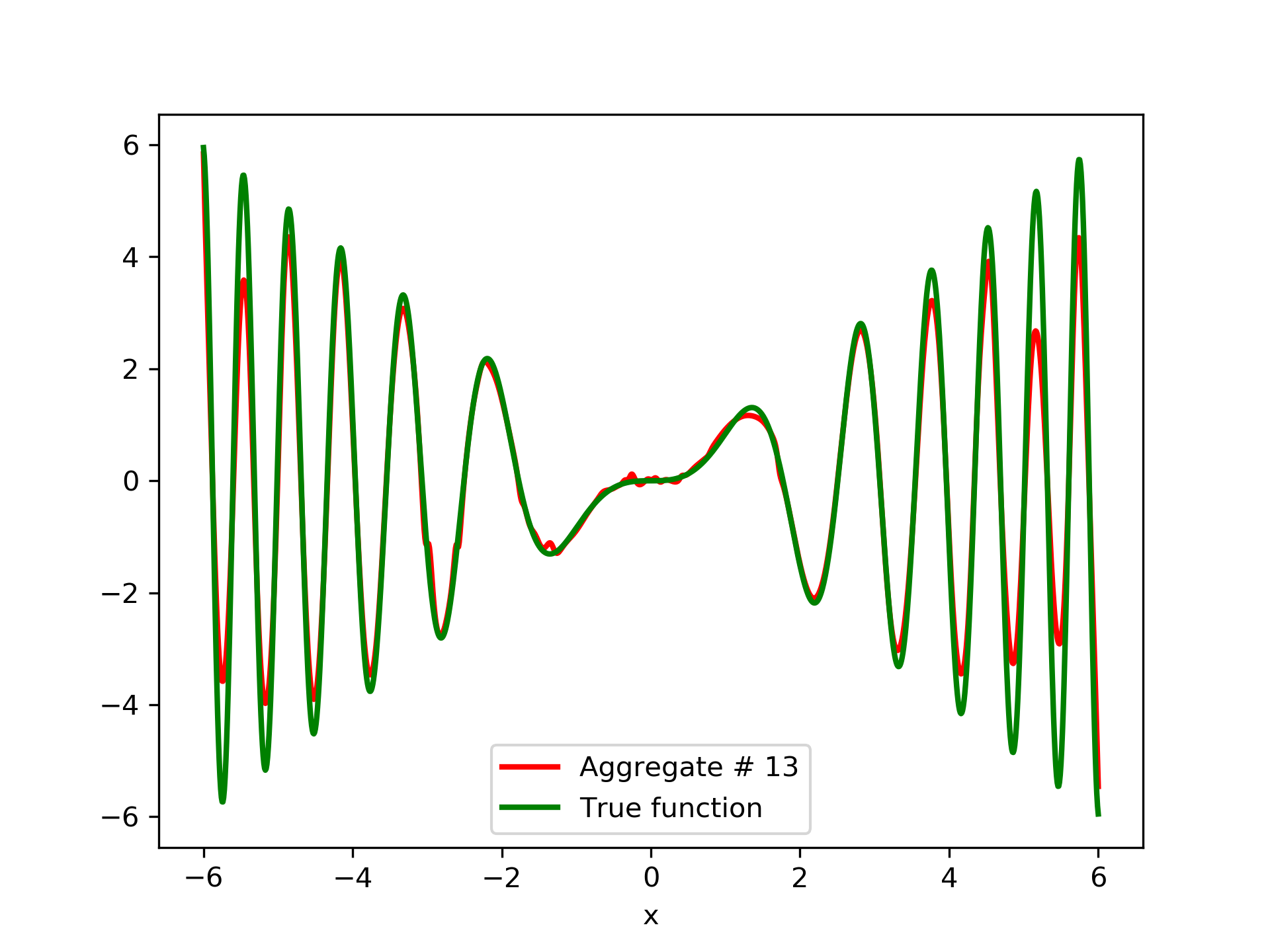}
\end{subfigure}\hfill
\begin{subfigure}{0.33\textwidth}
    \centering
    \includegraphics[width=\textwidth]{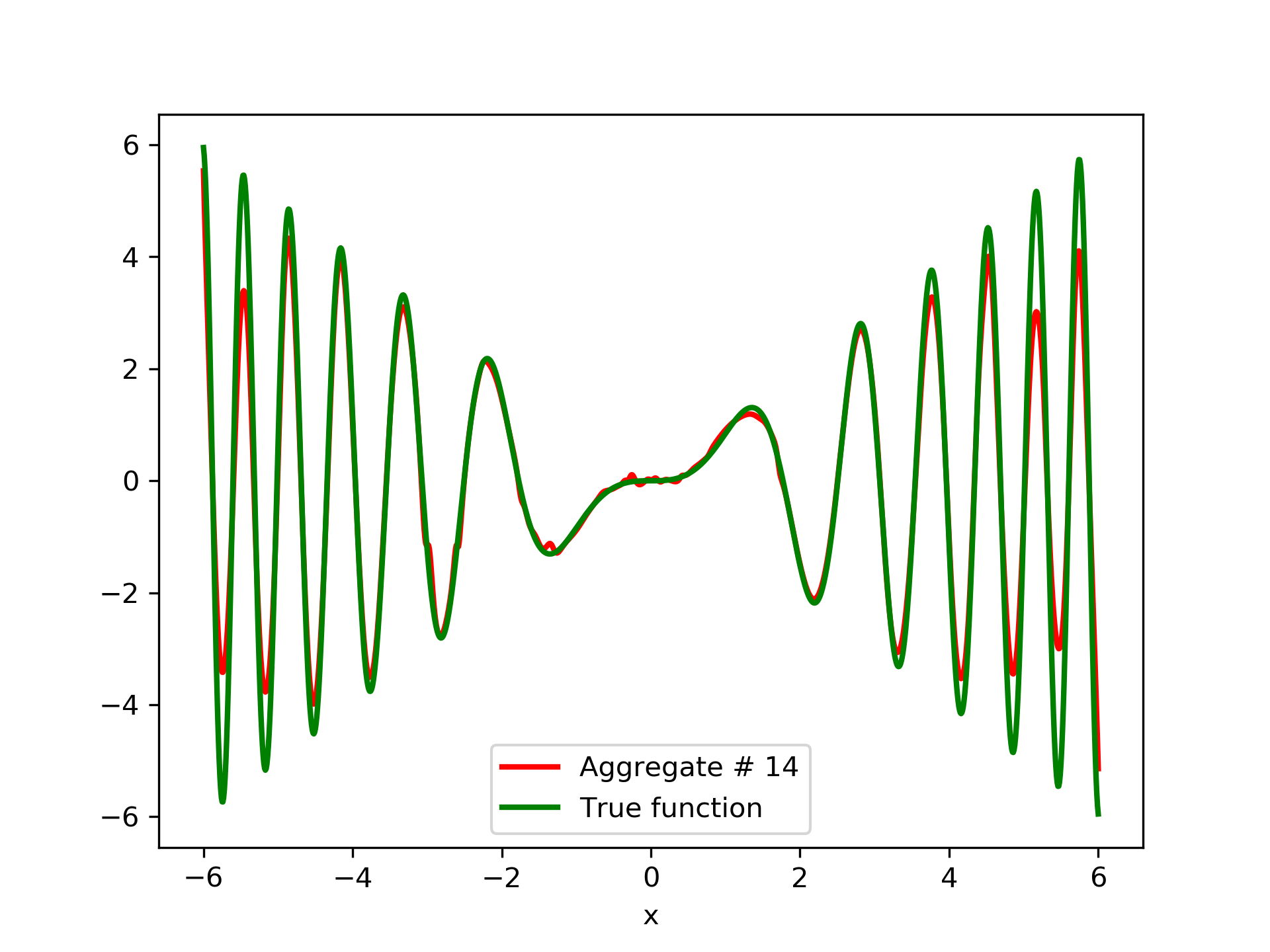}
\end{subfigure}\hfill
\begin{subfigure}{0.33\textwidth}
    \centering
    \includegraphics[width=\textwidth]{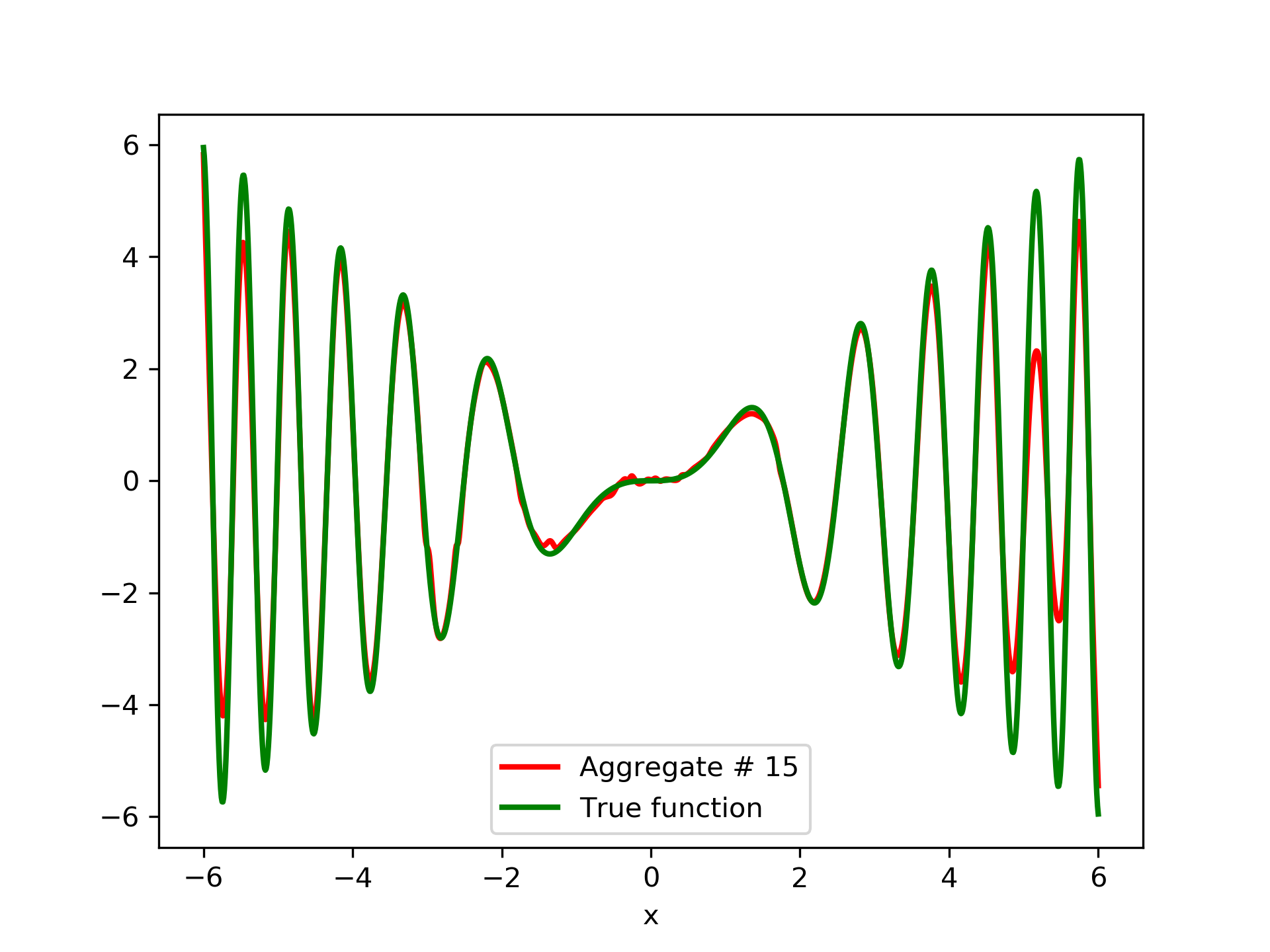}
\end{subfigure}
\caption{Plot of the aggregate ANNs related to table \ref{EXPF1W}, along with the plot of $f_1(x)$.}
\label{MXAGGREGATES}
\end{figure}
\clearpage
In fig. \ref{MXMEMBERS}, the member networks are plotted. Note that networks $N_2(x)$, $N_4(x)$, $N_5(x)$ and $N_{13}(x)$, clearly overfit $f_1(x)$. However the aggregate networks, plotted in figure \ref{MXAGGREGATES}, eliminate this effect gradually, by appropriate weighting. Note that the corresponding coefficients $a_2, a_4, a_5$ and $a_{13}$ are among the lowest ones in the linear combination composing the final aggregate network. 
Among the 15 models, model No. 11 has the minimum training MSE, which is 0.98583. The training MSE of final aggregate model is 0.31371, which is $68\%$ lower than that of model No. 11.  
\subsection{Test Case 3}
We have also experimented with the multidimensional function, $f_2(\mathbf{x})$, where we have set $d=4$. Again, we use MLP's with one hidden layer. 
For this function, we have used a grid of 15 equidistant points along each dimension, producing so a set $D$ containing a total of 50,625 points. 
We have chosen 2,000 of these points at random (seed is set to 12345) for the training set $T_r$, and the rest were used for the test set $T_s$. 
Detailed results of the experiments with $f_2(\mathbf{x})$, are listed in \mbox{Table \ref{tab:f3}.} Model No. 6 has a training  MSE of 0.04537, which is the lowest among the 10 trained models. The training  MSE of the final aggregate  is 0.02547, which is $43\%$ lower than that of model No. 6.

\begin{table}[h]
\caption{Experiment with $f_2(\mathbf{x})$ with $\mathbf{x}\in[-1.5, 1.5]^4$ and a training set of 2,000 points chosen randomly from a set of 50,625 equidistant points. Regularization factor: $\nu_{reg}=0.05$.}
\begin{center}
\begin{tabular}{|c|c|c|c|c|c|c|c|}
\hline
    ANN & Nodes   & Type    & MSE      &$\beta$   &  AG. MSE &  AG. MSE/TE & $a$  \\
\hline
 1 & 38 & sigmoid & 0.25331 &  0.00000 &  0.25331 & 0.29718 &  0.06849 \\
 \hline
 2 & 38 & tanh    & 0.14980 &  0.37494 &  0.09163 & 0.10700 &  0.11418 \\
 \hline
 3 & 37 & sigmoid & 0.35718 &  0.80263 &  0.07454 & 0.08540 &  0.04492 \\
 \hline
 4 & 37 & sigmoid & 0.31343 &  0.82541 &  0.06335 & 0.07445 &  0.04814 \\
 \hline
 5 & 39 & sigmoid & 0.54320 &  0.91340 &  0.05900 & 0.06983 &  0.02614 \\
 \hline
 6 & 39 & tanh    & 0.04537 &  0.40159 &  0.03420 & 0.04083 &  0.44982 \\
 \hline
 7 & 40 & sigmoid & 0.17129 &  0.86819 &  0.03096 & 0.03691 &  0.11413 \\
 \hline
 8 & 40 & tanh    & 1.09517 &  0.95779 &  0.02889 & 0.03429 &  0.03815 \\
 \hline
 9 & 41 & sigmoid & 0.33329 &  0.93684 &  0.02750 & 0.03331 &  0.06094 \\
 \hline
 10 & 41 & tanh   & 1.56689 &  0.96492 &  0.02547 & 0.03061 &  0.03508 \\
\hline
\end{tabular}
\end{center}
\label{tab:f3}
\end{table}
\subsection{Test Case 4}
\begin{figure}[ht]
\centering
\begin{subfigure}{0.8\textwidth}
    \centering
    \includegraphics[width=0.6\textwidth]{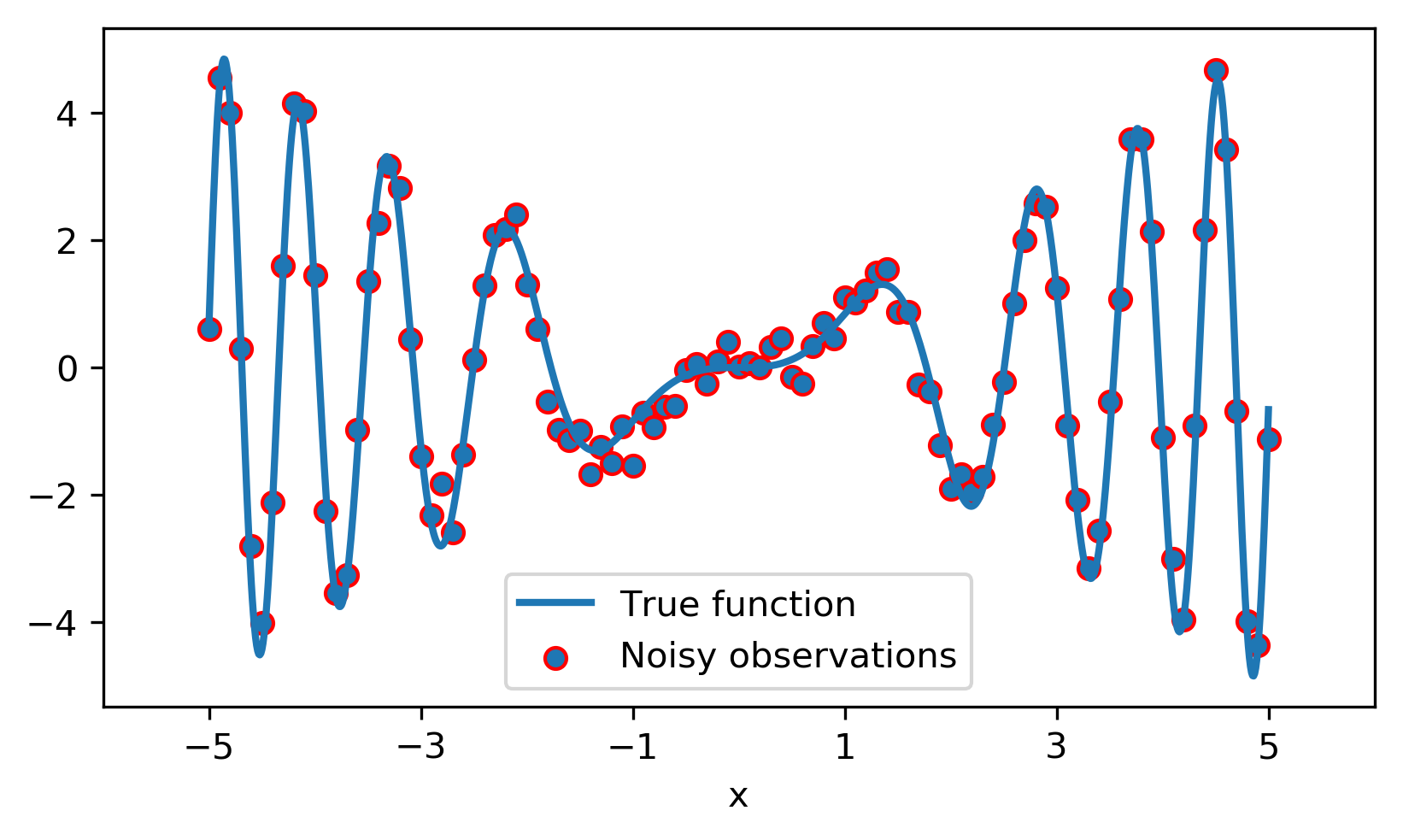}
\end{subfigure}
\caption{Plot of the noise-corrupted  observations used for training, alongside  $f_1(x)$}
\label{fig:noisy}
\end{figure}

\begin{figure}[h]
\centering
\begin{subfigure}{0.33\textwidth}
    \centering
    \includegraphics[width=\textwidth]{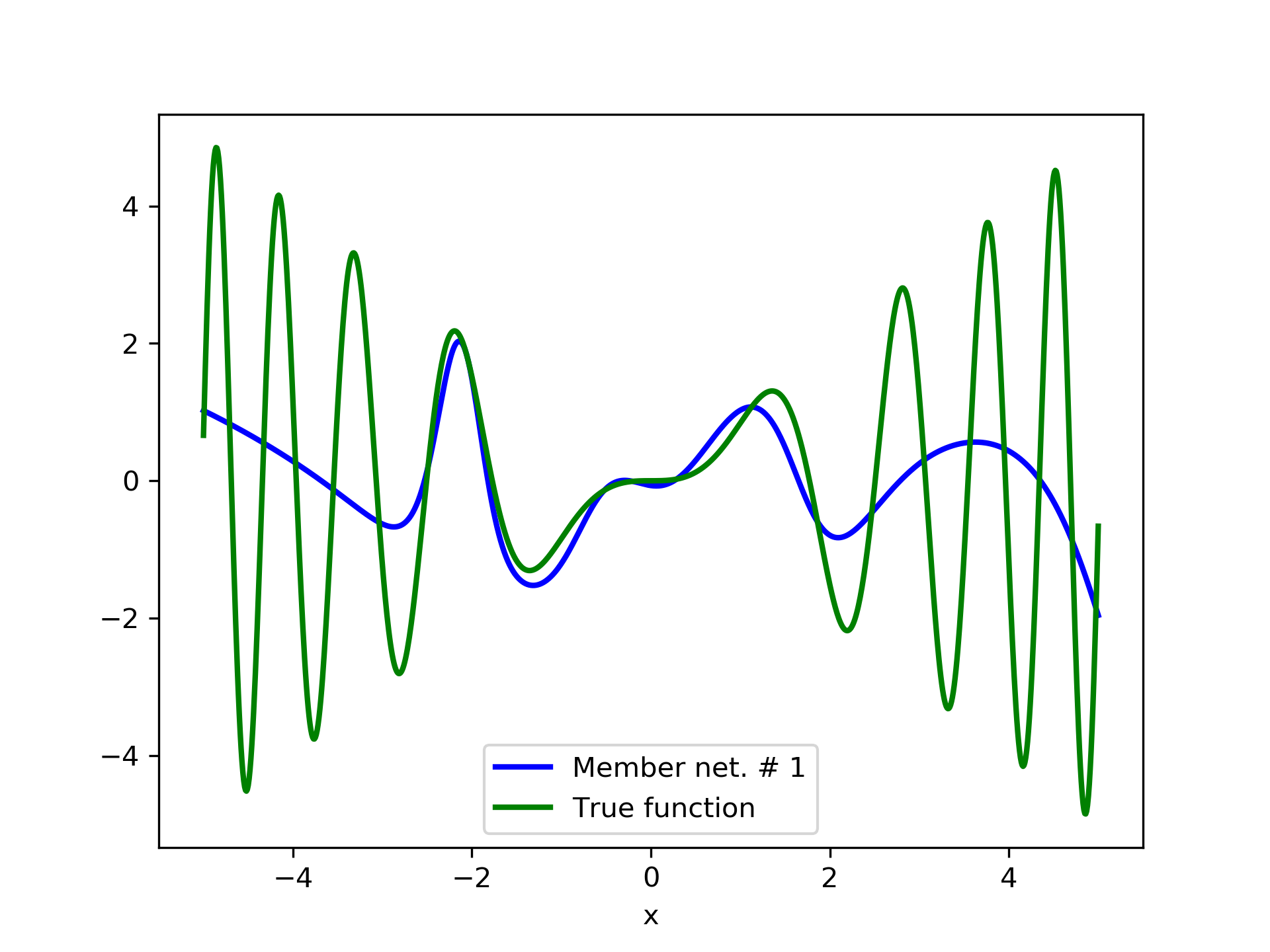}
\end{subfigure}\hfill
\begin{subfigure}{0.33\textwidth}
    \centering
    \includegraphics[width=\textwidth]{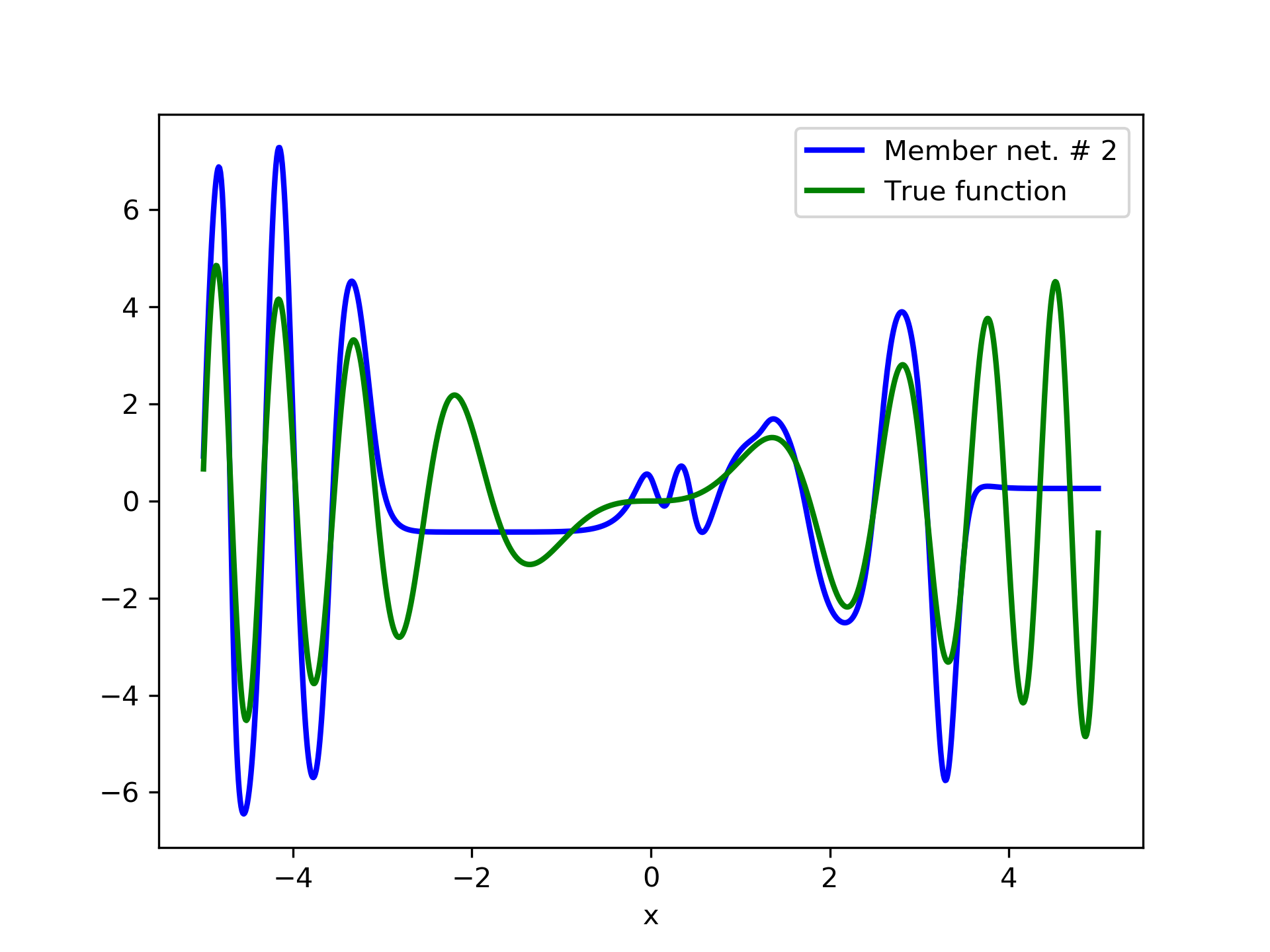}
\end{subfigure}\hfill
\begin{subfigure}{0.33\textwidth}
    \centering
    \includegraphics[width=\textwidth]{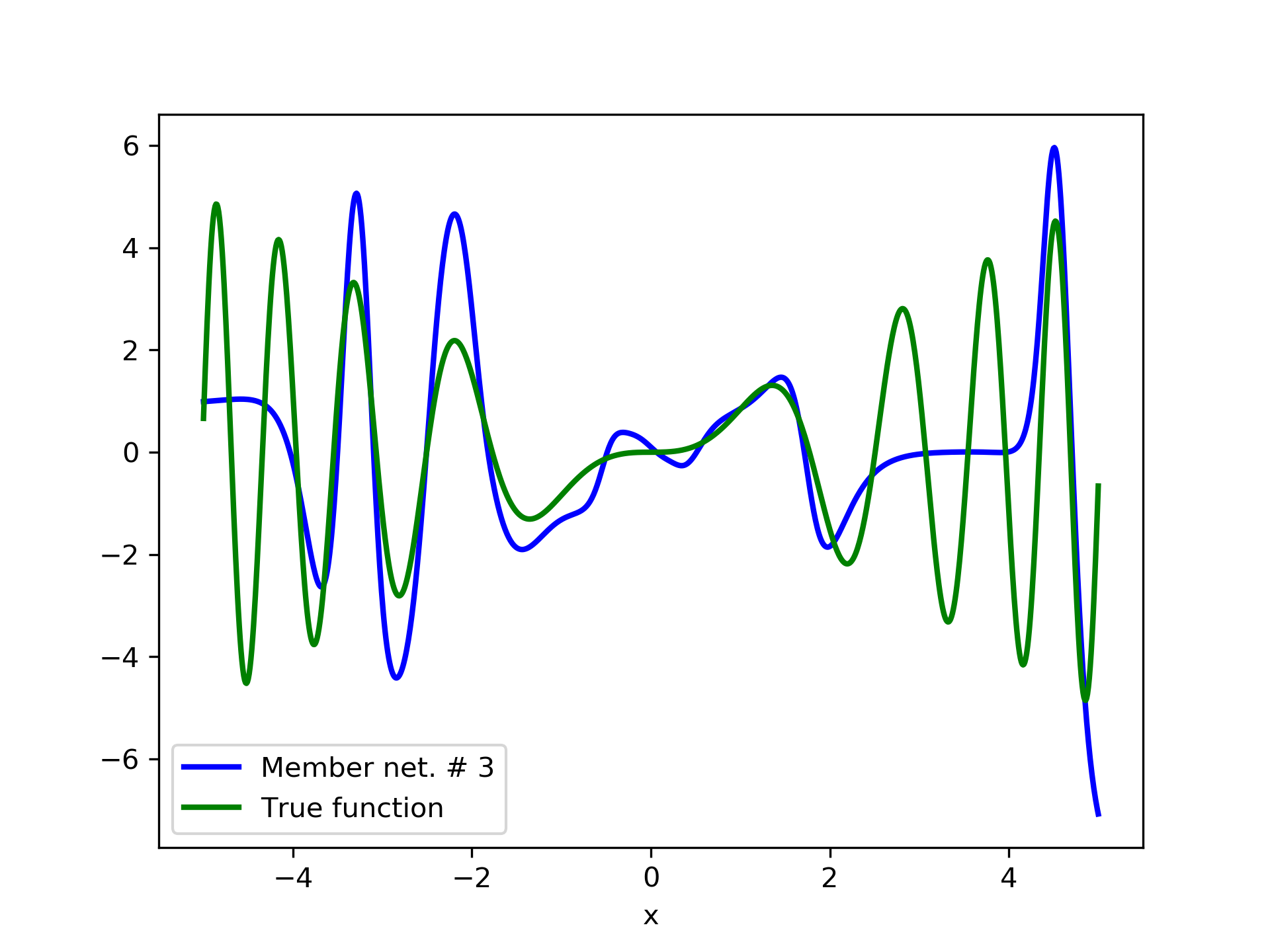}
\end{subfigure}
\begin{subfigure}{0.33\textwidth}
    \centering
    \includegraphics[width=\textwidth]{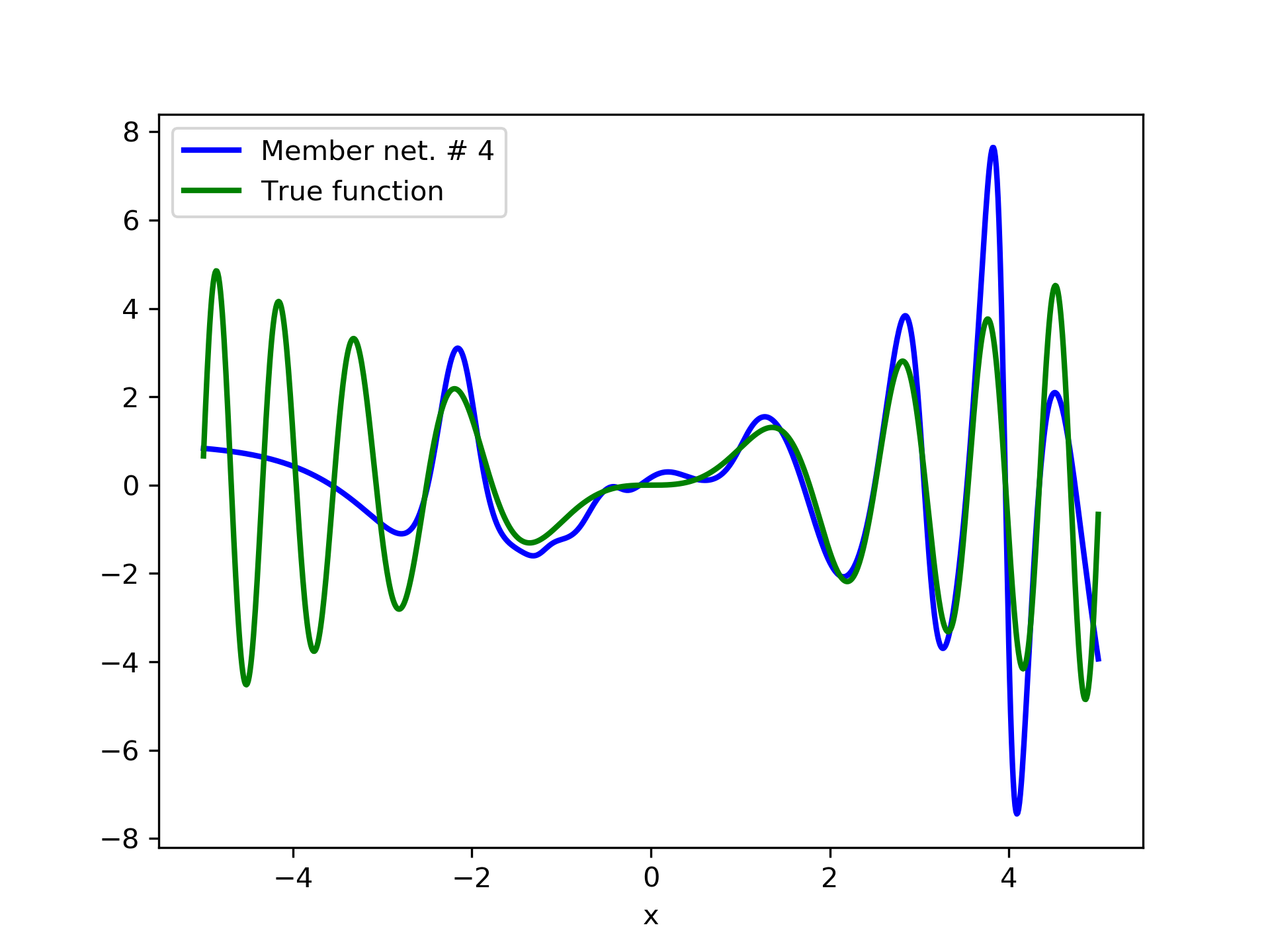}
\end{subfigure}\hfill
\begin{subfigure}{0.33\textwidth}
    \centering
    \includegraphics[width=\textwidth]{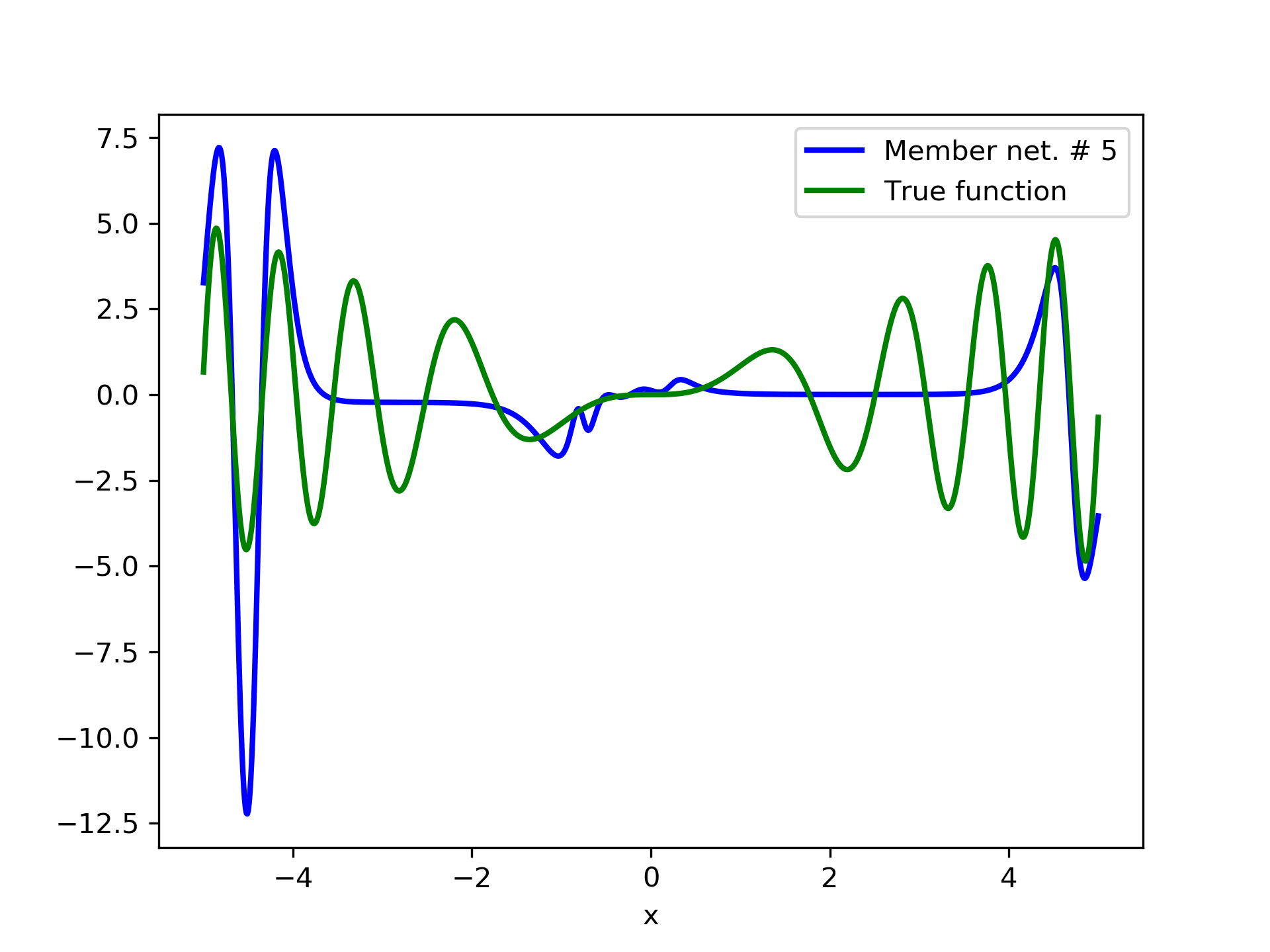}
\end{subfigure}\hfill
\begin{subfigure}{0.33\textwidth}
    \centering
    \includegraphics[width=\textwidth]{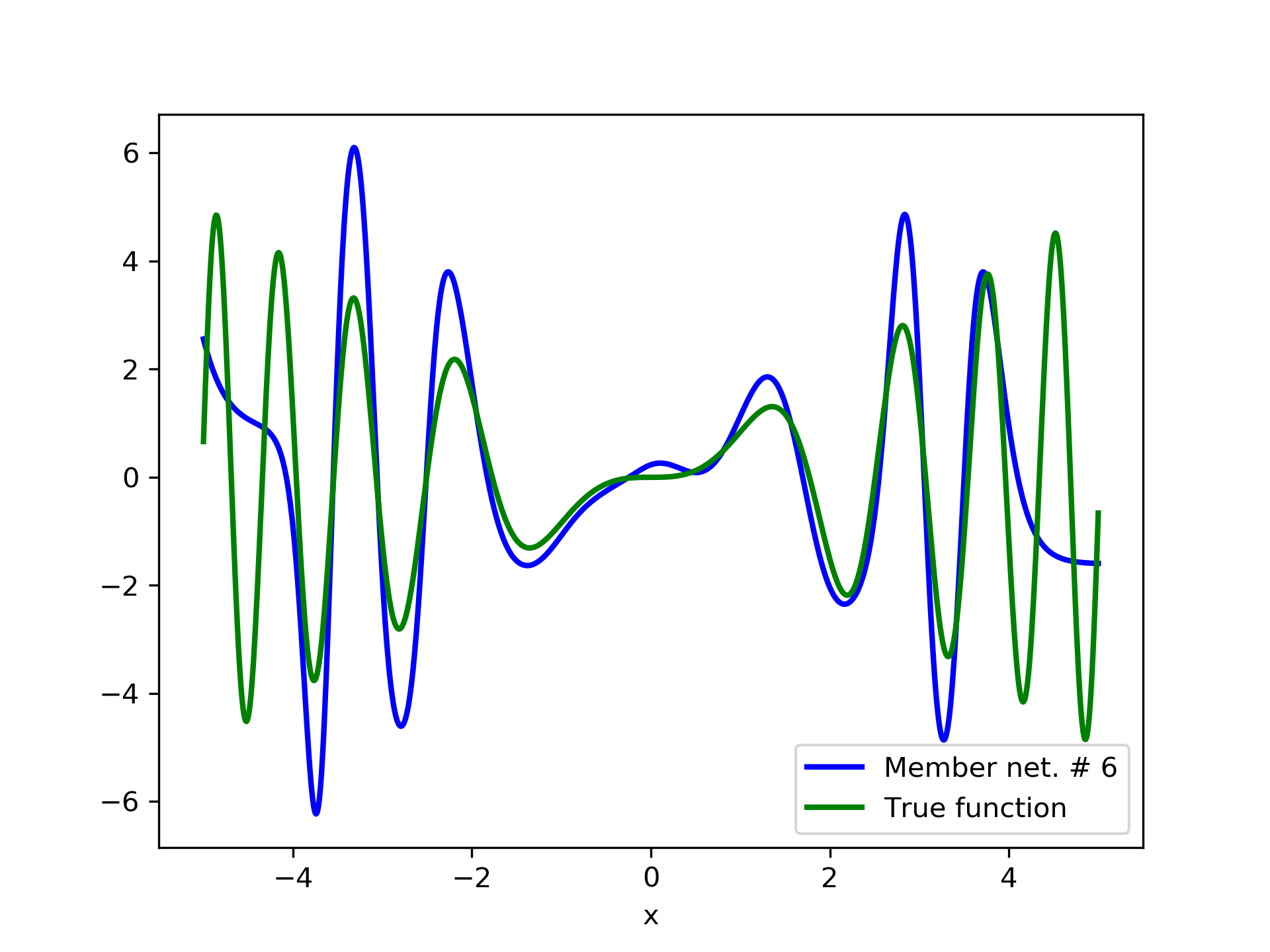}
\end{subfigure}
\begin{subfigure}{0.33\textwidth}
    \centering
    \includegraphics[width=\textwidth]{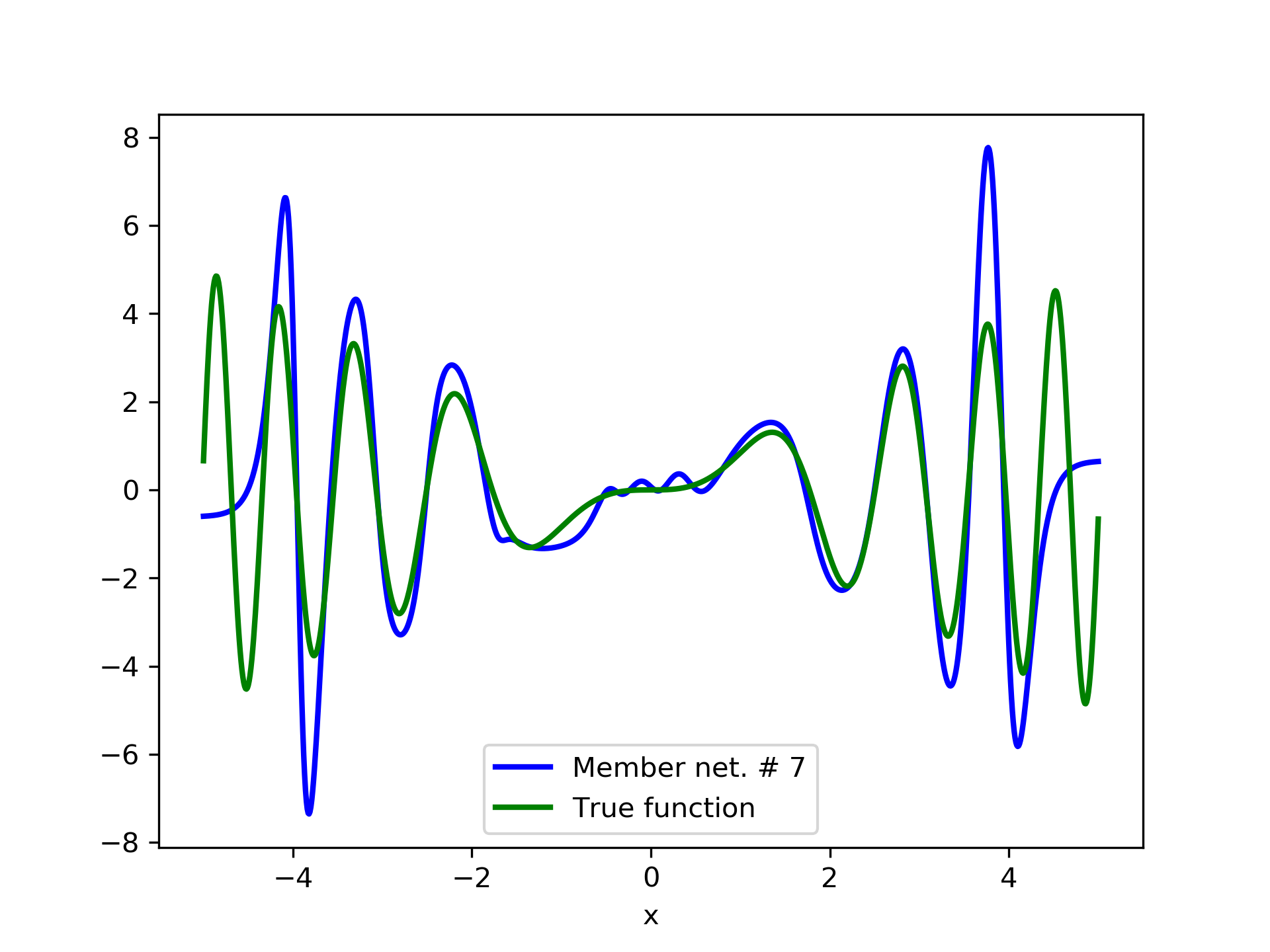}
\end{subfigure}\hfill
\begin{subfigure}{0.33\textwidth}
    \centering
    \includegraphics[width=\textwidth]{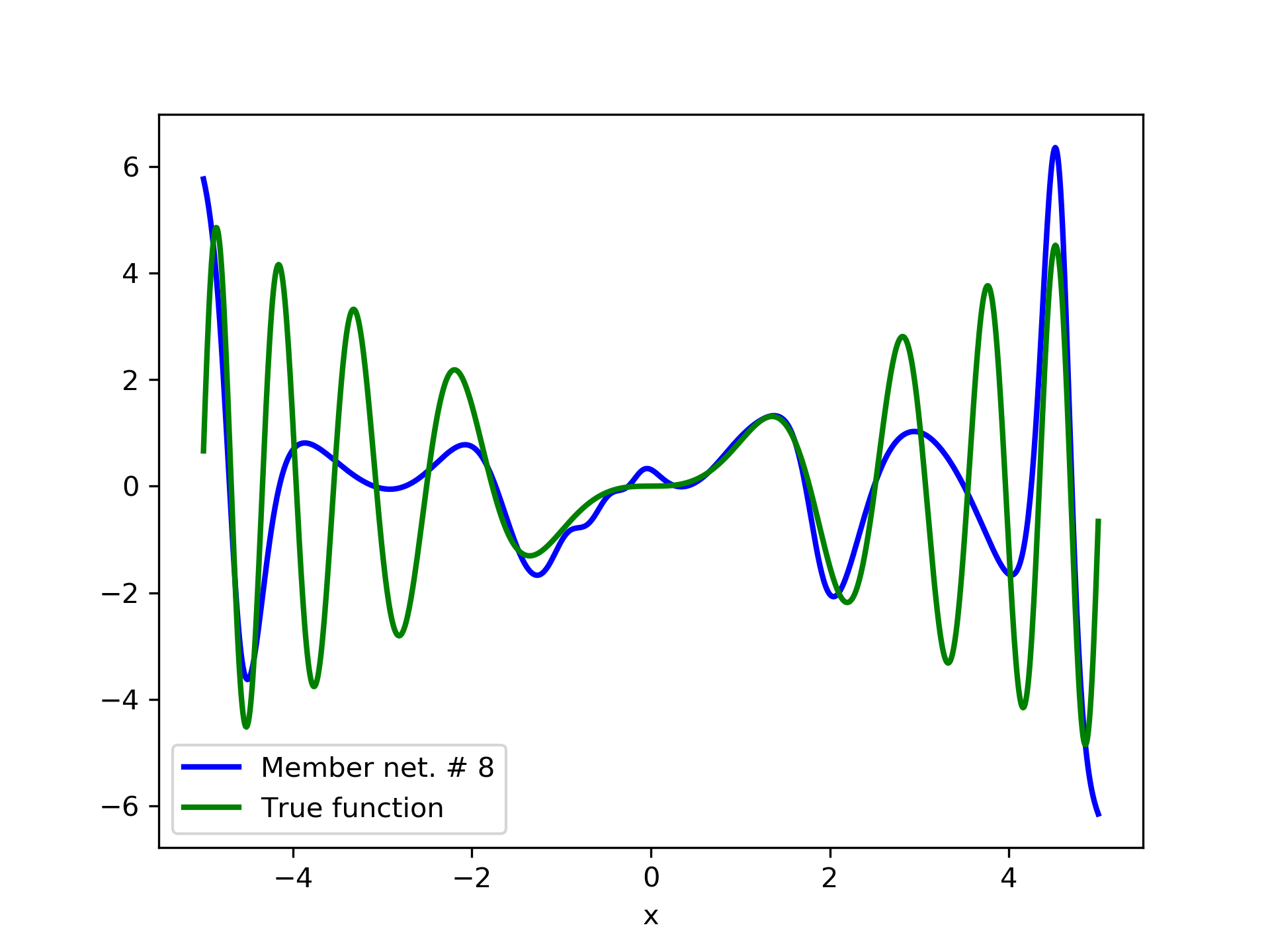}
\end{subfigure}\hfill
\begin{subfigure}{0.33\textwidth}
    \centering
    \includegraphics[width=\textwidth]{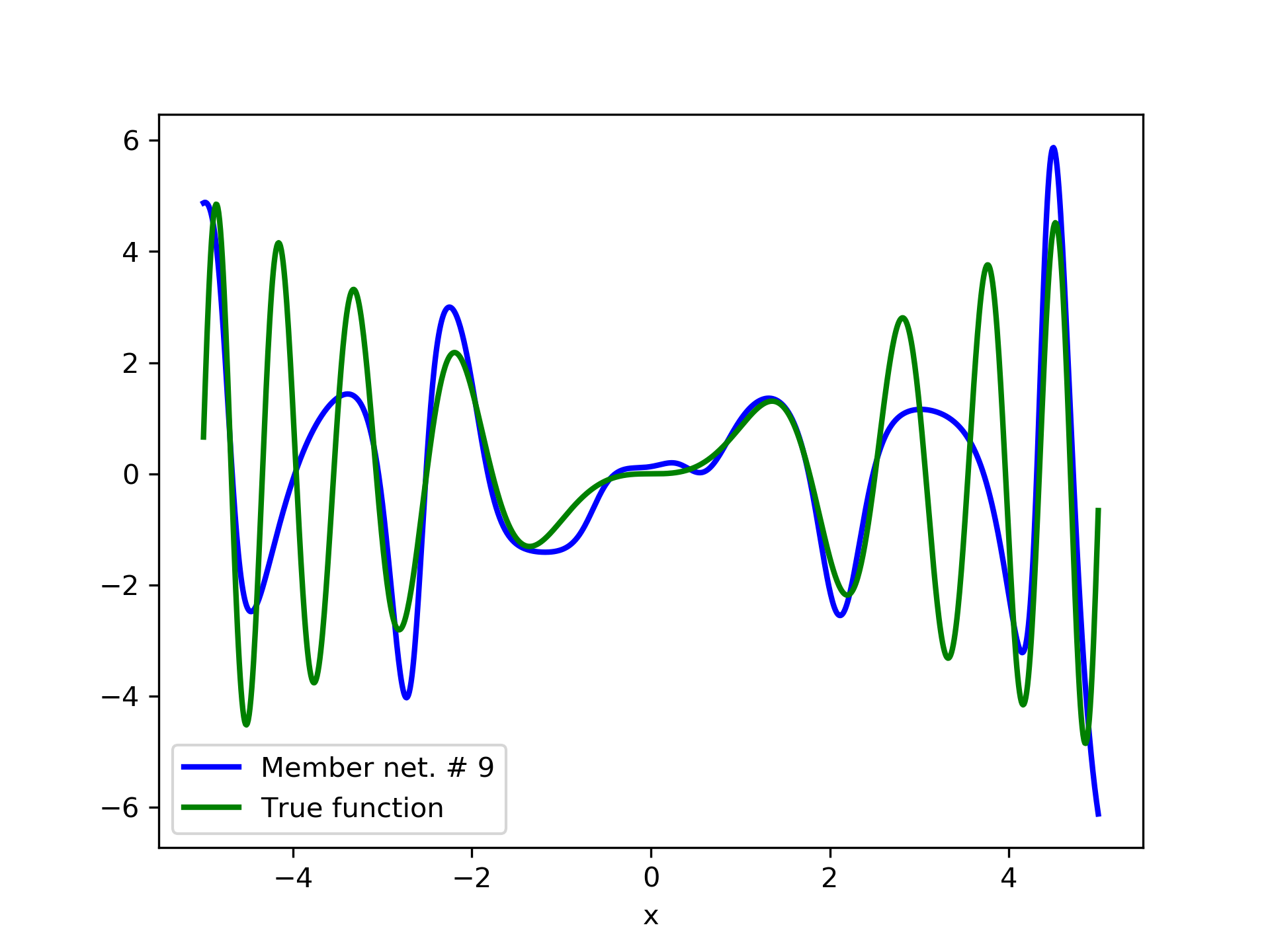}
\end{subfigure}
\begin{subfigure}{0.33\textwidth}
    \centering
    \includegraphics[width=\textwidth]{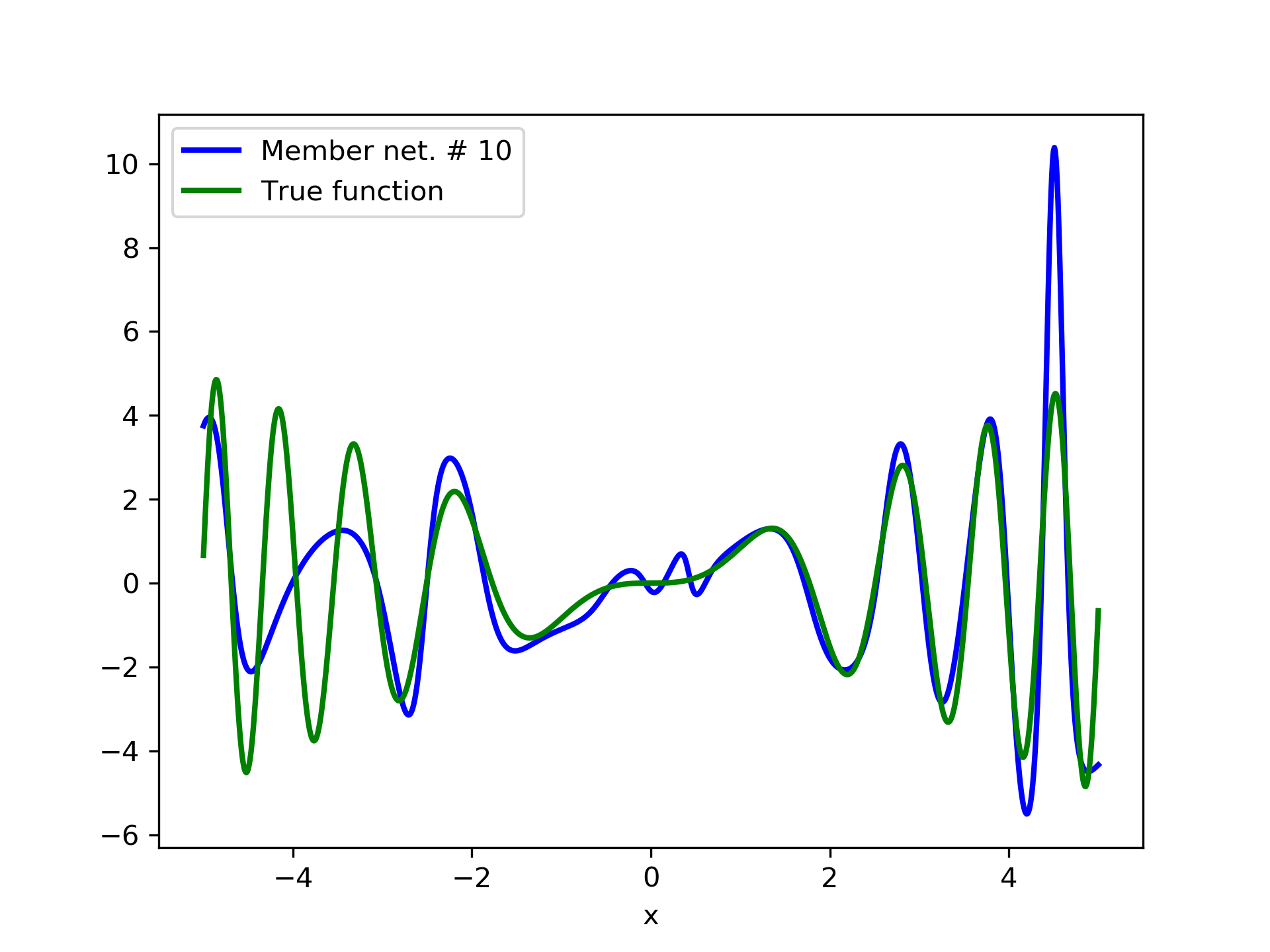}
\end{subfigure}\hfill
\begin{subfigure}{0.33\textwidth}
    \centering
    \includegraphics[width=\textwidth]{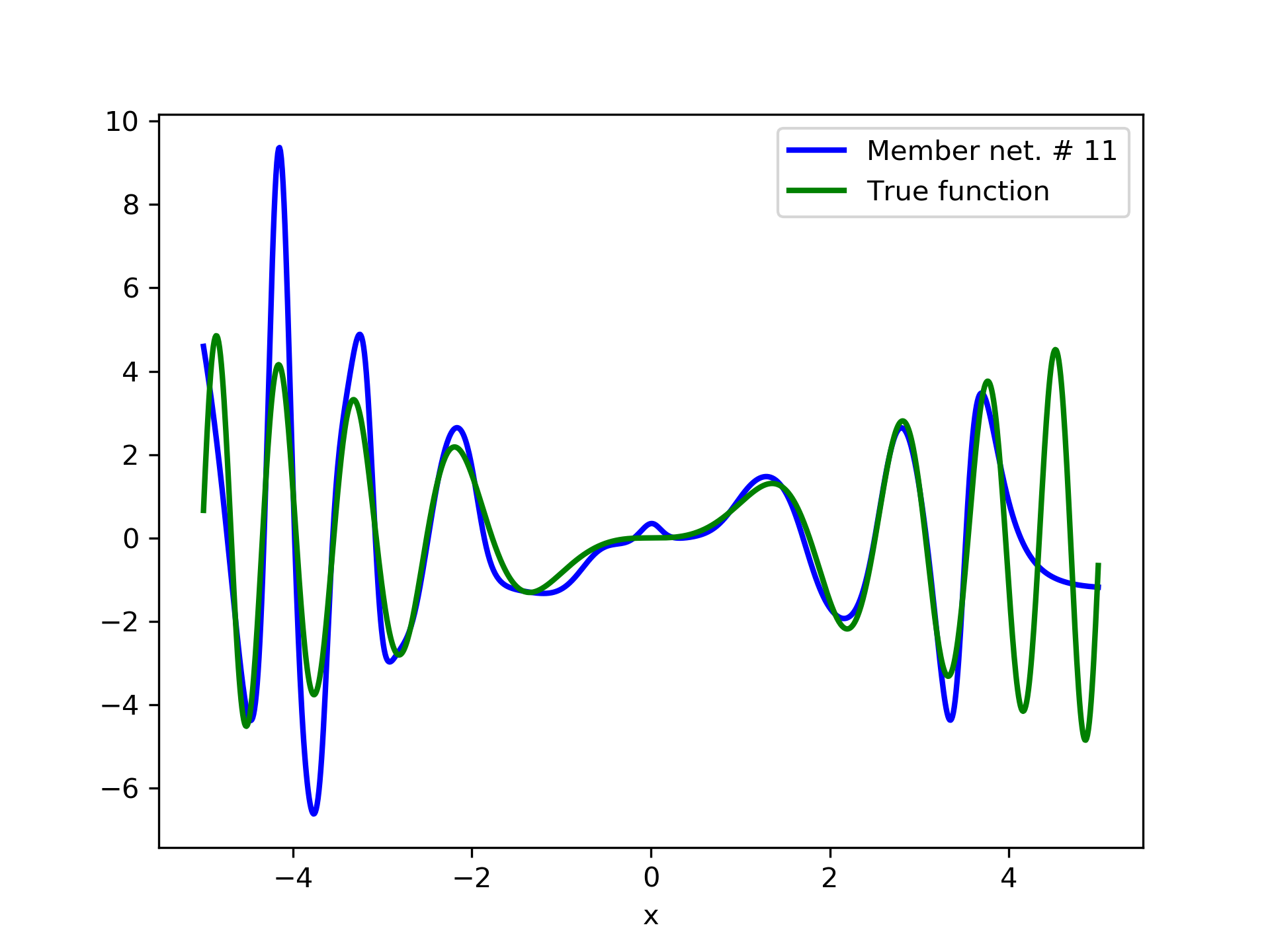}
\end{subfigure}\hfill
\begin{subfigure}{0.33\textwidth}
    \centering
    \includegraphics[width=\textwidth]{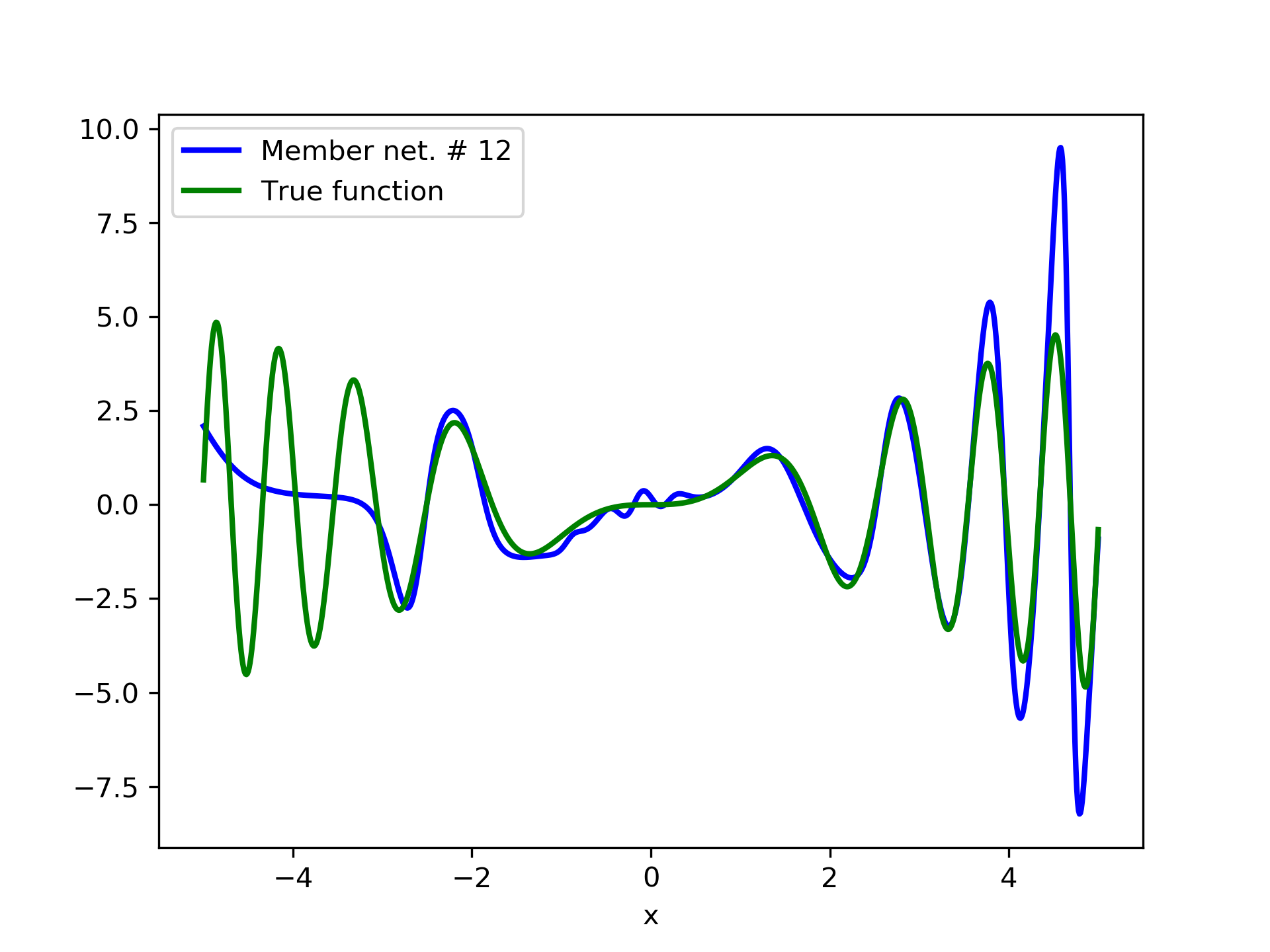}
\end{subfigure}
\begin{subfigure}{0.33\textwidth}
    \centering
    \includegraphics[width=\textwidth]{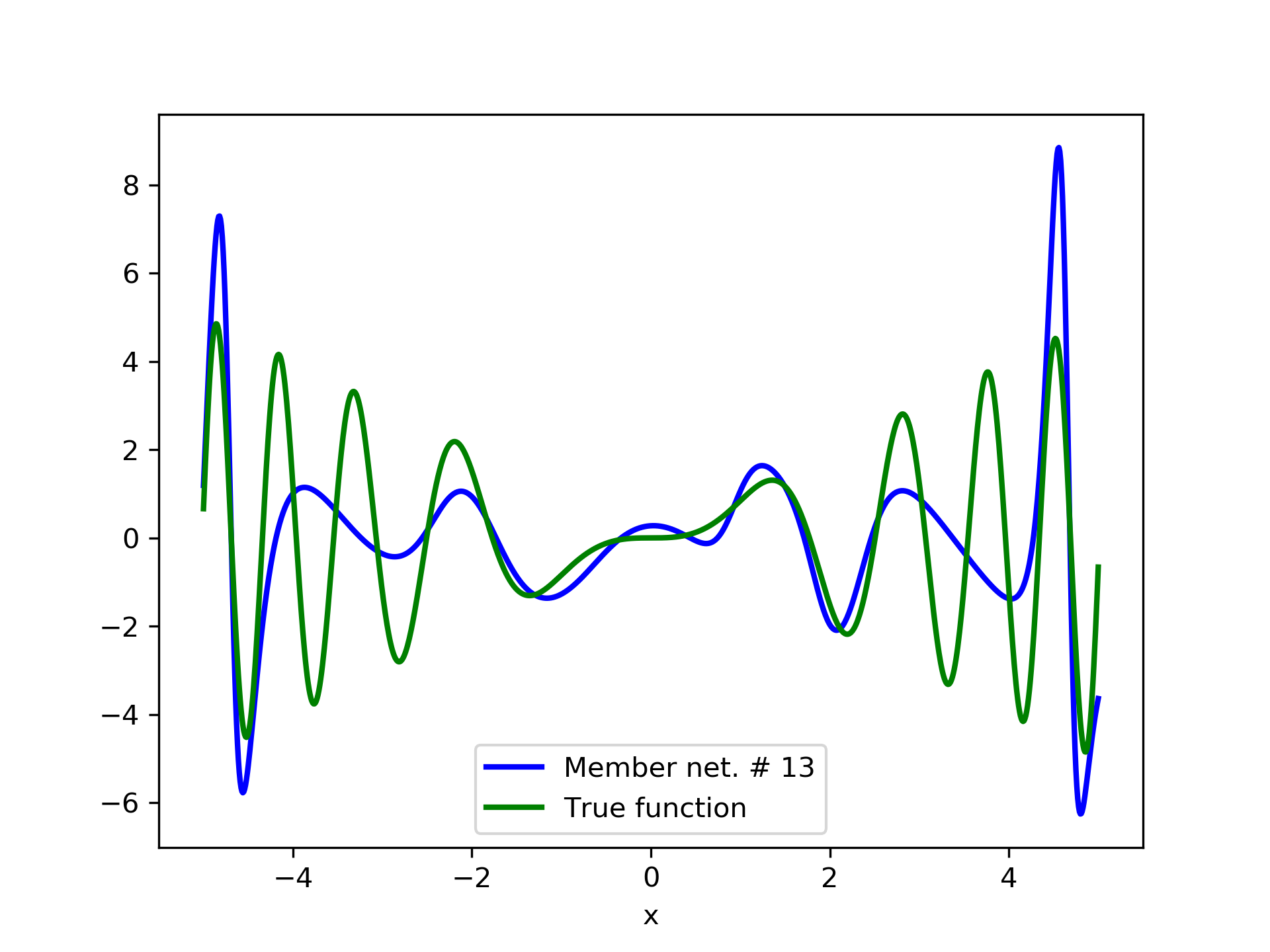}
\end{subfigure}\hfill
\begin{subfigure}{0.33\textwidth}
    \centering
    \includegraphics[width=\textwidth]{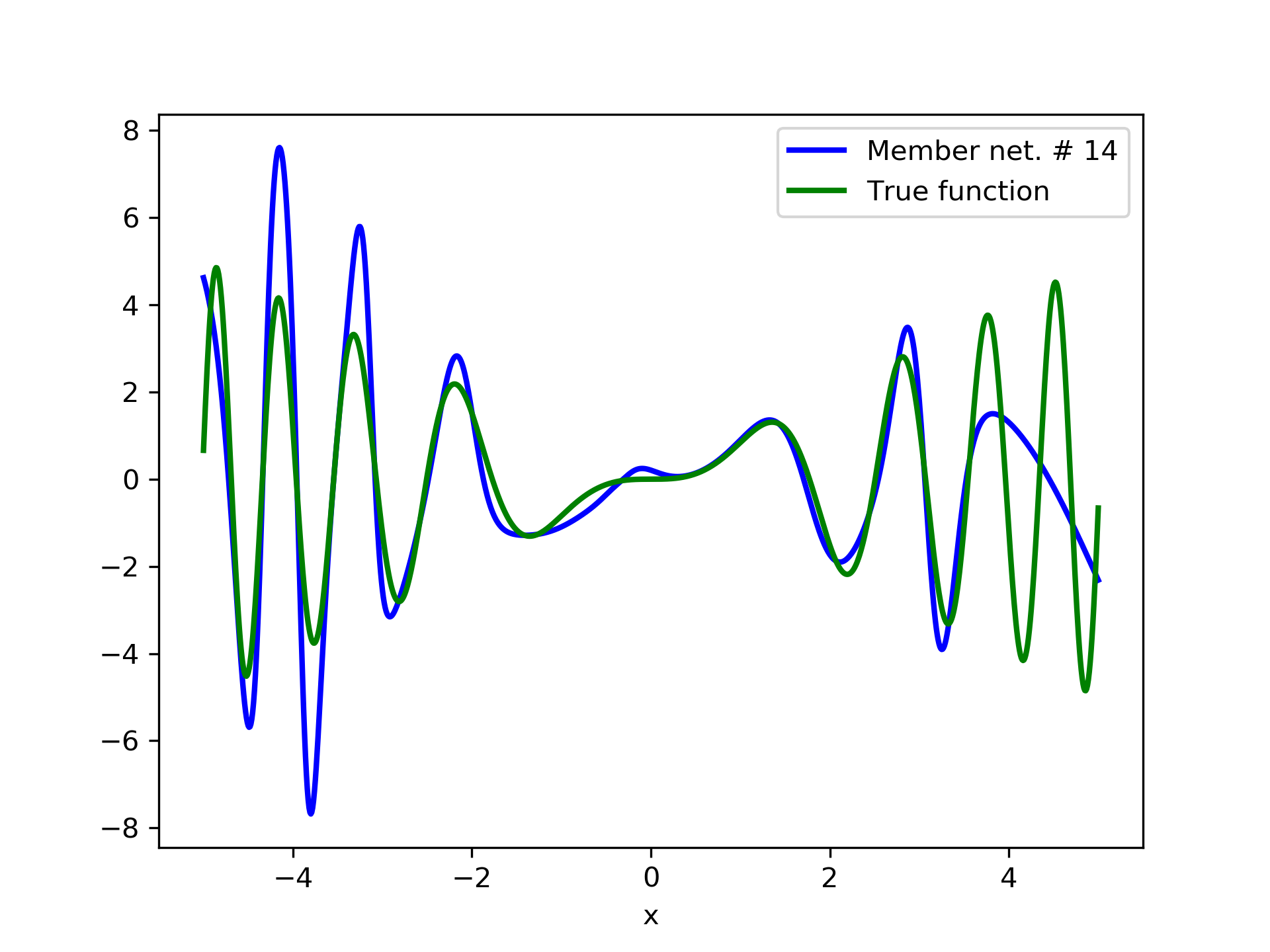}
\end{subfigure}\hfill
\begin{subfigure}{0.33\textwidth}
    \centering
    \includegraphics[width=\textwidth]{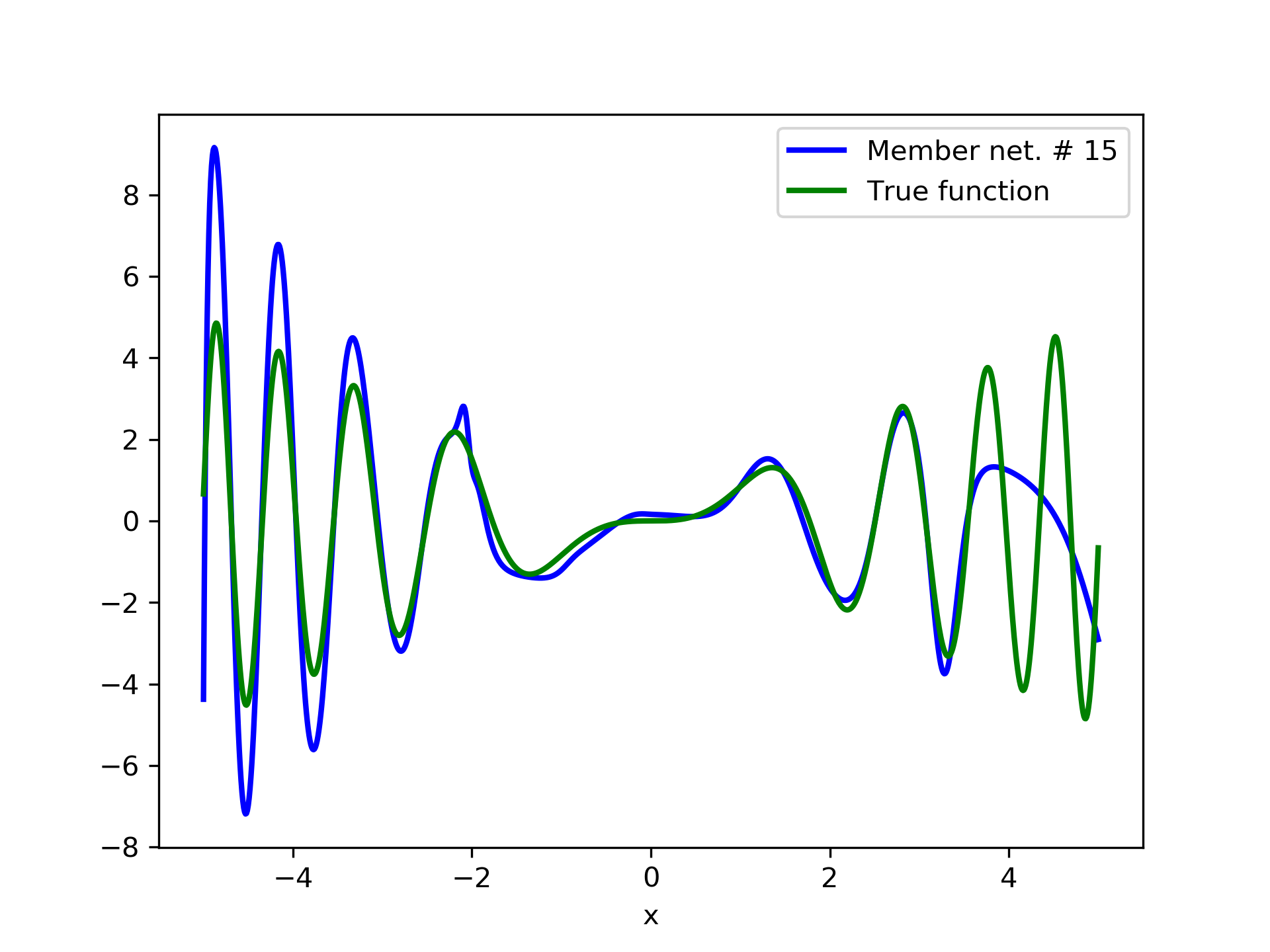}
\end{subfigure}
\caption{Plot of the ANN members related to table \ref{tab:noisy}, along with the plot of $f_1(x)$.}
\label{fig:noise_learned_member}
\end{figure}

\begin{figure}[h]
\centering
\begin{subfigure}{0.33\textwidth}
    \centering
    \includegraphics[width=\textwidth]{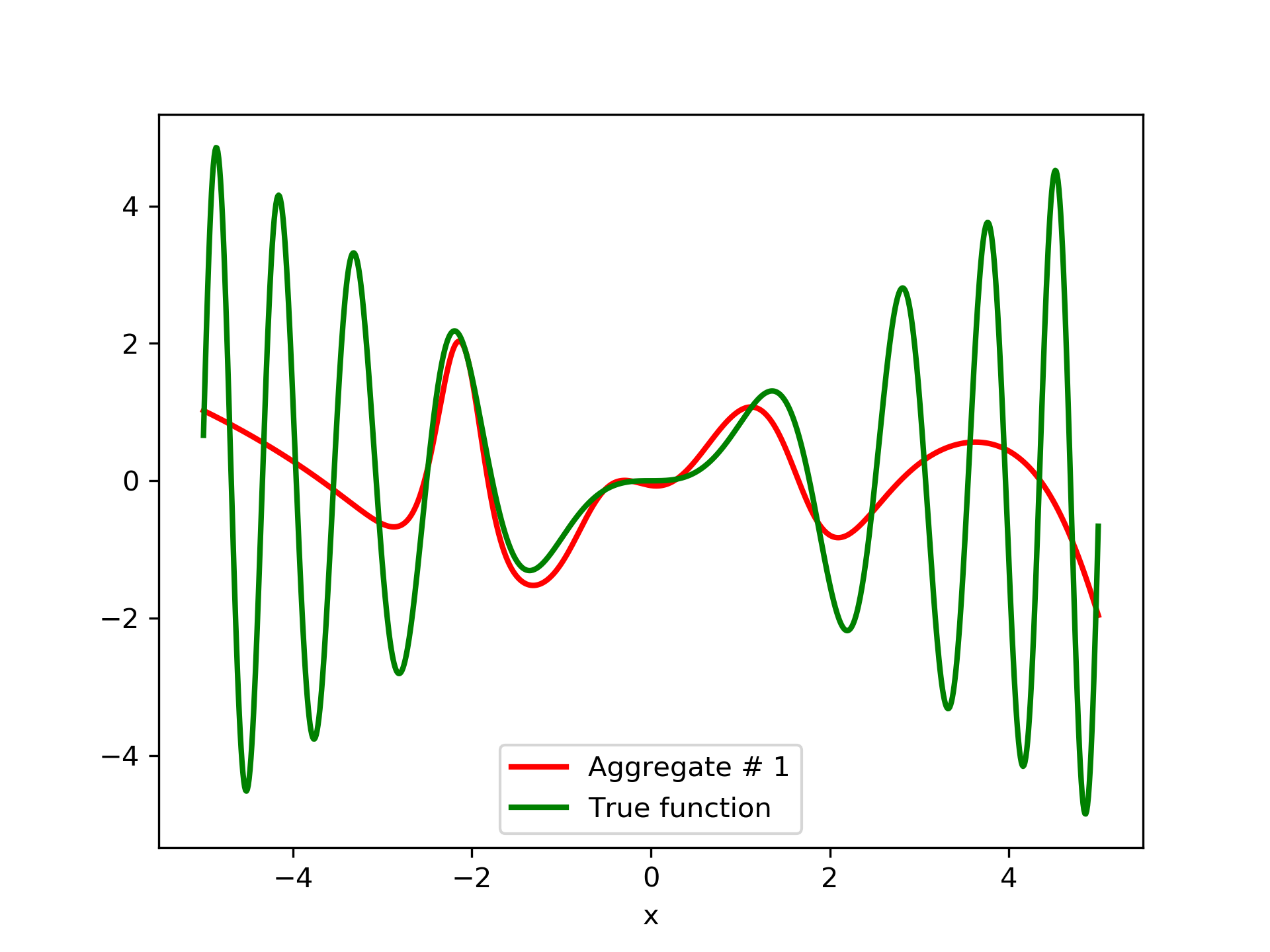}
\end{subfigure}\hfill
\begin{subfigure}{0.33\textwidth}
    \centering
    \includegraphics[width=\textwidth]{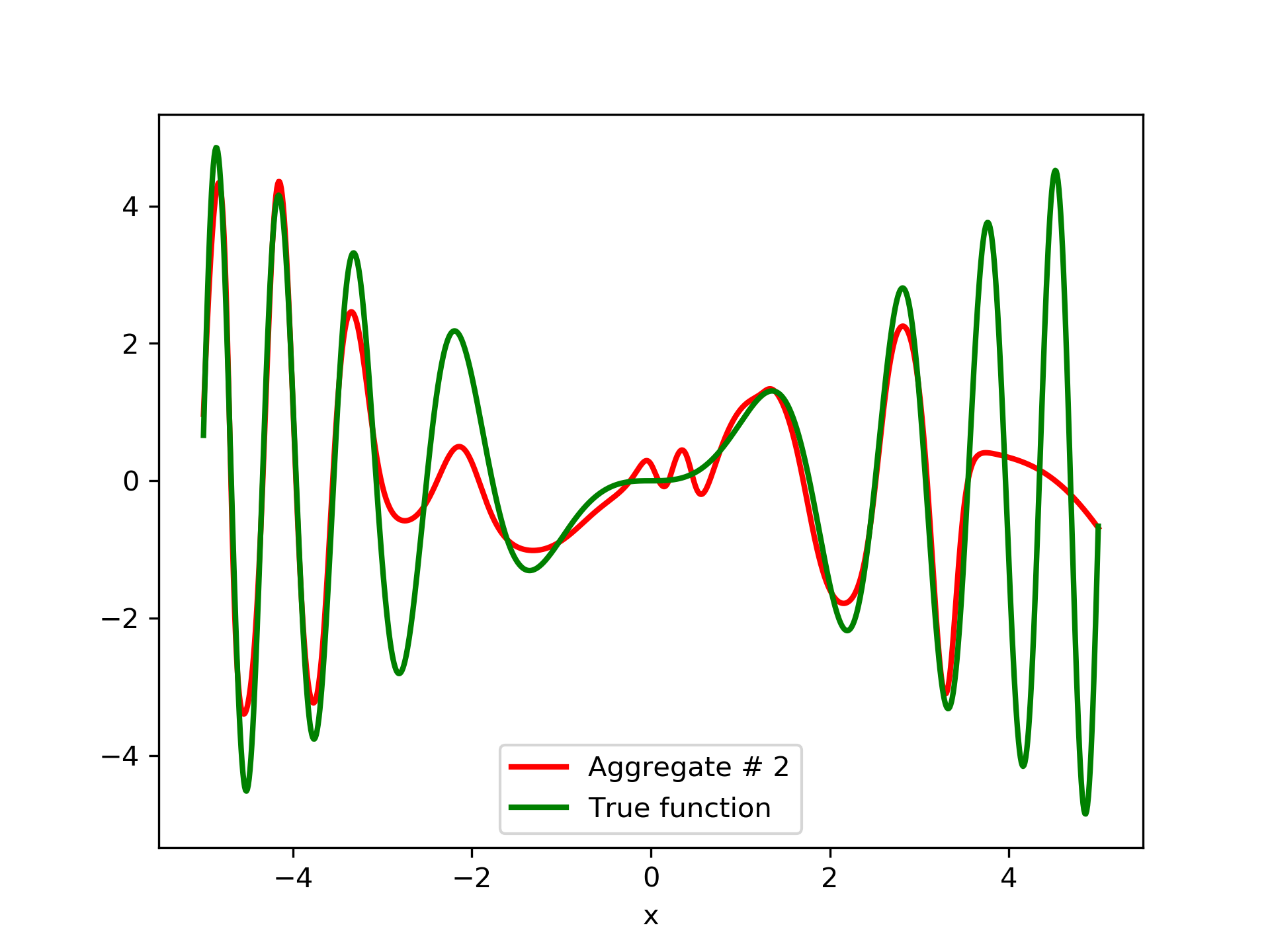}
\end{subfigure}\hfill
\begin{subfigure}{0.33\textwidth}
    \centering
    \includegraphics[width=\textwidth]{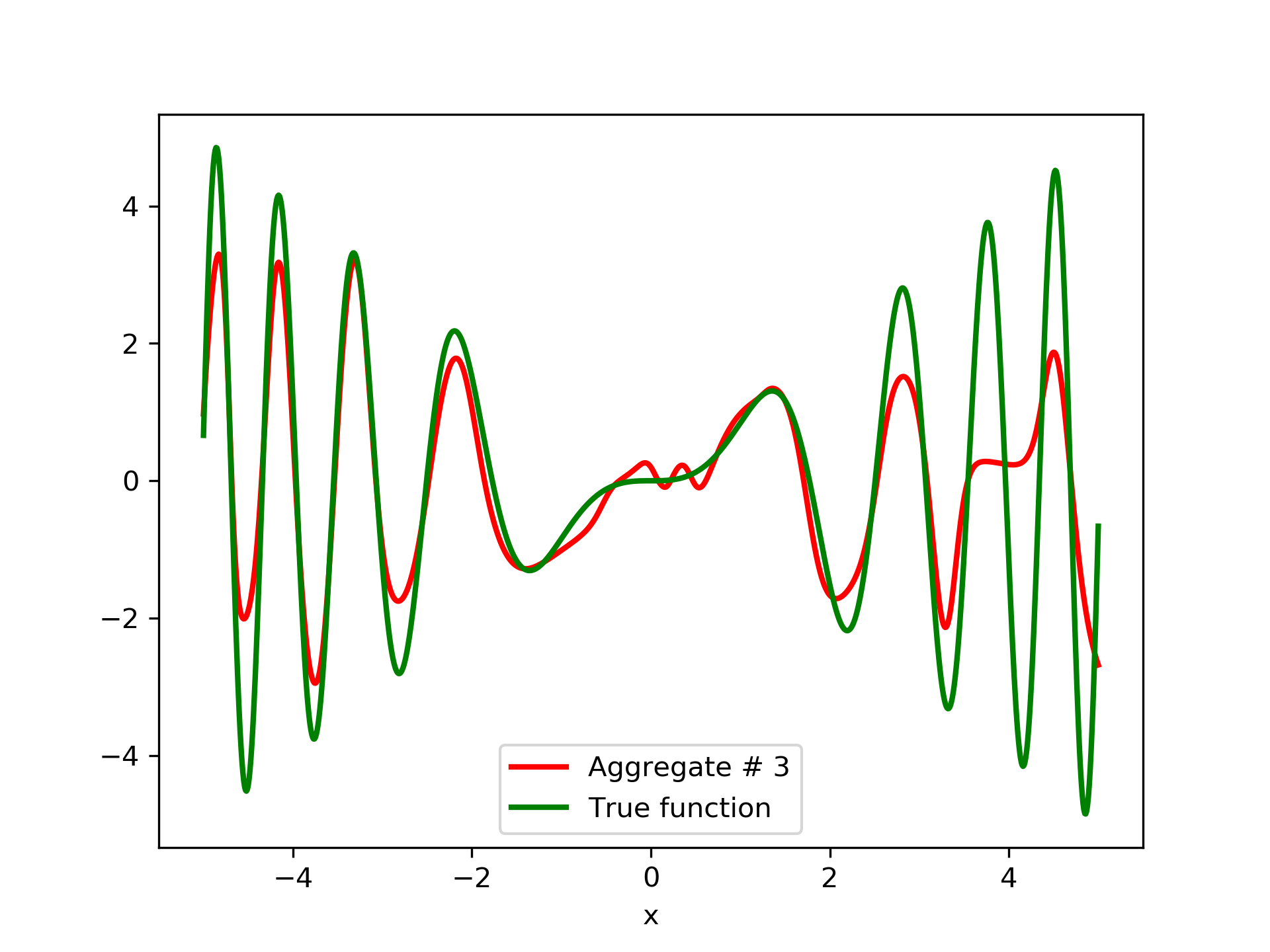}
\end{subfigure}\
\begin{subfigure}{0.33\textwidth}
    \centering
    \includegraphics[width=\textwidth]{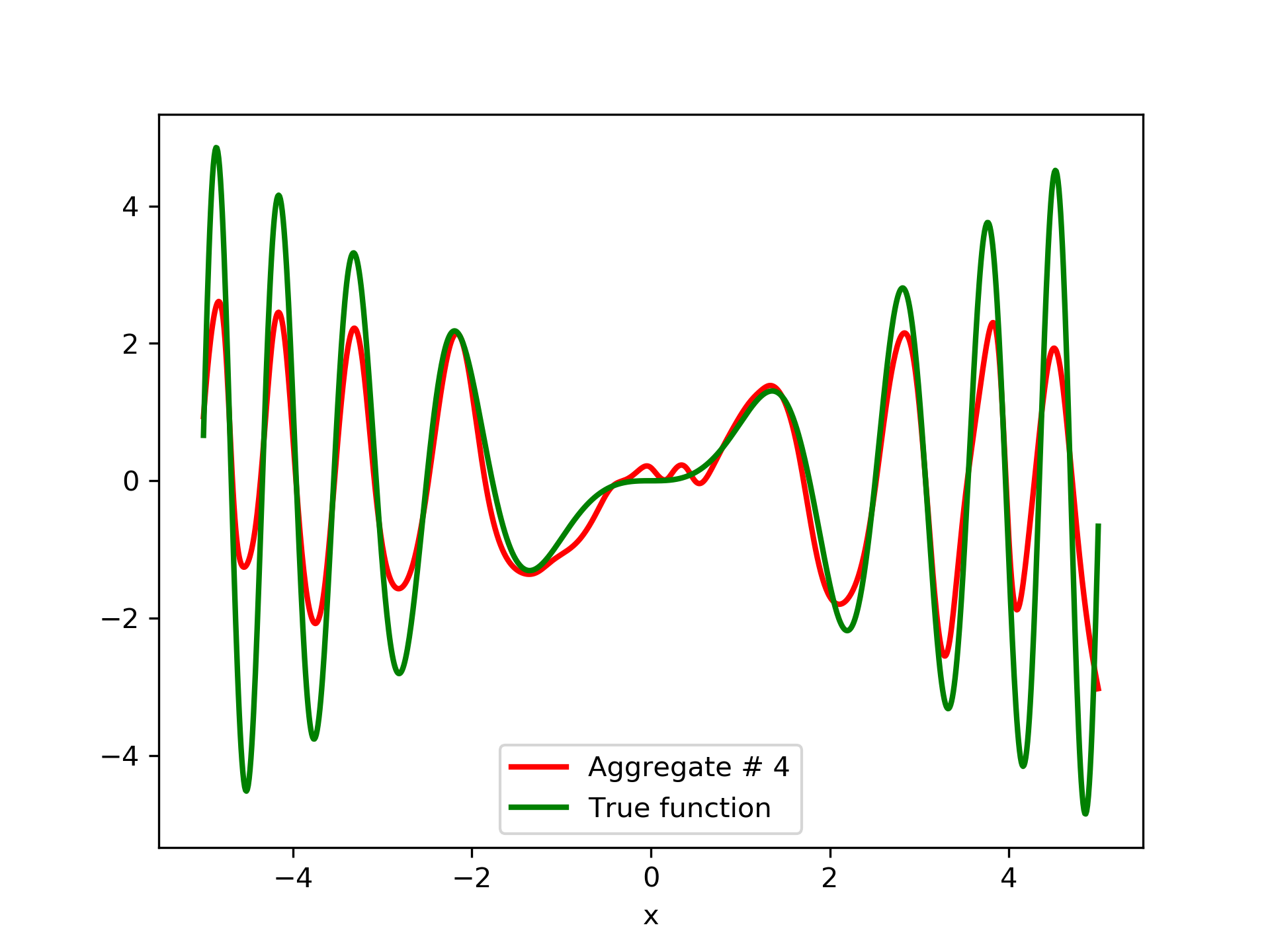}
\end{subfigure}\hfill
\begin{subfigure}{0.33\textwidth}
    \centering
    \includegraphics[width=\textwidth]{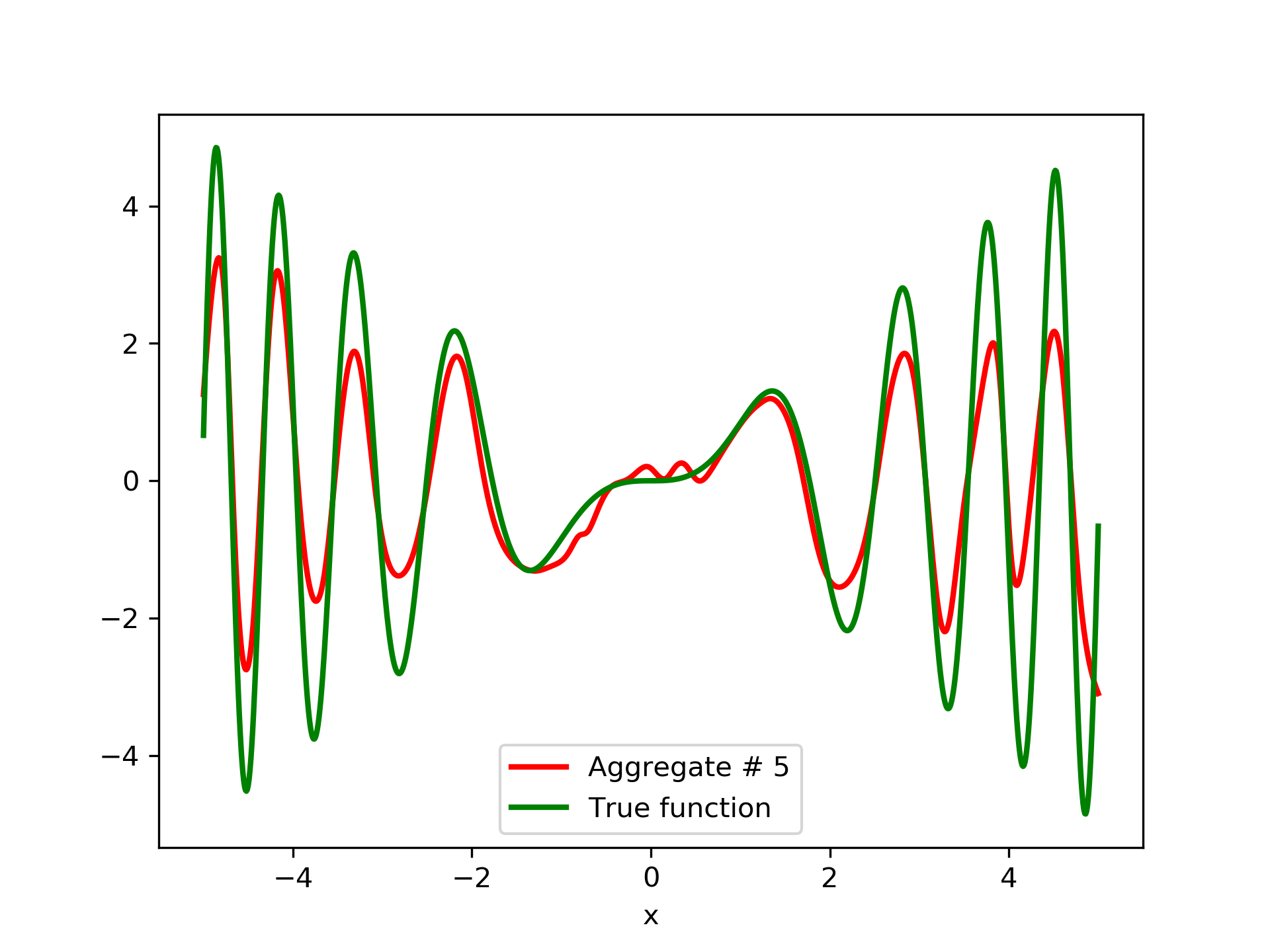}
\end{subfigure}\hfill
\begin{subfigure}{0.33\textwidth}
    \centering
    \includegraphics[width=\textwidth]{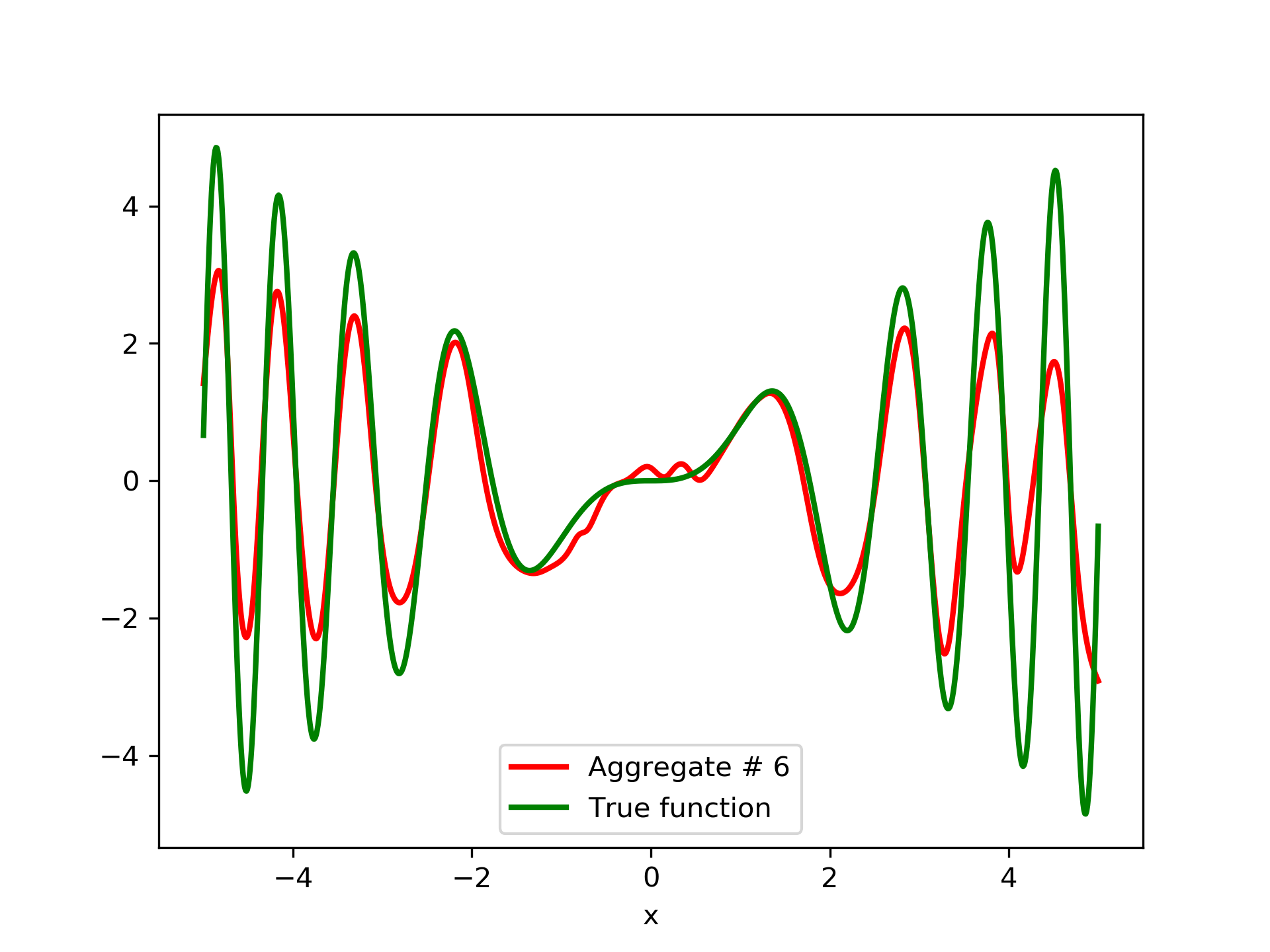}
\end{subfigure}
\begin{subfigure}{0.33\textwidth}
    \centering
    \includegraphics[width=\textwidth]{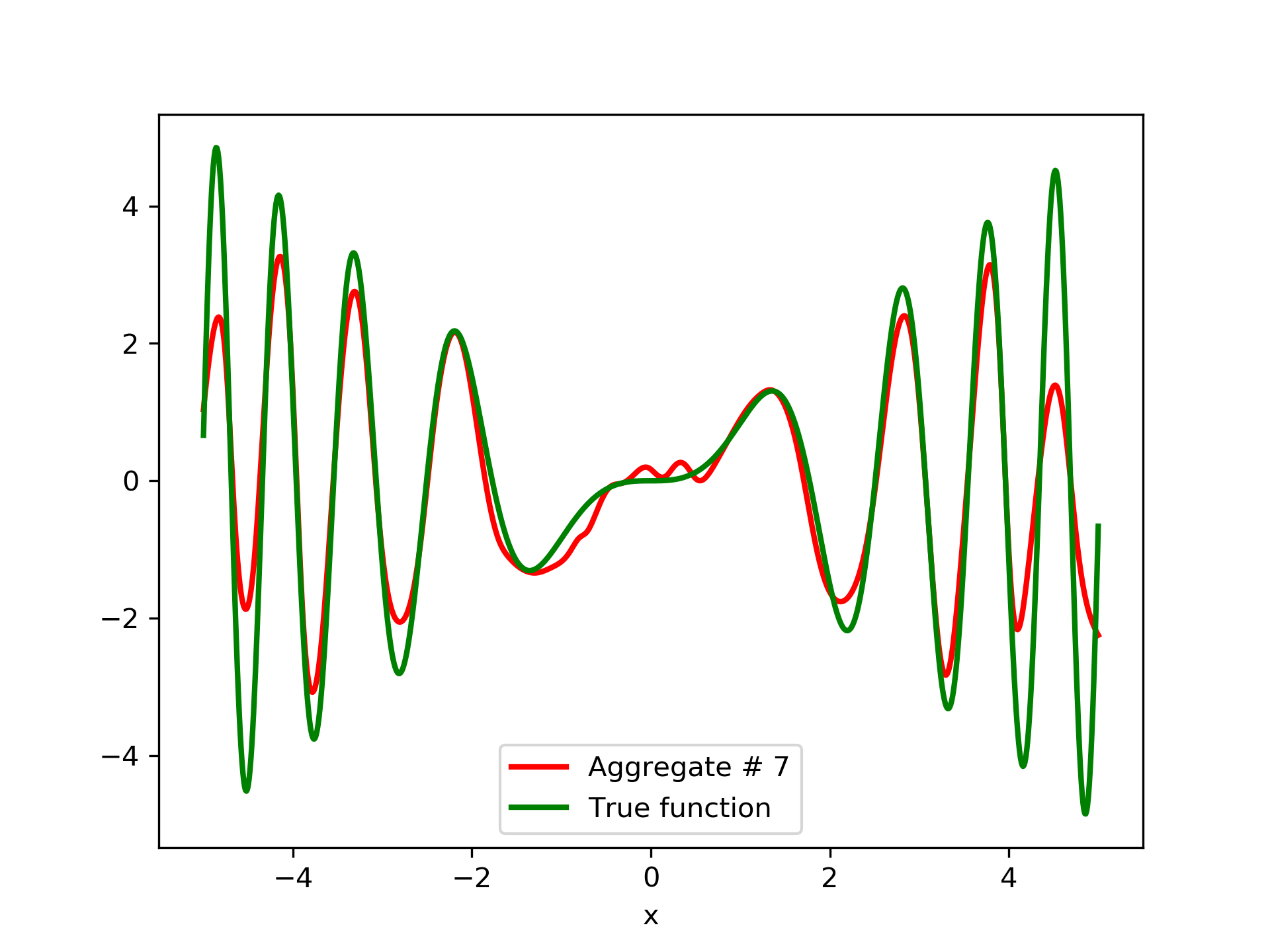}
\end{subfigure}\hfill
\begin{subfigure}{0.33\textwidth}
    \centering
    \includegraphics[width=\textwidth]{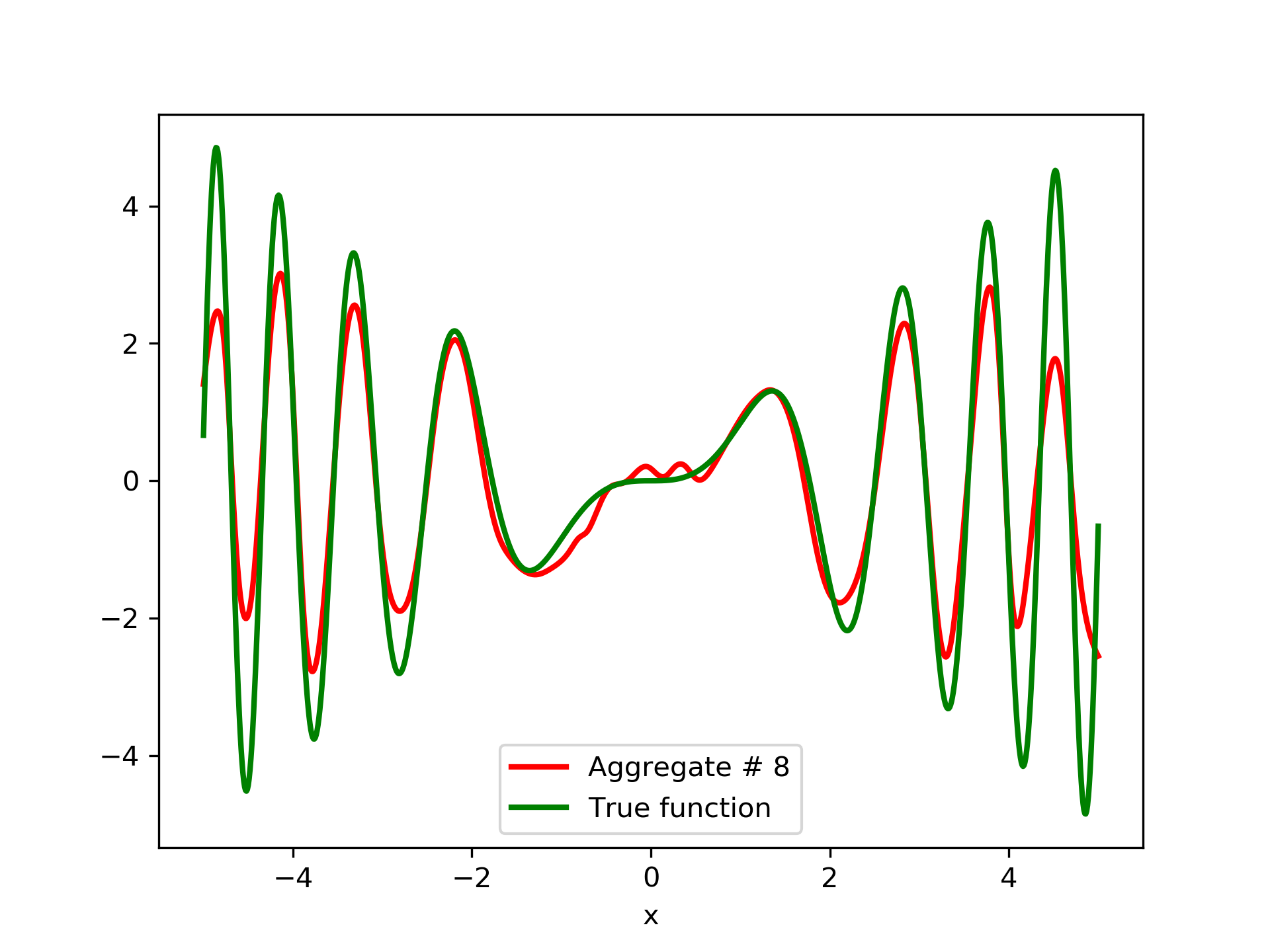}
\end{subfigure}\hfill
\begin{subfigure}{0.33\textwidth}
    \centering
    \includegraphics[width=\textwidth]{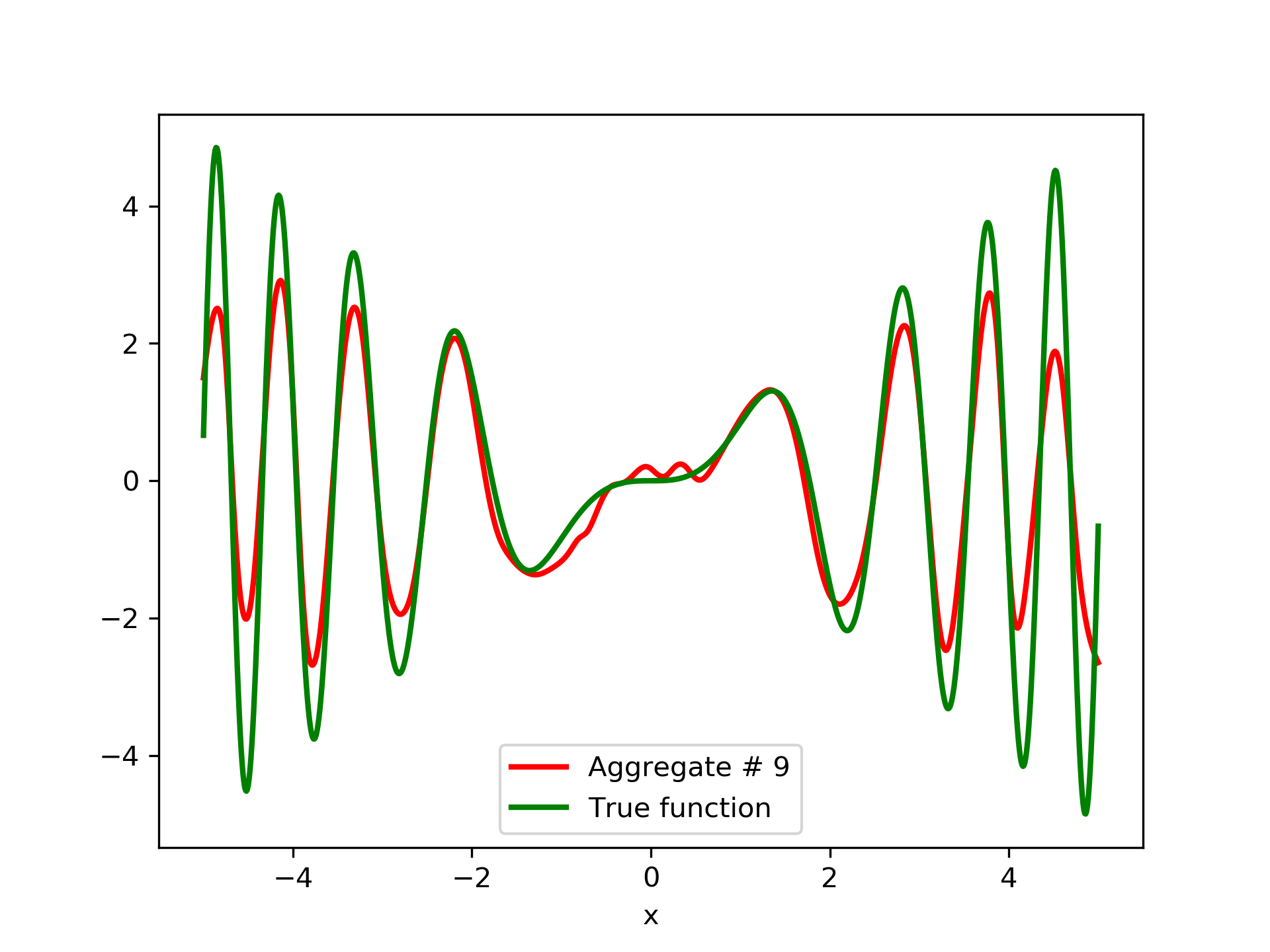}
\end{subfigure}
\begin{subfigure}{0.33\textwidth}
    \centering
    \includegraphics[width=\textwidth]{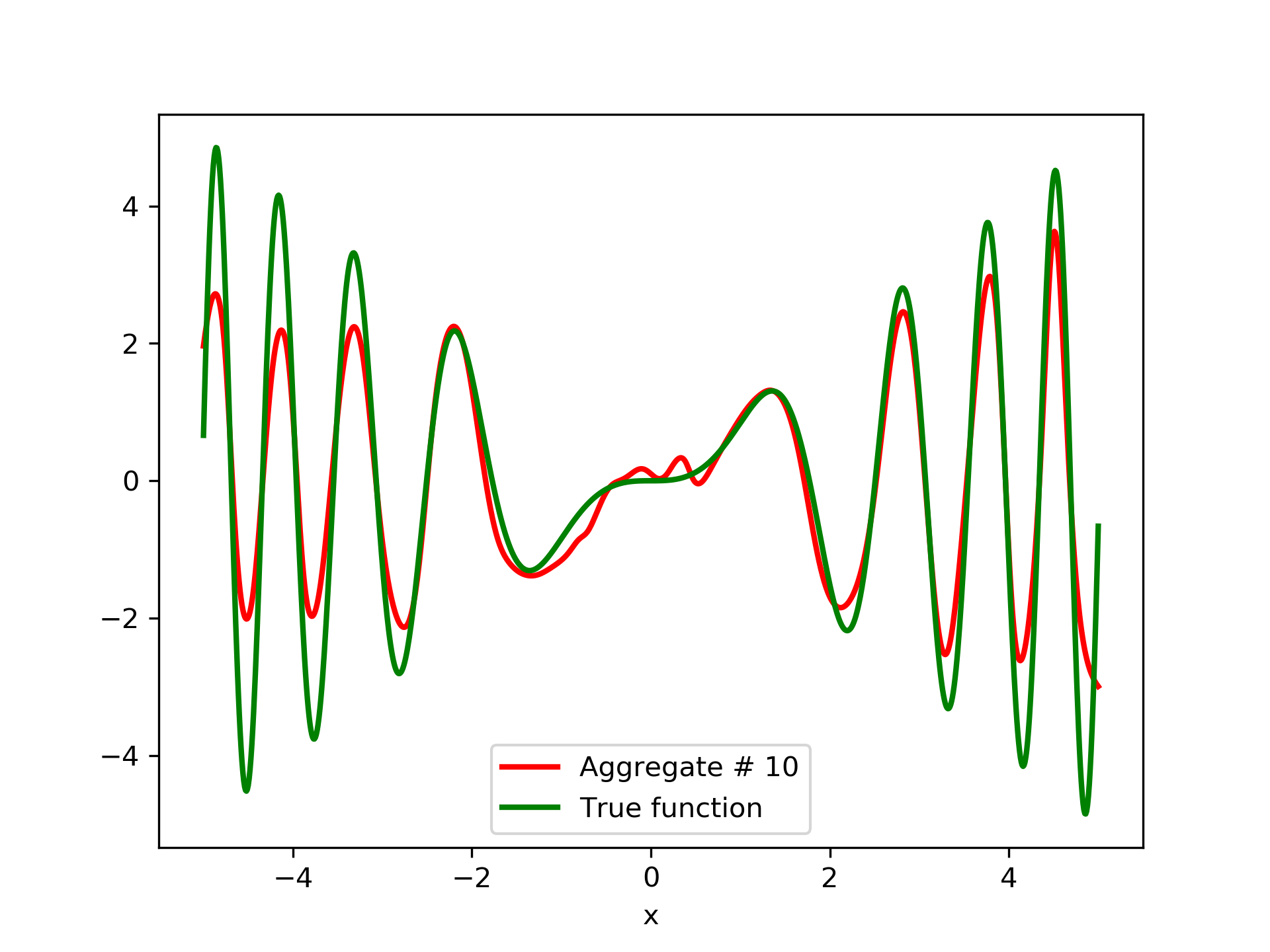}
\end{subfigure}\hfill
\begin{subfigure}{0.33\textwidth}
    \centering
    \includegraphics[width=\textwidth]{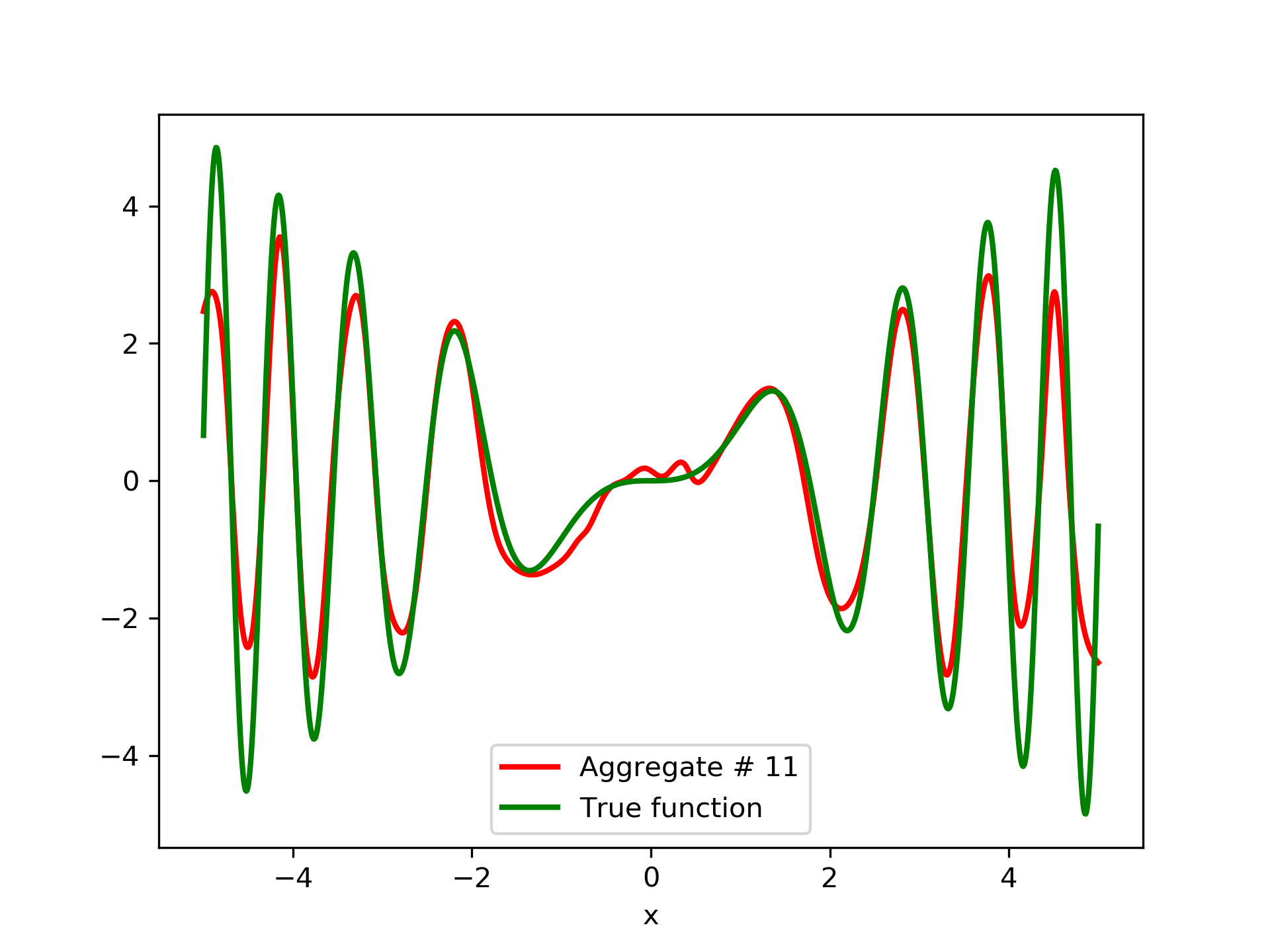}
\end{subfigure}\hfill
\begin{subfigure}{0.33\textwidth}
    \centering
    \includegraphics[width=\textwidth]{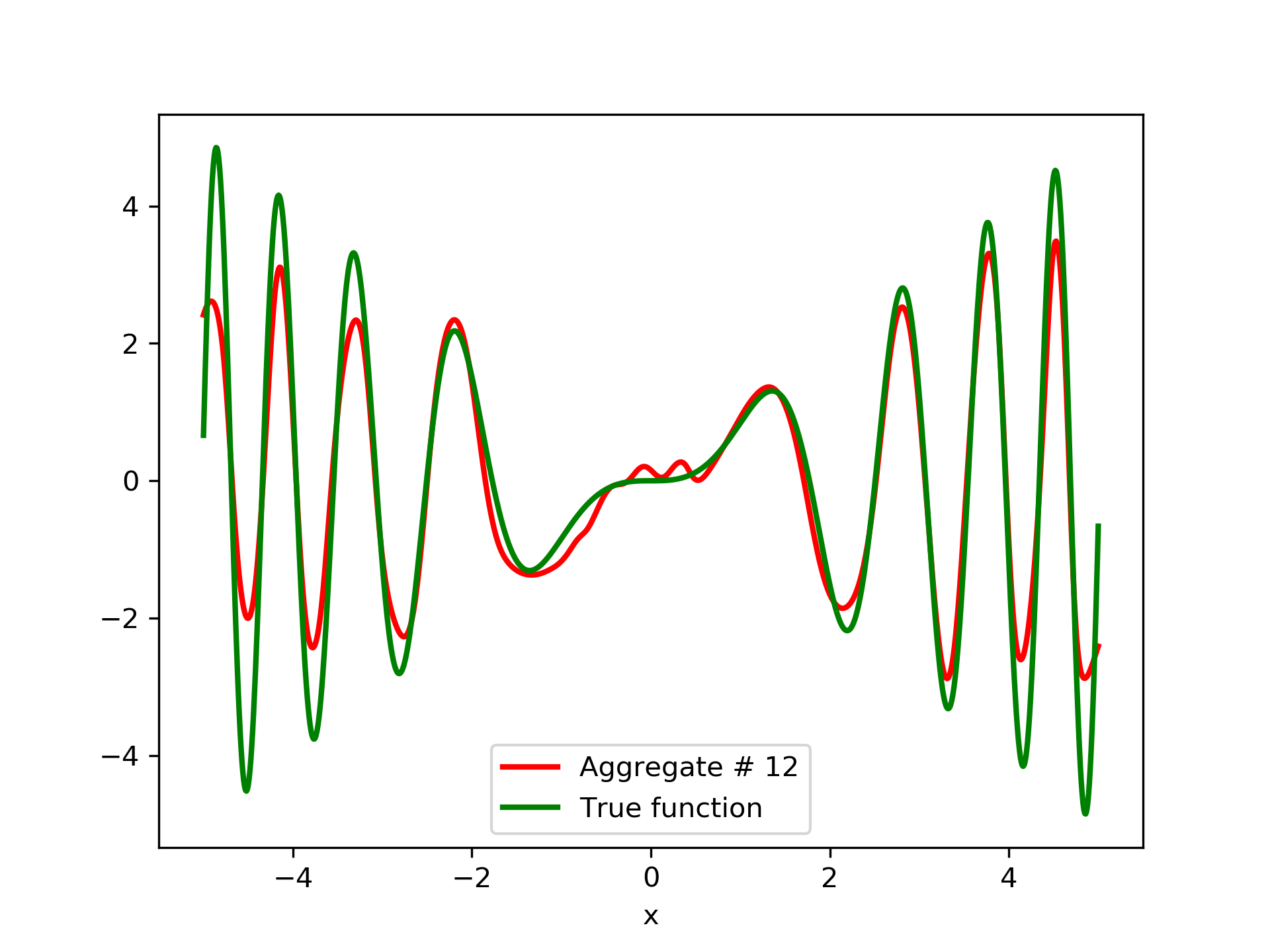}
\end{subfigure}
\begin{subfigure}{0.33\textwidth}
    \centering
    \includegraphics[width=\textwidth]{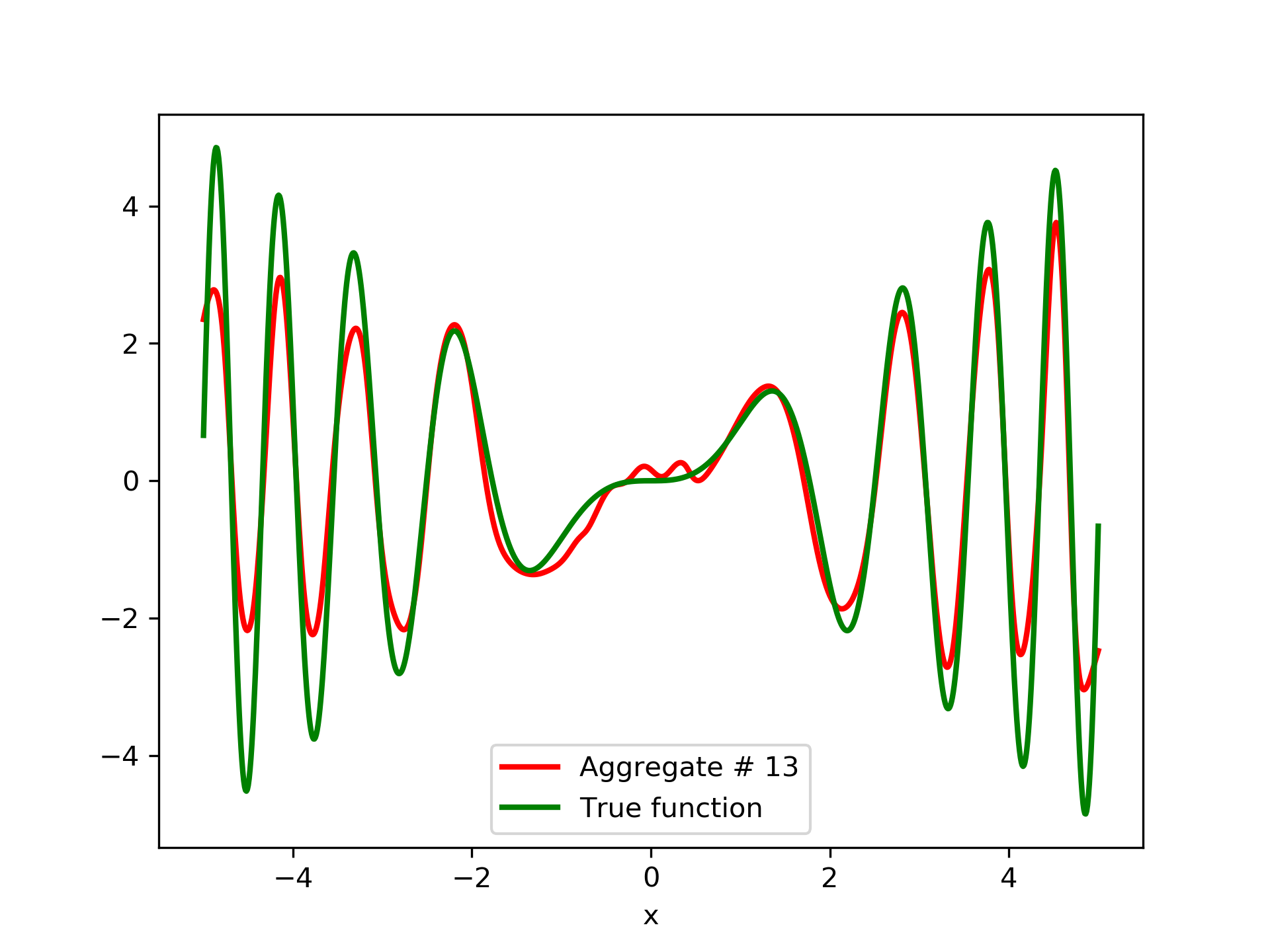}
\end{subfigure}\hfill
\begin{subfigure}{0.33\textwidth}
    \centering
    \includegraphics[width=\textwidth]{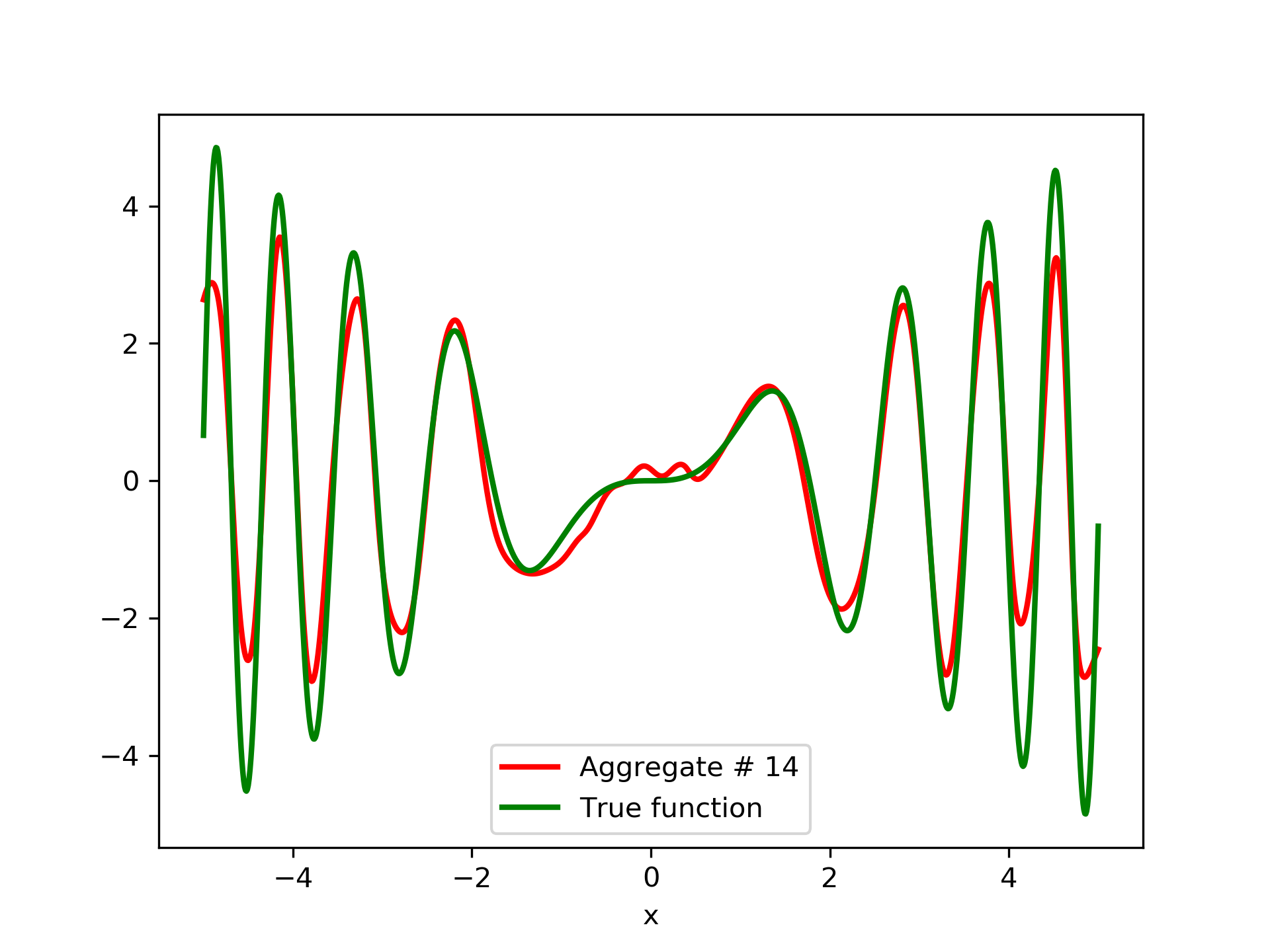}
\end{subfigure}\hfill
\begin{subfigure}{0.33\textwidth}
    \centering
    \includegraphics[width=\textwidth]{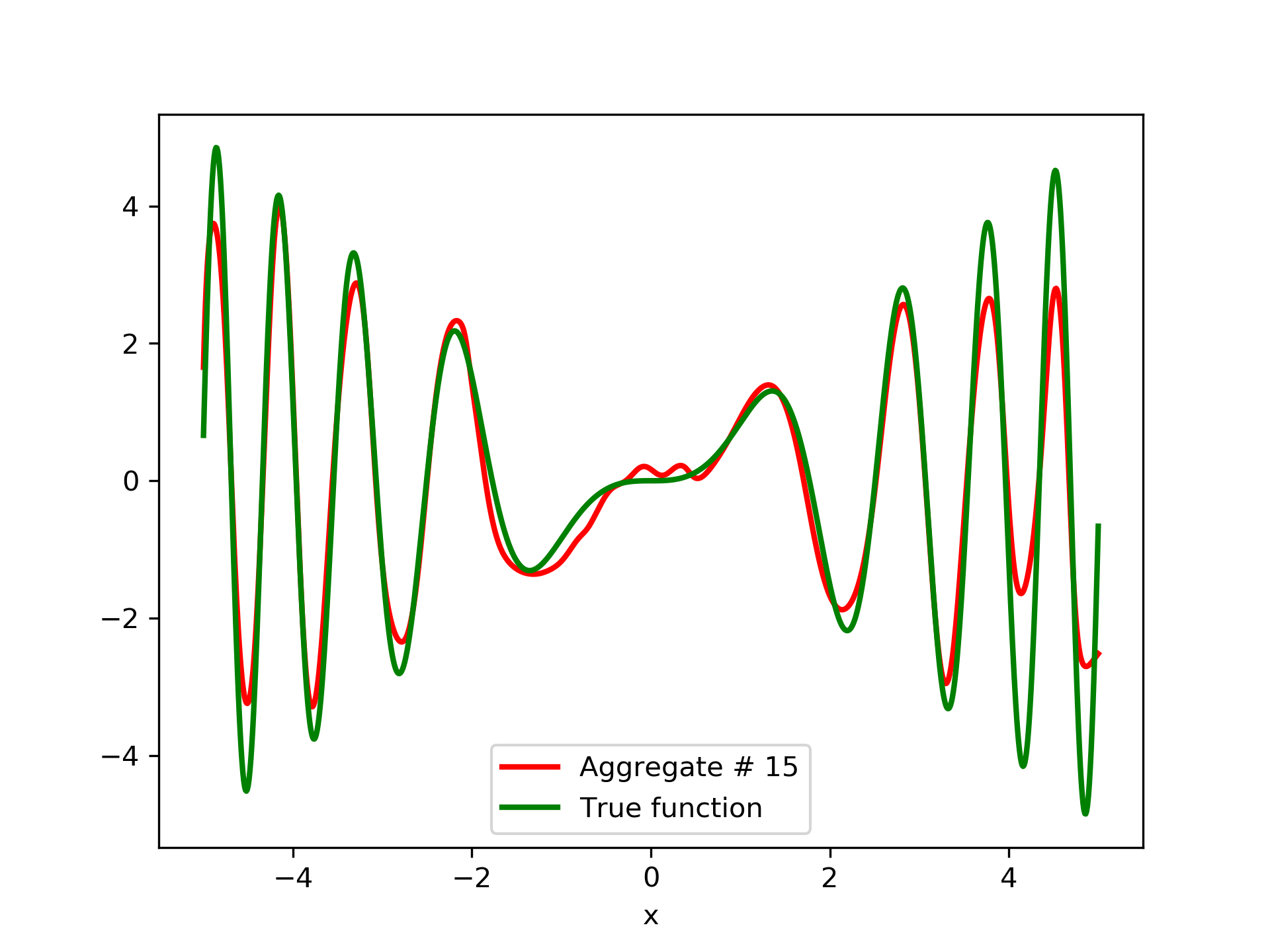}
\end{subfigure}
\caption{Plot of the aggregate ANNs related to table \ref{tab:noisy}, along with the plot of $f_1(x)$.}
\label{fig:noise_learned_agg}
\end{figure}

In this case we have experimented with noisy data.
Model function: $f_1(x)$ with $x \in[-5,5]$.
We construct a set $D$ containing $M_D = 1001$ equidistant points, $x_i = -5+\frac{i-1}{100}$. The training set $T_r$ contains $M_r = 101$ equidistant points, $ z_j = -5+\frac{j-1}{10}=x_{10j-9}$ and the target values are contaminated with white normal noise ($0.3N(0,1)$).
The test set is left intact, i.e. without any added noise.
The ensemble comprises 15 different networks.
In \mbox{fig. \ref{fig:noise_learned_member}}, the plots of $f_1(x)$ and of the trained member networks are displayed. The aggregate ANN's are plotted in fig. \ref{fig:noise_learned_agg}. 
The  noisy data are depicted in fig. \ref{fig:noisy}, where the dots correspond to the noisy training observations, while the solid line is a plot of $f_1(x)$.

Detailed results are listed in table \ref{tab:noisy}. 
As expected, the aggregate training error decreases by adding more networks and similar is the behavior of the test error. From  column ``AG. MSE/TE'' of table \ref{tab:noisy}, we note that the aggregate network generalizes extremely well. 
Note, model No. 12 has the lowest training MSE among all models which is 1.43211. The training MSE of the final aggregate is 0.45040, which is $68\%$ lower than that of model No. 12. 
\clearpage
\begin{table}[h]
\caption{Experiment with $f_1(x)$ with $x\in[-5,5]$ and a training set of 101 equidistant  points with added normal white noise. Regularization factor: $\nu_{reg}=0.1$.}
\begin{center}
\begin{tabular}{|c|c|c|c|c|c|c|c|}
\hline
     ANN & Nodes & Type & MSE & $\beta$ &  AG. MSE &  AG. MSE/TE &  $a$ \\
\hline
 1 &       24 &  tanh     &   3.42492 &  0.00000 &  3.42492 & 3.60286 &  0.06919 \\
 \hline
   2 &     25 &   sigmoid &  12.89758 &  0.81382 &  2.90180 & 3.04288 &  0.01583 \\
   \hline
   3 &     27 &   sigmoid &   4.71427 &  0.62760 &  1.91684 & 1.92987 &  0.05045 \\
   \hline
   4 &     23 &  softplus &   5.21665 &  0.98091 &  1.91559 & 1.93207 &  0.00264 \\
   \hline
   5 &     25 &   sigmoid &  10.43508 &  0.92285 &  1.85563 & 1.88489 &  0.01155 \\
   \hline
   6 &     29 &      tanh &   2.28159 &  0.55624 &  1.10990 & 1.12316 &  0.11939 \\
   \hline
   7 &     26 &  softplus &   7.02311 &  0.96502 &  1.10213 & 1.11369 &  0.00975 \\
   \hline
   8 &     24 &   sigmoid &   5.98359 &  0.95641 &  1.09196 & 1.10721 &  0.01271 \\
   \hline
   9 &     25 &      tanh &   3.65851 &  0.86828 &  1.03151 & 1.05286 &  0.04422 \\
   \hline
  10 &     25 &      tanh &   2.72174 &  0.74489 &  0.80691 & 0.79448 &  0.11498 \\
  \hline
  11 &     25 &  softplus &  12.74256 &  0.96355 &  0.78981 & 0.77719 &  0.01705 \\
  \hline
  12 &     27 &      tanh &   1.43211 &  0.66732 &  0.57738 & 0.58926 &  0.23319 \\
  \hline
  13 &     26 &  softplus &   9.88673 &  0.97901 &  0.57310 & 0.57708 &  0.01503 \\
  \hline
  14 &     25 &      tanh &   2.31302 &  0.84593 &  0.51341 & 0.51266 &  0.13040 \\
  \hline
  15 &     24 &      tanh &   2.36287 &  0.84637 &  0.45040 & 0.45868 &  0.15363 \\
\hline
\end{tabular}
\end{center}
\label{tab:noisy}
\end{table}
\section{Summary and Conclusions}\label{SUMMARY}
In this article we have presented an ensemble method that is free of the multicollinearity issue, and without having to resort to the ad-hoc choice of uniform weights.
Instead, we build an aggregate network, that does take in account the relative importance of the member networks. The aggregate network is incrementally built by blending it at every stage with a newly trained network under a negative correlation constraint. The blending is a convex linear combination as suggested in \citet{Breiman:1996}, in order to maintain  high generalization performance.
Indeed, this may be confirmed by inspecting the \mbox{``AG. MSE'' and ``AG.MES/TE''} columns in tables \ref{EXPF1}, \ref{EXPF1W}, \ref{tab:f3} and \ref{tab:noisy}. 
The experiment with noisy data, summarized in table  \ref{tab:noisy}, indicates that the method is robust and keeps on delivering quality generalization, as well as partial noise filtering.

Ensemble methods are useful in many applications. They have been used  with success
in econometrics and statistics for quite some time, \citet{Granger:1989}, \citet{Wallis:2011}. Since ensembles may contain many small networks, which are capable of being trained over datasets limited in size without over-fitting, they seem to be suitable for the interesting ``Few-Shot Learning'' case,  \citet{Wang:2020}. \\
The present method has been developed with regression problems in mind. Further work is necessary to extend it properly for handling classification tasks as well. 
\section*{Acknowledgements}
This work has been supported by a donation to AI Systems Lab (AISL) by GS
Gives and the Office of Naval Research (ONR) under Grant No. N00014-18-1-2278.

\bibliographystyle{achemso}
\medskip
\bibliography{ref}

\providecommand{\latin}[1]{#1}
\makeatletter
\providecommand{\doi}
  {\begingroup\let\do\@makeother\dospecials
  \catcode`\{=1 \catcode`\}=2 \doi@aux}
\providecommand{\doi@aux}[1]{\endgroup\texttt{#1}}
\makeatother
\providecommand*\mcitethebibliography{\thebibliography}
\csname @ifundefined\endcsname{endmcitethebibliography}
  {\let\endmcitethebibliography\endthebibliography}{}
\begin{mcitethebibliography}{27}
\providecommand*\natexlab[1]{#1}
\providecommand*\mciteSetBstSublistMode[1]{}
\providecommand*\mciteSetBstMaxWidthForm[2]{}
\providecommand*\mciteBstWouldAddEndPuncttrue
  {\def\EndOfBibitem{\unskip.}}
\providecommand*\mciteBstWouldAddEndPunctfalse
  {\let\EndOfBibitem\relax}
\providecommand*\mciteSetBstMidEndSepPunct[3]{}
\providecommand*\mciteSetBstSublistLabelBeginEnd[3]{}
\providecommand*\EndOfBibitem{}
\mciteSetBstSublistMode{f}
\mciteSetBstMaxWidthForm{subitem}{(\alph{mcitesubitemcount})}
\mciteSetBstSublistLabelBeginEnd
  {\mcitemaxwidthsubitemform\space}
  {\relax}
  {\relax}

\bibitem[Wolpert(1992)]{Wolpert:1992}
Wolpert,~D.~H. Stacked Generalization. \emph{Neural Networks} \textbf{1992},
  \emph{5}, 241--259\relax
\mciteBstWouldAddEndPuncttrue
\mciteSetBstMidEndSepPunct{\mcitedefaultmidpunct}
{\mcitedefaultendpunct}{\mcitedefaultseppunct}\relax
\EndOfBibitem
\bibitem[Breiman(1996)]{Breiman:1996}
Breiman,~L. Stacked regressions. \emph{Machine Learning} \textbf{1996},
  \emph{24}, 49--64\relax
\mciteBstWouldAddEndPuncttrue
\mciteSetBstMidEndSepPunct{\mcitedefaultmidpunct}
{\mcitedefaultendpunct}{\mcitedefaultseppunct}\relax
\EndOfBibitem
\bibitem[Perrone and Cooper(1993)Perrone, and Cooper]{Perrone:1993}
Perrone,~M.~P.; Cooper,~L.~N. When Networks Disagree: Ensemble Methods for
  Hybrid Neural Networks. Artificial Neural Networks for Speech and Vision.
  1993; pp 126--142\relax
\mciteBstWouldAddEndPuncttrue
\mciteSetBstMidEndSepPunct{\mcitedefaultmidpunct}
{\mcitedefaultendpunct}{\mcitedefaultseppunct}\relax
\EndOfBibitem
\bibitem[Perrone(1993)]{Perrone_thesis:1993}
Perrone,~M.~P. Improving Regression Estimation: Averaging Methods for Variance
  Reduction with Extensions to General Convex Measure Optimization. Ph.D.\
  thesis, Physics Department, Brown University, Providence, RI, USA, 1993\relax
\mciteBstWouldAddEndPuncttrue
\mciteSetBstMidEndSepPunct{\mcitedefaultmidpunct}
{\mcitedefaultendpunct}{\mcitedefaultseppunct}\relax
\EndOfBibitem
\bibitem[Hashem and Schmeiser(1993)Hashem, and Schmeiser]{HashemDER:1993}
Hashem,~S.; Schmeiser,~B. Approximating a Function and its Derivatives Using
  {MSE}-Optimal Linear Combinations of Trained Feedforward Neural Networks. In
  Proceedings of the Joint Conference on Neural Networks. 1993; pp
  617--620\relax
\mciteBstWouldAddEndPuncttrue
\mciteSetBstMidEndSepPunct{\mcitedefaultmidpunct}
{\mcitedefaultendpunct}{\mcitedefaultseppunct}\relax
\EndOfBibitem
\bibitem[Hashem and Schmeiser(1995)Hashem, and Schmeiser]{Hashem:1995}
Hashem,~S.; Schmeiser,~B. Improving Model Accuracy Using Optimal Linear
  Combinations of Trained Neural Networks. \emph{IEEE Transactions on Neural
  Networks} \textbf{1995}, \emph{6}, 792--794\relax
\mciteBstWouldAddEndPuncttrue
\mciteSetBstMidEndSepPunct{\mcitedefaultmidpunct}
{\mcitedefaultendpunct}{\mcitedefaultseppunct}\relax
\EndOfBibitem
\bibitem[Hashem(1993)]{Hashem:1993}
Hashem,~S. Optimal Linear Combinations of Neural Networks. Ph.D.\ thesis,
  School of Industrial Engineering, Purdue University, West Lafayette, IN, USA,
  1993\relax
\mciteBstWouldAddEndPuncttrue
\mciteSetBstMidEndSepPunct{\mcitedefaultmidpunct}
{\mcitedefaultendpunct}{\mcitedefaultseppunct}\relax
\EndOfBibitem
\bibitem[Hashem(1996)]{Hashem:1996}
Hashem,~S. Effects of Collinearity on Combining Neural Networks.
  \emph{Connection Science} \textbf{1996}, \emph{8}, 315--336\relax
\mciteBstWouldAddEndPuncttrue
\mciteSetBstMidEndSepPunct{\mcitedefaultmidpunct}
{\mcitedefaultendpunct}{\mcitedefaultseppunct}\relax
\EndOfBibitem
\bibitem[Hashem(1997)]{Hashem:1997}
Hashem,~S. Optimal Linear Combinations of Neural Networks. \emph{Neural
  Networks} \textbf{1997}, \emph{10}, 599--614\relax
\mciteBstWouldAddEndPuncttrue
\mciteSetBstMidEndSepPunct{\mcitedefaultmidpunct}
{\mcitedefaultendpunct}{\mcitedefaultseppunct}\relax
\EndOfBibitem
\bibitem[Krogh and Vedelsby(1994)Krogh, and Vedelsby]{Krogh:1994}
Krogh,~A.; Vedelsby,~J. Neural Network Ensembles, Cross Validation and Active
  Learning. Proceedings of the 7th International Conference on Neural
  Information Processing Systems. Cambridge, MA, USA, 1994; pp 231--238\relax
\mciteBstWouldAddEndPuncttrue
\mciteSetBstMidEndSepPunct{\mcitedefaultmidpunct}
{\mcitedefaultendpunct}{\mcitedefaultseppunct}\relax
\EndOfBibitem
\bibitem[Meir(1995)]{Meir:1995}
Meir,~R. {Bias, Variance and the Combination of Least Squares Estimators}. NIPS
  7. Cambridge MA, 1995\relax
\mciteBstWouldAddEndPuncttrue
\mciteSetBstMidEndSepPunct{\mcitedefaultmidpunct}
{\mcitedefaultendpunct}{\mcitedefaultseppunct}\relax
\EndOfBibitem
\bibitem[Hansen and Salamon(1990)Hansen, and Salamon]{Hansen:1990}
Hansen,~L.~K.; Salamon,~P. Neural network ensembles. \emph{IEEE Transactions on
  Pattern Analysis and Machine Intelligence} \textbf{1990}, \emph{12},
  993--1001\relax
\mciteBstWouldAddEndPuncttrue
\mciteSetBstMidEndSepPunct{\mcitedefaultmidpunct}
{\mcitedefaultendpunct}{\mcitedefaultseppunct}\relax
\EndOfBibitem
\bibitem[Leblanc and Tibshirani(1996)Leblanc, and Tibshirani]{Leblanc:1996}
Leblanc,~M.; Tibshirani,~R. Combining Estimates in Regression and
  Classification. \emph{Journal of the American Statistical Association}
  \textbf{1996}, \emph{91}, 1641--1650\relax
\mciteBstWouldAddEndPuncttrue
\mciteSetBstMidEndSepPunct{\mcitedefaultmidpunct}
{\mcitedefaultendpunct}{\mcitedefaultseppunct}\relax
\EndOfBibitem
\bibitem[Zhou \latin{et~al.}(2002)Zhou, Wu, and Tang]{ZHOU:2002}
Zhou,~Z.-H.; Wu,~J.; Tang,~W. Ensembling neural networks: Many could be better
  than all. \emph{Artificial Intelligence} \textbf{2002}, \emph{137}, 239 --
  263\relax
\mciteBstWouldAddEndPuncttrue
\mciteSetBstMidEndSepPunct{\mcitedefaultmidpunct}
{\mcitedefaultendpunct}{\mcitedefaultseppunct}\relax
\EndOfBibitem
\bibitem[Merz and Pazzani(1999)Merz, and Pazzani]{Merz:1999}
Merz,~C.~J.; Pazzani,~M.~J. A Principal Components Approach to Combining
  Regression Estimates. \emph{Machine Learning} \textbf{1999}, \emph{36},
  9--32\relax
\mciteBstWouldAddEndPuncttrue
\mciteSetBstMidEndSepPunct{\mcitedefaultmidpunct}
{\mcitedefaultendpunct}{\mcitedefaultseppunct}\relax
\EndOfBibitem
\bibitem[Liu and Yao(1999)Liu, and Yao]{LIU_YAO:1999}
Liu,~Y.; Yao,~X. Ensemble learning via negative correlation. \emph{Neural
  Networks} \textbf{1999}, \emph{12}, 1399--1404\relax
\mciteBstWouldAddEndPuncttrue
\mciteSetBstMidEndSepPunct{\mcitedefaultmidpunct}
{\mcitedefaultendpunct}{\mcitedefaultseppunct}\relax
\EndOfBibitem
\bibitem[Chen and Yao(2009)Chen, and Yao]{Chen_Yao:2009}
Chen,~H.; Yao,~X. Regularized Negative Correlation Learning for Neural Network
  Ensembles. \emph{IEEE Transactions on Neural Networks} \textbf{2009},
  \emph{20}, 1962--1979\relax
\mciteBstWouldAddEndPuncttrue
\mciteSetBstMidEndSepPunct{\mcitedefaultmidpunct}
{\mcitedefaultendpunct}{\mcitedefaultseppunct}\relax
\EndOfBibitem
\bibitem[Brown \latin{et~al.}(2005)Brown, Wyatt, and Ti\v{n}o]{Brown:2005}
Brown,~G.; Wyatt,~J.~L.; Ti\v{n}o,~P. Managing Diversity in Regression
  Ensembles. \emph{J. Mach. Learn. Res.} \textbf{2005}, \emph{6},
  1621--1650\relax
\mciteBstWouldAddEndPuncttrue
\mciteSetBstMidEndSepPunct{\mcitedefaultmidpunct}
{\mcitedefaultendpunct}{\mcitedefaultseppunct}\relax
\EndOfBibitem
\bibitem[Brown \latin{et~al.}(2005)Brown, Wyatt, Harris, and
  Yao]{Brown_Yao:2005}
Brown,~G.; Wyatt,~J.; Harris,~R.; Yao,~X. Diversity creation methods: a survey
  and categorization. \emph{Information Fusion} \textbf{2005}, \emph{6},
  5--20\relax
\mciteBstWouldAddEndPuncttrue
\mciteSetBstMidEndSepPunct{\mcitedefaultmidpunct}
{\mcitedefaultendpunct}{\mcitedefaultseppunct}\relax
\EndOfBibitem
\bibitem[Chan and Kasabov(2005)Chan, and Kasabov]{Chan:2005}
Chan,~Z. S.~H.; Kasabov,~N. Fast Neural Network Ensemble Learning via
  Negative-Correlation Data Correction. \emph{IEEE Transactions on Neural
  Networks} \textbf{2005}, \emph{16}, 1707--1710\relax
\mciteBstWouldAddEndPuncttrue
\mciteSetBstMidEndSepPunct{\mcitedefaultmidpunct}
{\mcitedefaultendpunct}{\mcitedefaultseppunct}\relax
\EndOfBibitem
\bibitem[Reeve and Brown(2018)Reeve, and Brown]{Reeve_Brown:2018}
Reeve,~H.~W.; Brown,~G. Diversity and degrees of freedom in regression
  ensembles. \emph{Neurocomputing} \textbf{2018}, \emph{298}, 55--68\relax
\mciteBstWouldAddEndPuncttrue
\mciteSetBstMidEndSepPunct{\mcitedefaultmidpunct}
{\mcitedefaultendpunct}{\mcitedefaultseppunct}\relax
\EndOfBibitem
\bibitem[Tumer and Ghosh(2002)Tumer, and Ghosh]{Tumer:2002}
Tumer,~K.; Ghosh,~J. Robust Combining of Disparate Classifiers through Order
  Statistics. \emph{Pattern Analysis \& Applications} \textbf{2002}, \emph{8},
  189--200\relax
\mciteBstWouldAddEndPuncttrue
\mciteSetBstMidEndSepPunct{\mcitedefaultmidpunct}
{\mcitedefaultendpunct}{\mcitedefaultseppunct}\relax
\EndOfBibitem
\bibitem[Ahmad and Zhang(2009)Ahmad, and Zhang]{AHMAD:2009}
Ahmad,~Z.; Zhang,~J. Selective combination of multiple neural networks for
  improving model prediction in nonlinear systems modelling through forward
  selection and backward elimination. \emph{Neurocomputing} \textbf{2009},
  \emph{72}, 1198 -- 1204\relax
\mciteBstWouldAddEndPuncttrue
\mciteSetBstMidEndSepPunct{\mcitedefaultmidpunct}
{\mcitedefaultendpunct}{\mcitedefaultseppunct}\relax
\EndOfBibitem
\bibitem[Abadi \latin{et~al.}(2015)Abadi, Agarwal, Barham, Brevdo, Chen, Citro,
  Corrado, Davis, Dean, Devin, Ghemawat, Goodfellow, Harp, Irving, Isard, Jia,
  Jozefowicz, Kaiser, Kudlur, Levenberg, Man\'{e}, Monga, Moore, Murray, Olah,
  Schuster, Shlens, Steiner, Sutskever, Talwar, Tucker, Vanhoucke, Vasudevan,
  Vi\'{e}gas, Vinyals, Warden, Wattenberg, Wicke, Yu, and Zheng]{tensorflow}
Abadi,~M. \latin{et~al.}  {TensorFlow}: Large-Scale Machine Learning on
  Heterogeneous Systems. 2015; \url{https://www.tensorflow.org/}, Software
  available from tensorflow.org\relax
\mciteBstWouldAddEndPuncttrue
\mciteSetBstMidEndSepPunct{\mcitedefaultmidpunct}
{\mcitedefaultendpunct}{\mcitedefaultseppunct}\relax
\EndOfBibitem
\bibitem[Wang \latin{et~al.}(2020)Wang, Yao, Kwok, and Ni]{Wang:2020}
Wang,~Y.; Yao,~Q.; Kwok,~J.; Ni,~L.~M. Generalizing from a Few Examples: A
  Survey on Few-Shot Learning. 2020\relax
\mciteBstWouldAddEndPuncttrue
\mciteSetBstMidEndSepPunct{\mcitedefaultmidpunct}
{\mcitedefaultendpunct}{\mcitedefaultseppunct}\relax
\EndOfBibitem
\bibitem[Granger(1989)]{Granger:1989}
Granger,~C. W.~J. Invited review combining forecasts—twenty years later.
  \emph{Journal of Forecasting} \textbf{1989}, \emph{8}, 167--173\relax
\mciteBstWouldAddEndPuncttrue
\mciteSetBstMidEndSepPunct{\mcitedefaultmidpunct}
{\mcitedefaultendpunct}{\mcitedefaultseppunct}\relax
\EndOfBibitem
\bibitem[Wallis(2011)]{Wallis:2011}
Wallis,~K.~F. Combining forecasts – forty years later. \emph{Applied
  Financial Economics} \textbf{2011}, \emph{21}, 33--41\relax
\mciteBstWouldAddEndPuncttrue
\mciteSetBstMidEndSepPunct{\mcitedefaultmidpunct}
{\mcitedefaultendpunct}{\mcitedefaultseppunct}\relax
\EndOfBibitem
\end{mcitethebibliography}


\appendix
\section{Zero-Bias Networks}
\label{appendix:A}
Let a network $\ub N(x)$
 and a training set $\ds T_r =\{x_i, y_i= y(x_i)\}_{i=1,M}$. Then the network 
\begin{equation}\label{ZERO}
N(x) \equiv \ub N(x) -\la \ub N\ra +\la y\ra,
\end{equation}
is by construction a zero-bias network on \{$\ds T_r$\}. 
Clearly the misfit is 
\begin{equation}
m(x)=N(x)-y(x)= \ub N(x) -\la \ub N\ra +\la y\ra-y(x),\hbox{\ satisfying\ } \la m\ra=0.
\end{equation}
At this point one may rewrite the misfit as:
\begin{equation}
m(x)= (\ub N(x) - y(x))-(\la \ub N-y\ra) \equiv \ub m(x)-\la \ub m\ra,
\end{equation}
and the MSE becomes:  
\begin{equation}
\la m^2\ra =\la (\ub m-\la \ub m\ra)^2\ra\ =\  \la \ub m^2\ra-\la \ub m\ra^2.
\end{equation}
Thus minimizing $\ds\la m^2\ra$ subject to $\la m\ra = 0$, is equivalent to minimizing \mbox{$\la \ub m^2\ra-\la \ub m\ra^2$}, subject to no constraints. The zero-bias network $N(x)$, is then recovered from eq. (\ref{ZERO}).

\section{Blending Restriction}
\label{appendix:B}
The blending coefficient may be further restricted to lay in a shorter interval, e.g. 
\begin{equation}
\label{RESTRI}
\epsilon \leq \beta_{k+1} \leq 1-\epsilon,\ \ \hbox{\ with\ \ } \epsilon \in [0,0.5)
\end{equation}
This imposes a slightly different constraint on the correlation $\la  M_km_{k+1} \ra$, instead of (\ref{INEQ}), i.e.:
\begin{equation}
 \la M_km_{k+1}\ra \ \  \leq\ \ \frac{1}{1-2\epsilon} \left[\min\{\la M_k^2\ra,\la m^2_{k+1}\ra\}  - \epsilon\left(\la M_k^2\ra+\la m^2_{k+1}\ra\right)\right].
\end{equation}
A typical value,  $\epsilon = 0.1$ is restricting $\beta_{k+1} \in [0.1,0.9]$.

\end{document}